\documentclass[preprint,sort&compress,11pt,3p]{elsarticle}
\usepackage{moreverb}
\usepackage{mathtools}
\usepackage{amsmath}
\usepackage{amssymb}
\usepackage{algorithmic}
\usepackage{graphics}
\usepackage{graphicx}
\usepackage{subfigure}
\usepackage{caption}
\usepackage{enumitem}
\usepackage{extarrows}
\usepackage{color}
\usepackage{framed}
\usepackage{wrapfig}
\usepackage{bm}
\usepackage{mathrsfs}
\usepackage{mathabx}
\usepackage{multirow}
\usepackage{longtable}
\usepackage{hyperref}
\usepackage{paralist}
\usepackage{indentfirst}
\usepackage{relsize}
\usepackage{extarrows}
\usepackage{lineno,xcolor}
\usepackage{upgreek}
\usepackage{bm}
\usepackage{mwe}
\usepackage{xcolor}
\usepackage{booktabs}
\usepackage[flushleft]{threeparttable}
\usepackage[linesnumbered,lined,ruled]{algorithm2e}

\newcommand{\eref}[1]{(\ref{#1})}

\hypersetup{
bookmarks=true,
bookmarksopen=true,
bookmarksnumbered=true,
unicode=false,
pdftoolbar=true,
pdfmenubar=true,
pdffitwindow=false,
pdfstartview={FitH},
pdftitle={My title},
pdfauthor={Author},
pdfsubject={Subject},
pdfcreator={Creator},
pdfproducer={Producer},
pdfkeywords={keywords},
pdfnewwindow=true,
colorlinks=true,
linkcolor=blue,
citecolor=blue,
filecolor=magenta,
urlcolor=blue
}

\journal{Mechanical Systems and Signal Processing}

\begin{document}
	
\title{\Large Incremental Bayesian Tensor Learning for Structural Monitoring Data Imputation and Response Forecasting}

\author[NU]{Pu Ren}
\ead{ren.pu@northeastern.edu}
\author[MU]{Xinyu Chen}
\author[MU]{Lijun Sun}
\ead{lijun.sun@mcgill.ca}
\author[NU,MIT]{Hao Sun\corref{cor}}
\ead{h.sun@northeastern.edu}
\cortext[cor]{Corresponding author. Tel: +1 617-373-3888}

\address[NU]{Department of Civil and Environmental Engineering, Northeastern University, Boston, MA 02115, USA}
\address[MU]{Department of Civil Engineering, McGill University, Montreal, Quebec H3A0G4, Canada}
\address[MIT]{Department of Civil and Environmental Engineering, MIT, Cambridge, MA 02139, USA}

\begin{abstract}
	\small
    There has been increased interest in missing sensor data imputation, which is ubiquitous in the field of structural health monitoring (SHM) due to discontinuous sensing caused by sensor malfunction. Recent development in Bayesian temporal factorization models for high-dimensional time series analysis has provided an effective tool solve both imputation and prediction problems. However, for large datasets, the default Bayesian temporal factorization model becomes less inefficient since the model has to be fully retrained when new data arrives. A potential solution is to train the model using a short time window covering only most recent data; however, by doing so, we may miss some critical dynamics and long-term dependencies which can only be identified from a longer time window. To address this fundamental issue in temporal factorization models, this paper presents an incremental Bayesian tensor learning scheme to achieve efficient imputation and prediction of structural response in long-term SHM. In particular, a spatiotemporal tensor is first constructed followed by Bayesian tensor factorization that extracts latent features for missing data imputation. To enable structural response forecasting based on long-term and incomplete sensing data, we develop an incremental learning scheme to effectively update the Bayesian temporal factorization model. The performance of the proposed approach is validated on continuous field-sensing data (including strain and temperature records) of a concrete bridge, based on the assumption that strain time histories are highly correlated to temperature recordings. The results indicate that the proposed probabilistic tensor learning framework is accurate and robust even in the presence of large rates of random missing, structured missing and their combination. The effect of rank selection on the imputation and prediction performance is also investigated. The results show that a better estimation accuracy can be achieved with a higher rank for random missing whereas a lower rank for structured missing. 
\end{abstract}


\maketitle

\section{Introduction}\label{S:1}
High-quality data plays a pivotal role in structural health monitoring (SHM) for condition assessment, damage detection, and decision making. However, during long-term monitoring, it is inevitable for imperfect and corrupted sensor measurements, especially in a harsh and noisy environment, which calls for effective approaches for imputation/recovery missing and noisy data. Furthermore, in order to conduct real-time early-warning of structural deterioration or even disastrous failure, forecasting/prediction of structural response has also received considerable attention. The general idea of time series analysis, in the context of imputation and forecasting, is to find key dynamic patterns from observations and establish a mapping function between the historical records and the estimation. Nevertheless, these tasks are rather challenging on account of complex spatiotemporal dependencies and inherent difficulty in large-scale and nonlinear characteristics of SHM data, especially in piratical applications.

There have been a number of attempts made to solve the data imputation and forecasting problems in the SHM community. On one hand, in the missing data recovery research, compressive sensing is one common and typical approach to rebuild the entire temporal signals based on the the sparsity assumption of the data in certain feature spaces \cite{bao2013compressive,bao2015compressive,huang2014robust,bao2020compressive,cao2017deformation}. Another interesting stream for data imputation is the use of probability methods (e.g., Gaussian process (GP) thanks to its great interpretation capacity for nonlinear dynamic processes), which has been comprehensively studied in outlier detection \cite{yuen2012novel,yuen2017outlier}, model calibration/updating \cite{sun2015statistical,sun2016bayesian,behmanesh2016effects,sun2017bayesian,song2019modeling,song2019hierarchical,uzun2019structural,chen2020sparse} and system identification \cite{yin2017vibration,avendano2017gaussian,avendano2018gaussian,avendano2019modelling,kopsaftopoulos2013functional,amer2019probabilistic}. For instance, Wan \emph{et al.} \cite{wan2019bayesian} employed Bayesian multi-task learning with multi-dimensional GP priors to recover SHM data. Chen \emph{et al.} \cite{chen2018novel} explored the possibility of probability density function estimation for data loss compensation with warping transformations. Some recent surveys have reported the great potential in data imputation by considering both the sensor information and time series, which is usually conceptualized as spatiotemporal. Yang \emph{et al.} \cite{yang2016harnessing} developed a low-rank matrix completion method with $\ell_1$-norm and a nuclear norm for imputation of random missing data. This approach is powerful but has limitations due to an ideal assumption that the data is randomly missing, which is less common in practical SHM (e.g., data might be missing for a continuous duration). Chen \emph{et al.} \cite{chen2019analyzing} investigated the inter-sensor relationship of stochastic structural responses with non-parametric copulas, which flexibly captured the spatial dependency for strain data. Moreover, the sequential broad learning (SBL) approach was recently presented for efficiently reconstructing structural response \cite{kuok2020model}, which is however short for spatial consideration. On the other hand, for the sake of data-driven structural response forecasting, the majority of existing research focus on the time-dependent response approximation based on high-quality collected data (e.g., data missing is not considered). In particular, the widely-accepted and well-studied methods are based on the linear combination of previous observations, for example, dynamic linear models \cite{fan2017bridge,goulet2017bayesian,wang2019modeling} and autoregressive (AR) models \cite{park2007structural,bornn2009structural,okasha2011reliability}. Distinctively, Wan and Ni \cite{wan2018bayesian} examined the capability of a GP-based Bayesian approach for underlying nonlinear dynamic system response prediction from a statistic perspective. Besides, deep learning techniques, such as the convolutional neural network (CNN) \cite{wu2018deep,fan2020dynamic,zhang2020physicsES}, the long-short term memory (LSTM) network \cite{zhang2019deep,zhang2020physics}, and the variational autoencoder (VAE) \cite{mylonas2020deep}, have also been proven to be a decent alternative for extracting spatial features for dynamic response reconstruction and prediction.

Despite the rapid development of data science in SHM, there still remain three representative challenges for the specific aim of data imputation and response forecasting. Firstly, very little work has been devoted to the spatial dependency and correlation in the time series analysis. The second is the lack of consideration on vast and continuous missing scenarios (e.g., data missing for a long continuous period such as one day or consecutive days). Lastly, almost all of the present studies on response forecasting are based on high-quality data instead of imperfect measurements with missing values. To this end, in light of the recent renaissance in tensor learning \cite{kolda2009tensor,anandkumar2014tensor,janzamin2020spectral}, which has already greatly contributed to image processing \cite{zhao2015bayesian,zhang2015compression,lu2016tensor,du2016pltd,zhang2016tensor,shi2018feature}, recommender systems \cite{karatzoglou2010multiverse,ifada2014tensor,seko2018matrix}, and traffic data analysis \cite{yu2016temporal,tan2013tensor,asif2016matrix,tan2016short,takeuchi2017autoregressive,deng2016latent,chen2018spatial,chen2019bayesian,sun2019bayesian,chen2020nonconvex,chen2020low}. In the context of SHM, we can naturally consider the data as multivariate time-series matrix and then apply temporal factorization models (e.g., \cite{sun2019bayesian}) where the low-rank representation can effectively characterize the complex spatial and temporal dependencies rooted in the data. However, a fundamental limitation of these factorization-based models lies in their inefficiency in dealing with streaming data: the model has to be fully retrained whenever new data arrives to the system. This poses a critical challenge for SHM which requires efficient models to account for continuous monitoring. To address this issues, in this paper we propose an incremental Bayesian learning scheme, based on Chen and Sun \cite{sun2019bayesian}, that enables imputation of SHM data and forecasting of structural response in a long-term horizon for temporal/continuous SHM. Instead of training on the full data, we propose an incremental updating scheme leveraging locally streaming data, resulting in more efficient and more accurate imputation/prediction for long-term SHM data. In particular, we employ the proposed learning approach for (1) reconstruction of spatiotemporal missing data in SHM and (2) forecasting of structural response under the scenario of missing/incomplete data. It is worthy to mention that, different from \cite{yang2016harnessing}, tensor factorization in the context of Bayesian inference \cite{sun2019bayesian} provides a principled selection mechanism for suitable likelihood models and allows for uncertainty quantification in parameter estimation and prediction \cite{rai2014scalable}. In addition, inspired by the strong correlation between strain data and temperature data \cite{xu2010monitoring,duan2011strain,xia2017service,zhu2018thermal,wang2019modeling}, this research sheds new light on integrating physics into the tensor model, resulting in an interpretative low-rank data structure.

The main contribution of this paper can be summarized as follows. Firstly, to the best of our knowledge, it is the first time to realize response forecasting with incomplete data in SHM applications, based on reliable latent features instead of directly using the corrupted data. Secondly, by constructing one-dimensional time series data into a matrix (i.e., second-order tensor) structure (sensor locations $\times$ time steps), we can easily capture the spatiotemporal features of the data for accurate imputation and forecasting. Thirdly, the physics relationship between strain and temperature is introduced to optimize the tensor structure. Fourthly, we propose an incremental learning scheme to tackle practical continuous monitoring problems and speed up the tensor factorization process through efficient updating. We further validate the proposed approach on a concrete bridge with multi-year recordings of strain and temperature time histories.

The rest of the paper is organized as follows, in addition to this Introduction section. Section \ref{S:2} begins by laying out the theoretical dimensions of this work, and is concerned with the proposed methodology. In Section \ref{S:2.1}, we describe the problem definition and general principle of data imputation and response forecasting under the data missing scenarios. In Section \ref{S:2.1-add}, we introduce the incremental Bayesian tensor learning architecture and two adaptive factor updating stages. In Section \ref{S:2.2}, \ref{S:2.3} and \ref{S:2.4}, we circumstantially present the Bayesian generation and inference procedure, as well as the autoregressive process for temporal feature modeling. Section \ref{S:3} elaborates the experimental validation results of the proposed method, focusing on three key themes: imputation and forecasting performance with respect to different missing rates, uncertainty quantification and rank analysis. Section \ref{S:4} concludes the current work and the outlook of future directions. 

\section{Methodology}\label{S:2}
In this section, we formulate the problem of SHM data imputation and response forecasting in the context of incremental Bayesian tensor learning, and present the spatiotemporal dependency modeling procedure via matrix factorization. 

\subsection{Problem description}\label{S:2.1}
The goal of continuous/steaming SHM data imputation and forecasting is to estimate the missing values and predict the future structural response given partially observed data collected from a sensor network. The multidimensional time series data, with missing values, can be represented by matrix $\mathbf{Y} \in \mathbb{R}^{M\times T}$, where $M$ denotes the number of sensor locations and $T$ is the number of time stamps for a certain continuously monitoring period. The imputation process aims to firstly learn a factorized spatial feature $\mathbf{U}$ and a temporal feature $\mathbf{X}$ based on the observed data $\mathbf{Y}$, and then reconstruct the response with imputed values. Afterwards, given $\mathbf{y}_{:,t}$ signifying the multivariate data at time $t$, the course of response forecasting utilizes the well-trained spatial factor $\mathbf{U}$ and the updated temporal factor $\mathbf{X}^{*}$ to map $L$ ($\geq 1$) historical sensing data to future $\mathcal{T}$ ($\geq 1$) structural responses, given by
\begin{equation}
    \label{forecasting_illusration}
    [\mathbf{y}_{:,t-L+1}, \cdots, \mathbf{y}_{:,t}] 
    \xlongrightarrow[\mathbf{X}^{*}]{\mathbf{U}}
    [\mathbf{y}_{:,t+1}, \cdots, \mathbf{y}_{:,t+\mathcal{T}}]
\end{equation}
which essentially establishes a temporal forecasting process.

\subsection{Incremental learning scheme}\label{S:2.1-add}
\begin{figure}[t!]
    \centering
    \includegraphics[width=0.75\linewidth]{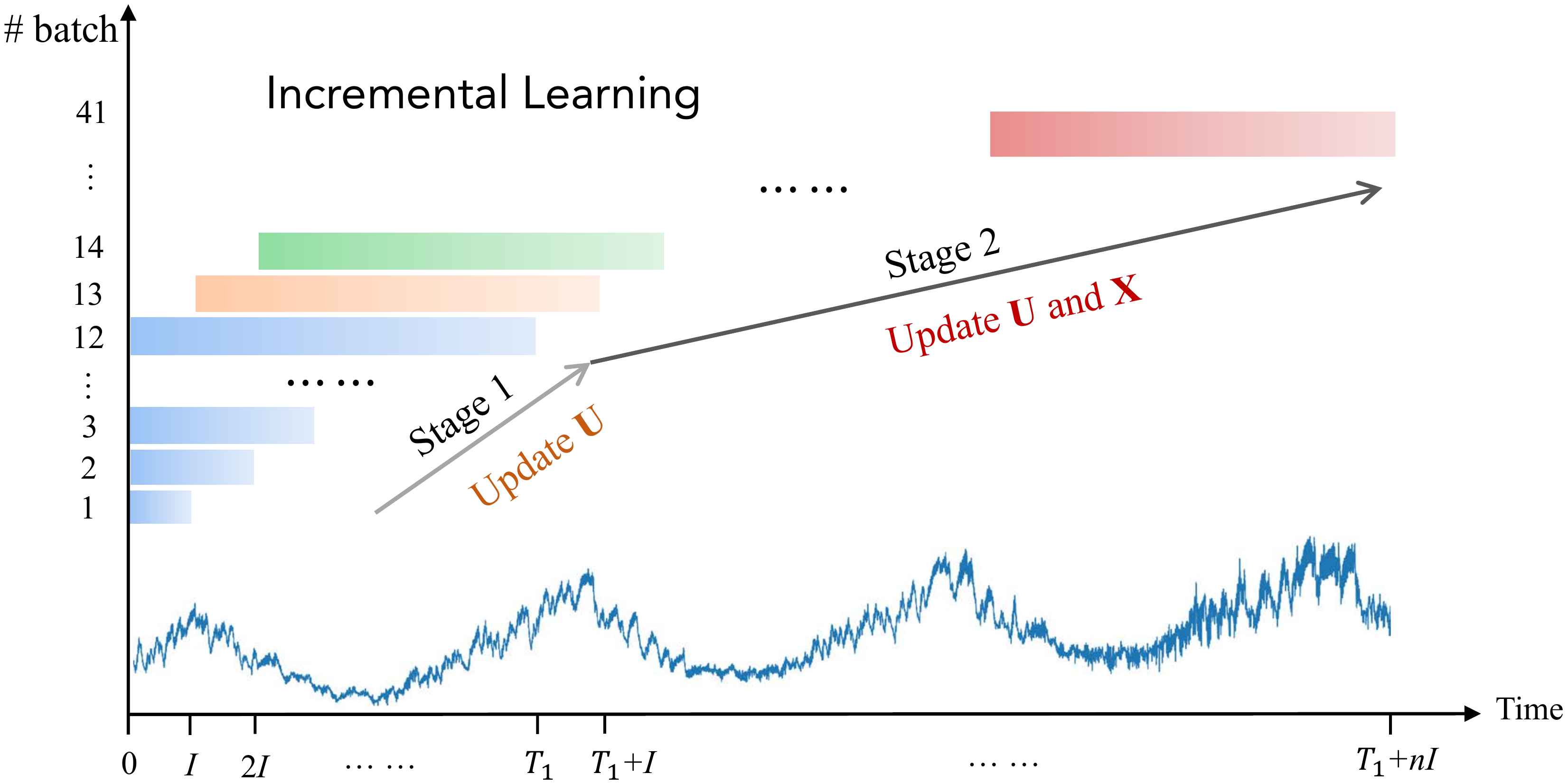}
    \caption{The proposed incremental learning scheme. Note that the dynamic batch window size is  $I,2I, \cdots, T_1$ for each short-time updating period respectively, while the the long-period fixed batch window remains constant (e.g. $T_1$) after reaching a critical point (e.g., one year).}
    \label{fig:batch_window}
\end{figure}

For continuous SHM, data streams over time where imputation and forecasting should be ideally done in a real-time manner accounting for new records. This typically requires online learning with model re-training involved, resulting in significant computational burden especially when large-scale data analysis is performed. To this end, inspired by the work in \cite{deng2016latent}, we present an incremental learning scheme as illustrated in Figure \ref{fig:batch_window}. Instead of retraining the entire model when new data arrives, in the incremental scheme we only take the up-to-date information from sensors within a certain number of time steps for intermittent training/updating. The benefits of this proposed scheme are two-fold: (1) possessing efficiency and alleviating the computational burden induced by online model re-training for every time step, and (2) maintaining satisfactory accuracy thanks to the use of streaming sensing data. 


Specifically, there are two updating stages for data imputation and response forecasting in continuous SHM: (1) short-time dynamic batch window and (2) long-period fixed batch window. The dynamic batch window stage learns the latent spatial attribute $\mathbf{U}$ from previous records during the time interval $[0, I]$ when $I$ is small (e.g., one month), where $I$ denotes the length of data for forward imputation period. Then we fix $\mathbf{U}$ for response forecasting within $[I,2I]$ and execute the next spatial information updating using records during $[0,2I]$. The rest of the first dynamic tensor learning stage will follow the same manner until the total imputation time reaches one critical time stamp $T_1$ (e.g., one year), where $\mathbf{U}$ is incrementally updated. In the second stage, we update the spatial attribute $\mathbf{U}$ and the temporal factor $\mathbf{X}$ simultaneously due to the constant temporal dimension of the fixed batch window. For instance, $\mathbf{U}$ and $\mathbf{X}$ will be updated in the time period $[I, T_1+I]$ for imputation, then we still keep $\mathbf{U}$ unchanged and perform forecasting within $[T_1+I,T_1+2I]$. Subsequently, we repeat the imputation and forecasting procedure with the fixed batch window as the continuous monitoring proceeds. The basic concept of the proposed incremental learning scheme for semi-online tensor learning is presented in Figure \ref{fig:batch_window}. Such a scheme enables recovery of missing data incrementally (e.g., every $I$-unit increment) and forecasting of structural response on the fly for long-term SHM. The process can be realized through a Bayesian tensor learning approach \cite{sun2019bayesian}, which is introduced in the following.

\subsection{Hierarchical Bayesian modeling for tensor decomposition}\label{S:2.2}

\begin{figure}[t!]
    \centering
    \includegraphics[width=0.85\linewidth]{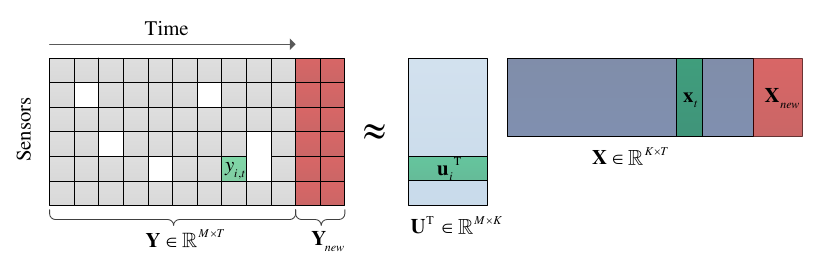}
    \caption{A graphic illustration of matrix factorization. Note that the white boxes represent the missing values while the grey boxes denote the observed data.}
    \label{fig:MF}
\end{figure}

Naturally, spatiotemporal SHM data observed from $M$ sensor locations with $T$ time stamps can be constructed in the form of a two-dimensional tensor,  $\mathbf{Y} \in \mathbb{R}^{M\times T}$. Due to inevitable data missing in practical applications, we define an indicator set for the observed elements in $\mathbf{Y}$ as $\Omega = \{(i,t)|y_{i,t} \text{ is observed}\}$. To characterize the spatiotemporal dependencies, we employ the general idea of second-order tensor (matrix) decomposition to approximate the multidimensional data through the sum of $K$ rank-1 tensors, namely,
\begin{equation}
    \label{MF}
    \mathbf{Y} \approx \sum_{r=1}^{K} \mathbf{u}_r \circ \mathbf{x}_r  = 
    \mathbf{U}^\top \mathbf{X},
\end{equation}
\noindent where $K$ is a positive integer referring to the tensor rank, and the symbol $\circ$ stands for the vector outer product. Here, $\mathbf{u}_1, \mathbf{u}_2, \cdots, \mathbf{u}_K \in \mathbb{R}^M$ and $\mathbf{x}_1, \mathbf{x}_2, \cdots, \mathbf{x}_K \in \mathbb{R}^T$ form the rank-1 components of the matrix $\mathbf{Y}$. Furthermore, with this formulation, we assume $\mathbf{U}$ as the spatial latent factor whose rows are $\mathbf{u}_r$'s, and $\mathbf{X}$ to be the temporal latent feature whose rows are $\mathbf{x}_r$'s. Element-wise, $y_{i,t}$ is estimated by the inner product of $\mathbf{u}_i$ and $\mathbf{x}_t$, where $\mathbf{u}_i \in \mathbb{R}^K$ represents the latent spatial feature at sensor $i$ and $\mathbf{x}_t \in \mathbb{R}^K$ is the latent temporal embedding at time $t$, expressed as
\begin{equation}
    \label{element_wise_MF}
    y_{i,t} \approx \mathbf{u}_i^\top \mathbf{x}_t.
\end{equation}
The basic concept of matrix factorization is illustrated in Figure \ref{fig:MF}.

Next, we introduce the fully Bayesian method for tensor learning \cite{sun2019bayesian}. To begin this process, the likelihood of the observed SHM data $y_{i,t}$ is given by:
\begin{equation}
    \label{y_it}
    y_{i,t} \sim \mathcal N(\mathbf{u}_i^\top \mathbf{x}_t, \tau_\epsilon^{-1}),
\end{equation}
where $\mathcal N(\cdot)$ denotes the Gaussian distribution with mean $\mathbf{u}_i^\top \mathbf{x}_t$ and precision $\tau_\epsilon$. Secondly, to model the spatial factor, the prior distribution over the spatial feature vectors (i.e., $\mathbf{u}_i$) is assumed to be multivariate Gaussian, viz.,
\begin{equation}
    \label{u_i}
    \mathbf{u}_i \sim \mathcal N(\boldsymbol{\upmu}_u, \mathbf{\Lambda}_u^{-1}). 
\end{equation}
We further place conjugate Gaussian-Wishart priors on the spatial feature parameters $\boldsymbol{\Theta}_u = \{\boldsymbol{\upmu}_u,\mathbf{\Lambda}_u\}$, i.e., mean $\boldsymbol{\upmu}_u \in  \mathbb{R}^K$ and variance $\mathbf{\Lambda}_u \in \mathbb{R}^{K \times K}$, written as \cite{sun2019bayesian}
\begin{equation}
    \label{mu_u}
    \begin{split}
        p(\boldsymbol{\Theta}_u|\boldsymbol{\upmu}_0, \beta_0, \mathbf{W}_0, v_0) &= p(\boldsymbol{\upmu}_u|\mathbf{\Lambda}_u)p(\mathbf{\Lambda}_u)\\   
        & =\mathcal{N}(\boldsymbol{\upmu}_u|\boldsymbol{\upmu}_0, (\beta_0 \mathbf{\Lambda}_u)^{-1}) \mathcal{W}(\mathbf{\Lambda}_u|\mathbf{W}_0, v_0).
    \end{split}
\end{equation}
Here, $\boldsymbol{\upmu}_0, \beta_0, \mathbf{W}_0, v_0$ are hyper-parameters; $\mathcal W(\cdot)$ denotes the Wishart distribution with $v_0$ degrees of freedom and a $K \times K$ scale matrix $\mathbf{W}_0$, namely,
\begin{equation}
    \label{wishart}
    \mathcal{W}(\mathbf{\Lambda}_u|\mathbf{W}_0, v_0) = \frac{1}{C}|\mathbf{\Lambda}_u|^{\frac{v_0-K-1}{2}}
    \text{exp} \left(-\frac{1}{2} \text{Tr} \left(\mathbf{W}_0^{-1} \mathbf{\Lambda}_u\right)\right),
\end{equation}
where $C$ is the normalizing constant and Tr($\cdot$) denotes the matrix trace defined as the sum of all the elements on the main diagonal of the matrix. 

\begin{figure}[t!]
    \centering
    \includegraphics[width=0.65\linewidth]{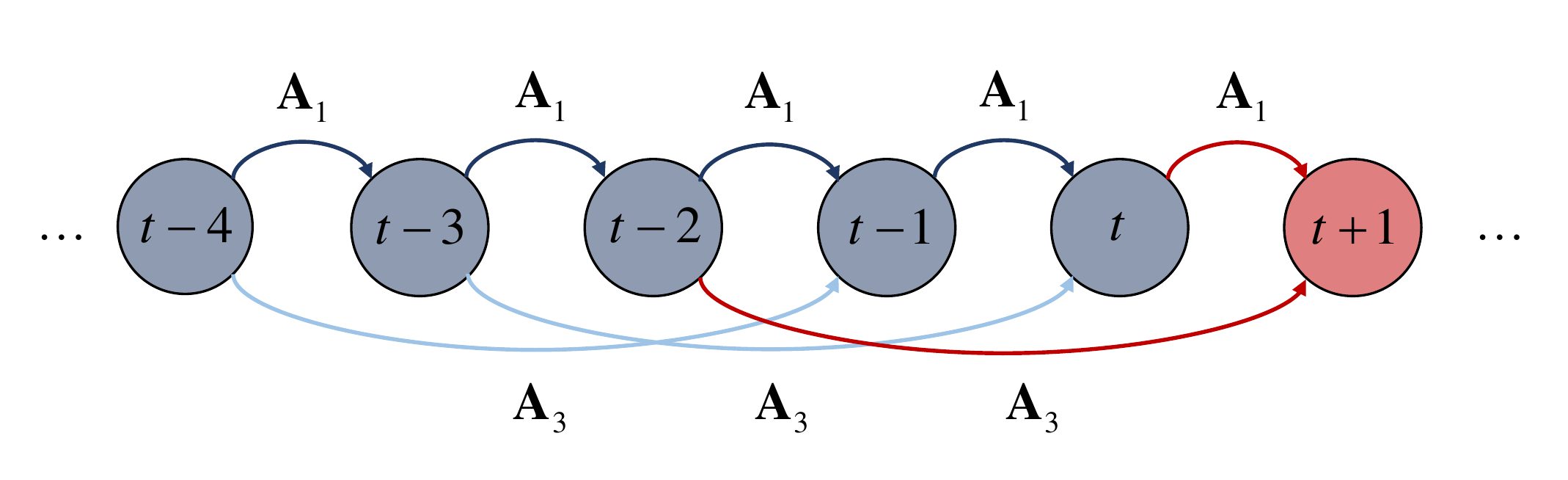}
    \caption{Auto-regressive model for temporal dependencies.}
    \label{fig:time_series}
\end{figure}

Although probabilistic modeling of the spatial factors is straightforward, it is tricky to capture the time-evolving patterns and predict the dynamic trends in the Bayesian learning. Here, we consider incorporating the AR process into the matrix/tensor factorization model for describing the temporal dependencies \cite{sun2019bayesian}. Generally, an AR model is characterized by a time lag set and a weight parameter vector. However, different from the traditional AR model which is more applicable for low-dimensional data, we make two modifications to handle the multi-dimensional time-series issue. The first distinction is that we introduce a flexible AR structure on time lags $\mathcal{L}$  \cite{yu2016temporal}. Instead of applying a small-size lag set (e.g., $\mathcal{L}=\{1\}$) which only learns the simple temporal patterns (e.g., daily similarity), we try to use more complex time lags to infer seasonal or yearly trends for long-term forecasting. The second alteration is changing the tensor structure of AR model parameters for convenience. Let the time lags set be $\mathcal{L}=\{l_1, l_2, \cdots, l_d\}$, where $d$ is the order of the AR model. In our case, the weight parameter $\mathbf{A}_j$ ($j \in \{1,2, \cdots, d\}$) should be a $K \times K$ matrix since the elements in the AR model are formed as column vectors (i.e., $\mathbf{x}_t \in \mathbb{R}^{K \times 1}$) in the temporal feature matrix $\mathbf{X}$. The graphic illustration is shown in Figure \ref{fig:time_series} with the example of $\mathcal{L}=\{1,3\}$.

In addition, there is no diagonal restriction on $\mathbf{A}_j$ due to the complicated causal relationship between factors. Thus, the reorganized formulation of the AR process can be written as:
\begin{equation}
    \begin{split}
        \label{AR}
        \mathbf{x}_{t+1} 
        & \approx \mathbf{A}_1 \mathbf{x}_{t+1-l_1} + \mathbf{A}_2 \mathbf{x}_{t+1-l_2} + \cdots + \mathbf{A}_d \mathbf{x}_{t+1-l_d} \\ 
        & =  \underbrace{[\mathbf{A}_1, \mathbf{A}_2, \cdots, \mathbf{A}_d]}_{\mathbf{A}}
        \underbrace{[\mathbf{x}_{t+1-l_1}, \mathbf{x}_{t+1-l_2}, \cdots, \mathbf{x}_{t+1-l_d}]^\top}_{\mathbf{z}_{t+1}}.
    \end{split}
\end{equation}
For simplicity, we define a time-invariant matrix $\mathbf{A} \in \mathbb{R}^{(Kd) \times K}$ and a historical observation vector $\mathbf{z}_{t+1} \in \mathbb{R}^{(Kd) \times 1}$ shown in Eq. \eref{AR}. As a result, by assuming the prior distribution for the temporal factor $\mathbf{x}_t$ as multivariate Gaussian, we have the mean vector as $\mathbf{A}^\top \mathbf{z}_{t}$ for the forecasting process. 
Therefore, the piecewise modeling of the temporal feature matrix is summarized as:
\begin{equation}
    \label{x_t}
    \begin{split}
        \mathbf{x}_t &\sim \mathcal N({\widetilde{\boldsymbol{\upmu}}_x}, \widetilde{\boldsymbol{\Sigma}}_x)\\
        & \sim 
        \begin{cases} 
            \mathcal N(\mathbf{0}, \mathbf{I}_x), & \mbox{if } t \in \{1, 2, \cdots, l_d\}, \\
            \mathcal N(\mathbf{A}^\top \mathbf{z}_t, \boldsymbol{\Sigma}), & \mbox{otherwise},
        \end{cases}
    \end{split}
\end{equation}
where $\mathbf{0} \in \mathbb{R}^{K \times 1}$ is a zero vector and $\mathbf{I}_x \in \mathbb{R}^{K \times K}$ is an identity matrix. 


Likewise, a conjugate Matrix Normal Inverse Wishart prior is applied to the hyper-parameters $\boldsymbol{\Theta}_x = \{\mathbf{A}, \boldsymbol{\Sigma}\}$ in the forecasting process \cite{sun2019bayesian}:
\begin{equation}
    \label{A_Sigma}
    \begin{split}
        p(\boldsymbol{\Theta}_x|\mathbf{\Lambda}_0, \mathbf{V}_0, \boldsymbol{\Psi}_0, v_0) 
        & = p(\mathbf{A}|\mathbf{\Sigma})p(\mathbf{\Sigma})\\
        & = \mathcal{MN}(\mathbf{A}|\mathbf{\Lambda}_0, \mathbf{V}_0, \boldsymbol{\Sigma})
        \mathcal{IW}(\boldsymbol{\Sigma}|\boldsymbol{\Psi}_0, v_0),
    \end{split}
\end{equation}
where $\mathcal{MN}(\cdot)$ is Matrix Normal distribution and $\mathcal{IW}(\cdot)$ denotes Inverse Wishart function. Herein, the Inverse-Wishart distribution $\boldsymbol{\Sigma} \sim \mathcal{IW}(\boldsymbol{\Psi}_0, v_0)$ is equivalent to $\boldsymbol{\Sigma}^{-1} \sim \mathcal{W}(\boldsymbol{\Psi}_0^{-1}, v_0)$. Besides, the probability density function (PDF) for $\mathbf{A}$ is given by
\begin{equation}
    \label{matrix_normal}
    p(\mathbf{A}|\mathbf{\Lambda}_0, \mathbf{V}_0, \boldsymbol{\Sigma}) = (2\pi)^{-\frac{K^{2}d}{2}} |\mathbf{V}_0|^{-\frac{K}{2}} |\mathbf{\Sigma}|^{-\frac{Kd}{2}} \text{exp}\left(-\frac{1}{2}\left[\text{Tr}(\boldsymbol{\Sigma}^{-1}(\mathbf{A}-\mathbf{\Lambda}_0)^\top \mathbf{V}_0 (\mathbf{A}-\mathbf{\Lambda}_0))\right]\right). 
\end{equation}
in which $\mathbf{\Lambda}_0 \in \mathbb{R}^{(Kd) \times K}$ is the mean matrix parameter, $\mathbf{V}_0 \in \mathbb{R}^{(Kd)\times(Kd)}$ represents the row-variance matrix, and $\mathbf{\boldsymbol{\Sigma}} \in \mathbb{R}^{K\times K}$ denotes the column-variance matrix parameter.

The final stage of model generation is to deal with the precision parameter $\tau_\epsilon$ as shown in Eq. \eref{y_it}. In particular, a conjugate Gamma prior over $\tau_\epsilon$ is introduced to make the generative model robust in consideration of the indeterminate noise effect in SHM data:
\begin{equation}
    \label{gamma}
    \tau_\epsilon \sim \text{Gamma}(a_0, b_0)
\end{equation}
Here we define $\boldsymbol{\Theta}_\tau = \{a_0, b_0\}$ where $a_0$ and $b_0$ represent the shape parameter and the rate parameter, respectively. The PDF of $\tau_\epsilon$ has the form as follows
\begin{equation}
    \label{gamma_func}
    p(\tau_\epsilon|a_0,b_0) = \frac{b_0^{a_0}}{\Gamma(a_0)}
    \tau_\epsilon^{a_0-1} \text{exp}(-b_0 \tau_\epsilon).
\end{equation}

\begin{figure}[t!]
    \centering
    \includegraphics[width=0.45\linewidth]{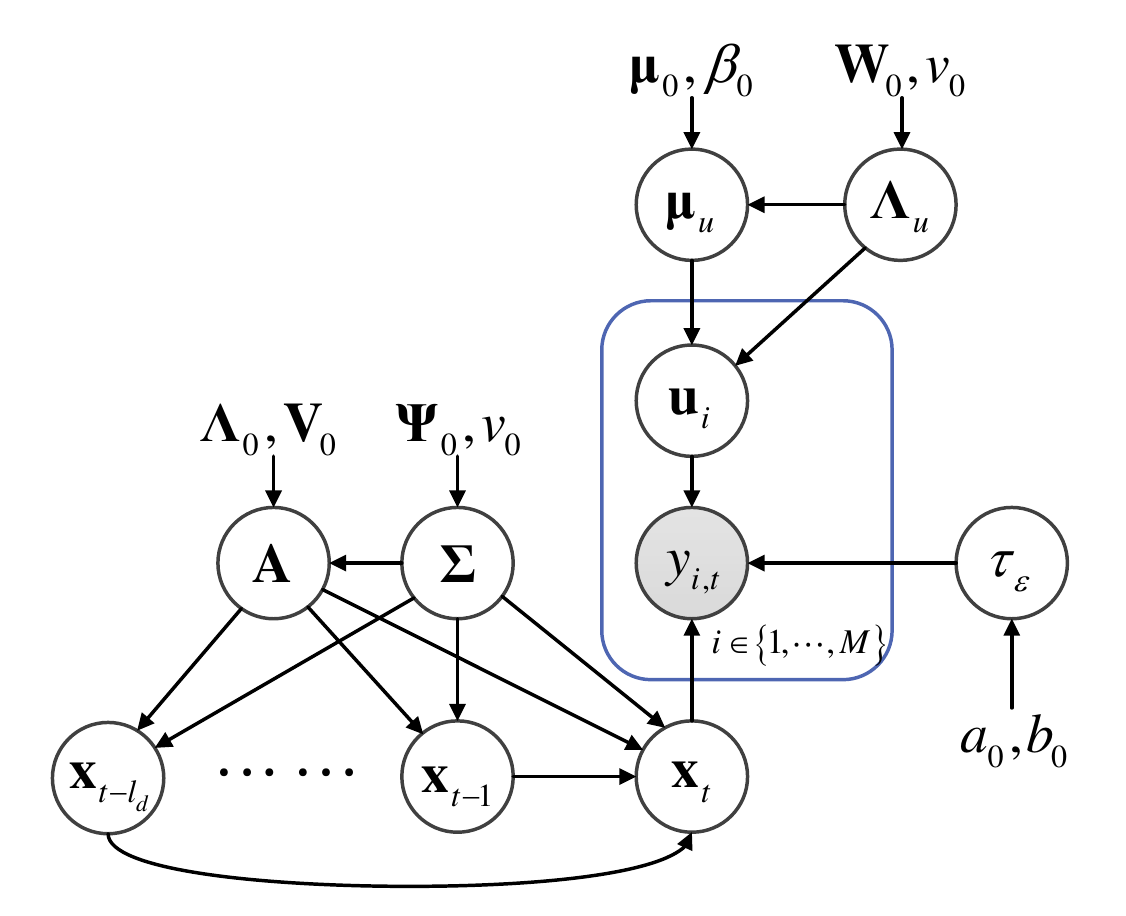}
    \caption{Probabilistic graphic model for Bayesian tensor learning \cite{sun2019bayesian}}
    \label{fig:graphic_model}
\end{figure}

The graphic model representing the generative Bayesian tensor learning described above is depicted in Figure \ref{fig:graphic_model}. The grey node $y_{i,t}$ ($(i,t) \in \Omega$) is the observed SHM data, while $\mathbf{u}_i$, $\mathbf{x}_t$ and $\tau_\epsilon$ are the parameters in the likelihood distribution \ref{y_it}. 
In our experiments, we initialize the scalars as: $\beta_0=1, v_0=K, a_0=b_0=1\times10^{-6}$. The vector $\boldsymbol{\upmu}_0$ and the matrix $\mathbf{\Lambda}_0$ are set as a zeros. The remaining matrices $\{\mathbf{W}_0, \mathbf{V}_0,\boldsymbol{\Psi}_0\}$ are all set to be identity matrix but with different dimensions. 


\subsection{Missing data imputation}\label{S:2.3}
Following the Bayesian modeling formulation in \cite{zhao2015bayesian} and \cite{salakhutdinov2008bayesian}, we infer the predictive distribution over missing entries below:
\begin{equation}
    \label{prediction}
    \begin{split}
    p(y_{i,t}^{*}|\mathbf{Y}, \boldsymbol{\Theta}_0^{u}, \boldsymbol{\Theta}_0^{x}) 
    &= \iint p(y_{i,t}^{*}|\mathbf{u}_i, \mathbf{x}_t, \tau_\epsilon) 
    p(\mathbf{U}, \mathbf{X}, \tau_\epsilon|\mathbf{Y}, \boldsymbol{\Theta}_u, \boldsymbol{\Theta}_x, \boldsymbol{\Theta}_\tau) \\
    & p(\boldsymbol{\Theta}_u, \boldsymbol{\Theta}_x, \boldsymbol{\Theta}_\tau|\boldsymbol{\Theta}_0^{u}, \boldsymbol{\Theta}_0^{x}) \text{d} \{\mathbf{U}, \mathbf{X}, \tau_\epsilon\}
    \text{d} \{\boldsymbol{\Theta}_u, \boldsymbol{\Theta}_x, \boldsymbol{\Theta}_\tau\}
    \end{split}
\end{equation}
where $\boldsymbol{\Theta}_0^{u} = \{\boldsymbol{\upmu}_0, \beta_0, \mathbf{W}_0, v_0\}$ and $\boldsymbol{\Theta}_0^{x} = \{\mathbf{\Lambda}_0, \mathbf{V}_0, \boldsymbol{\Psi}_0, v_0\}$ are assemblies of the hyper-parameters. Noteworthy, the exact solution of Eq. \eref{prediction} cannot be obtained analytically due to the intricate integration over all the latent variables and hyper-parameters. Therefore, we seek to use Markov Chain Monte Carlo (MCMC) sampling \cite{andrieu2010particle} to approximate the inference. The underlying logic of MCMC sampling is that we can draw dependent sequences of samples representing the posterior distribution. Thus, we can describe the predictive distribution in Eq. \eref{prediction} as:
\begin{equation}
    \label{mcmc}
    p(y_{i,t}^{*}|\mathbf{Y}, \boldsymbol{\Theta}_0^{u}, \boldsymbol{\Theta}_0^{x}) \approx 
    \frac{1}{N} \sum_{n=1}^{N} p\left(y_{i,t}^{*}|\mathbf{u}_i^{(n)}, \mathbf{x}_t^{(n)}, \tau_\epsilon^{(n)}\right),
\end{equation}
where $\big\{\mathbf{u}_i^{(n)}, \mathbf{x}_t^{(n)}, \tau_\epsilon^{(n)}\big\}$ denote the $n^{\text{th}}$ simulated sample from the posterior distribution of interest. Herein, we introduce the Gibbs sampling \cite{gilks1992adaptive} to generate the posterior samples, which is a sequential sampling approach by sweeping through each variable to sample from its conditional distribution with the remaining variables fixed to their current values. In addition, thanks to the use of conjugate priors in the Bayesian model generation, we can easily derive the conditional distributions since the posterior distribution is in the same probability distribution family as the prior distribution. The Gibbs sampling procedure for all the parameters and hyper-parameters are described below.
    
\subsubsection{Sampling spatial features}\label{S:2.3.1}
We sample the spatial hyper-parameters $\boldsymbol{\Theta}_u$ first. Considering the likelihood in Eq. \eref{u_i} and the prior in Eq. \eref{mu_u}, the posterior distribution is given by a Gaussian-Wishart distribution \cite{sun2019bayesian}:
\begin{equation}
    \label{sample_mu_u}
    \begin{split}
        p(\boldsymbol{\upmu}_u, \mathbf{\Lambda}_u|\mathbf{U}, \boldsymbol{\Theta}_0^{u}) &= 
        \mathcal{N} (\boldsymbol{\upmu}_u|\boldsymbol{\upmu}_0^{*}, (\beta_0^{*} \mathbf{\Lambda}_u)^{-1}) \mathcal{W} (\mathbf{\Lambda}_u|\mathbf{W}_0^{*}, v_0^{*}) \\
        & \propto \prod_{i=1}^{M} \mathcal{N}(\mathbf{u}_i|\boldsymbol{\upmu}_u, \mathbf{\Lambda}_u^{-1})\times  \mathcal{N}(\boldsymbol{\upmu}_u|\boldsymbol{\upmu}_0, (\beta_0 \mathbf{\Lambda}_u)^{-1}) \times \mathcal{W} (\mathbf{\Lambda}_u|\mathbf{W}_0, v_0)
    \end{split}
\end{equation}
where
\begin{equation}
    \label{sample_mu_u_para}
    \begin{split}
        & \boldsymbol{\upmu}_0^{*} = \frac{\beta_0 \boldsymbol{\upmu}_0 + M \widebar{\mathbf{u}}}
        {\beta_0 + M}, \mbox{  } \beta_0^{*} = \beta_0 + M, \mbox{  } v_0^{*} = v_0 + M, \\
        & (\mathbf{W}_0^{*})^{-1} = \mathbf{W}_0^{-1} + M \widebar{\mathbf{S}} + \frac{\beta_{0}M}{\beta_0 + M} (\boldsymbol{\upmu}_0 - \widebar{\mathbf{u}})(\boldsymbol{\upmu}_0 - \widebar{\mathbf{u}})^\top.
    \end{split}
\end{equation}
Here, $\widebar{\mathbf{u}}$ and $\widebar{\mathbf{S}}$ are two statistical parameters defined as:
\begin{equation}
    \label{stat}
    \widebar{\mathbf{u}} = \frac{1}{M} \sum_{i=1}^{M} \mathbf{u}_i, \mbox{  }
    \widebar{\mathbf{S}} = \frac{1}{M} \sum_{i=1}^{M} (\mathbf{u}_i - \widebar{\mathbf{u}})(\mathbf{u}_i - \widebar{\mathbf{u}})^\top.
\end{equation}
The conditional distribution over spatial features $\mathbf{u}_i$, conditioned on temporal features $\mathbf{X}$, partially observed sensor data $\mathbf{Y}$, precision $\tau_\epsilon$ and all other hyper-parameters of interest can be obtained \cite{sun2019bayesian}:
\begin{equation}
    \label{sample_u_i}
    \begin{split}
            p(\mathbf{u}_i|\mathbf{Y}, \mathbf{X}, \boldsymbol{\Theta}_u, \tau_\epsilon) & = \mathcal{N} (\mathbf{u}_i|\boldsymbol{\upmu}_u^{*},  (\mathbf{\Lambda}_u^{*})^{-1}) \\
            & \propto \prod_{t=1}^{T} \mathcal{N}(y_{i,t}|\mathbf{u}_i^\top \mathbf{x}_t, \tau_\epsilon) \times \mathcal{N}(\mathbf{u}_i|\boldsymbol{\upmu}_u, (\mathbf{\Lambda}_u)^{-1}),
    \end{split}
\end{equation}
where 
\begin{equation}
    \label{sample_u_i_para}
    \begin{split}
        \mathbf{\Lambda}_u^{*} &= \mathbf{\Lambda}_u + \tau_\epsilon \sum_{t=1}^{T} \mathbf{x}_t \mathbf{x}_t^\top, \\
        \mathbf{\upmu}_u^{*} &= (\mathbf{\Lambda}_u^{*})^{-1} \left(\tau_\epsilon \sum_{i=1}^{T} \mathbf{x}_t y_{i,t} + \mathbf{\Lambda}_u \boldsymbol{\upmu}_u \right),
        \mbox{  } (i,t)\in \Omega. 
    \end{split}
\end{equation}

\subsubsection{Sampling temporal features}\label{S:2.3.2}
Following the sampling procedure for spatial features, we infer the conditional distribution of the hyper-paramters $\mathbf{\Theta}_x$ with the likelihood in Eq. \eref{x_t} and the prior in Eq. \eref{A_Sigma}, namely,
\begin{equation}
    \label{sample_A}
    \begin{split}
        p(\mathbf{A}, \boldsymbol{\Sigma}|\mathbf{X}, \boldsymbol{\Theta}_0^{x}) &= \mathcal{MN}(\mathbf{A}|\mathbf{\Lambda}_0^{*}, \mathbf{V}_0^{*}, \boldsymbol{\Sigma}) \mathcal{IW}(\boldsymbol{\Sigma}|\boldsymbol{\Psi}_0^{*}, v_0^{*}) \\
        & \propto \prod_{t=1}^{T} \mathcal{N}(\mathbf{x}_t|\widetilde{\boldsymbol{\upmu}}_x, \widetilde{\boldsymbol{\Sigma}}_x) \times \mathcal{MN} (\mathbf{A}|\mathbf{\Lambda}_0, \mathbf{V}_0, \boldsymbol{\Sigma}) \times \mathcal{IW}(\boldsymbol{\Sigma}|\boldsymbol{\Psi}_0, v_0).
    \end{split}
\end{equation}
Matching the coefficients of the hyper-parameters in Eq. \eref{sample_A}, we can obtain the updated parameters as follows \cite{sun2019bayesian}:
\begin{equation}
    \label{sample_A_para}
    \begin{split}
        \mathbf{V}_0^{*} &= (\mathbf{V}_0^{-1} + \mathbf{Q}^\top \mathbf{Q})^{-1}, \\ \boldsymbol{\Lambda}_0^{*} &= \mathbf{V}_0^{*} (\mathbf{V}_0^{-1} \mathbf{\Lambda}_0 + \mathbf{Q}^\top \mathbf{P}), \\
        v_0^{*} &= v_0 + T - l_d, \\
        \boldsymbol{\Psi}_0^{*} &= \boldsymbol{\Psi}_0 + \mathbf{P}^\top \mathbf{P} + \boldsymbol{\Lambda}_0^\top \mathbf{V}_0^{-1} \boldsymbol{\Lambda}_0 - (\boldsymbol{\Lambda}_0^{*})^\top (\mathbf{V}_0^{*})^{-1} \boldsymbol{\Lambda}_0^{*}.
    \end{split}
\end{equation}
These two matrices $\mathbf{P} \in \mathbb{R}^{(T-l_d) \times K}$ and $\mathbf{Q} \in \mathbb{R}^{(T-d) \times (Kd)}$ are defined for simplicity and convenience, expressed as
\begin{equation}
    \label{P_Q}
    \begin{split}
    \mathbf{P} &= [\mathbf{x}_{l_d+1}^\top, \cdots, \mathbf{x}_{T}^\top]^\top, \\
    \mathbf{Q} &= [\mathbf{z}_{l_d+1}^\top, \cdots, \mathbf{z}_{T}^\top]^\top.
    \end{split}
\end{equation}

After sampling the hyper-parameters, we further derive the conditional distribution of the temporal factor $\mathbf{x}_t$, whose posterior distribution follows Gaussian, given by
\begin{equation}
    \label{sample_x_t}
    \begin{split}
        p(\mathbf{x}_t|\mathbf{Y}, \mathbf{U}, \boldsymbol{\Theta}_x, \tau_\epsilon) &= \mathcal{N} (\mathbf{x}_t|\boldsymbol{\upmu}_x^{*}, \boldsymbol{\Sigma}_x^{*}) \\
        &\propto \prod_{i=1}^{M} \mathcal{N}(y_{i,t}|\mathbf{u}_i^\top \mathbf{x}_t, \tau_\epsilon) \times \mathcal{N}(\mathbf{x}_t|\widetilde{\boldsymbol{\upmu}}_x, \widetilde{\boldsymbol{\Sigma}}_x).
    \end{split}
\end{equation}
Nevertheless, sampling $\mathbf{x}_t$ is complicated due to the piecewise Bayesian modeling on the temporal feature parameters. Here, we introduce four auxiliary variables $\{\mathbf{C}, \mathbf{D}, \mathbf{E}, \mathbf{F}\}$ considering the function of the AR process. The general updating formulation can thus be written as \cite{sun2019bayesian}
\begin{equation}
    \label{sample_x_t_para}
    \begin{split}
        \boldsymbol{\Sigma}_x^{*} &= \left(\tau_\epsilon \sum_{i=1}^{M} \mathbf{u}_i \mathbf{u}_i^\top + \mathbf{C} + \mathbf{D} \right)^{-1}, \\
        \boldsymbol{\upmu}_x^{*} &= \boldsymbol{\Sigma}_x^{*} \left(\tau_\epsilon \sum_{i=1}^{M} \mathbf{u}_i y_{i,t} + \mathbf{E} + \mathbf{F} \right), \mbox{ } (i,t) \in \Omega.
    \end{split}
\end{equation}
where the variables $\mathbf{C}$ and $\mathbf{E}$ are given by
\begin{equation}
    \label{C}
    \mathbf{C} = 
    \begin{cases}
        \sum_{j=1, l_d < t+l_j \leq T}^{d}\mathbf{A}_j^\top \boldsymbol{\Sigma}^{-1} \mathbf{A}_j, & \mbox{if } t \in \{1,2, \cdots, T-l_1\}, \\
        \mathbf{0}, & \mbox{otherwise},
    \end{cases}
\end{equation}
\begin{equation}
    \label{E}
    \mathbf{E} = 
    \begin{cases}
        \sum_{j=1, l_d < t+l_j \leq T}^{d}\mathbf{A}_j^\top \boldsymbol{\Sigma}^{-1} \boldsymbol{\phi}_{t+l_j}, & \mbox{if } t \in \{1,2, \cdots, T-l_1\}, \\
        \mathbf{0}, & \mbox{otherwise},
    \end{cases}
\end{equation}
with $\boldsymbol{\phi}_{t+l_j}$ being defined as
\begin{equation}
    \label{phi}
    \boldsymbol{\phi}_{t+l_j} = \mathbf{x}_{t+l_j} - \sum_{p=1, p \neq j}^{d} \mathbf{A}_p \mathbf{x}_{t+l_j-l_p}.
\end{equation}
In addition, the variables $\mathbf{D}$ and $\mathbf{F}$ can be written as
\begin{equation}
    \label{D}
    \mathbf{D} = 
    \begin{cases}
        \mathbf{I}_x, & \mbox{if } t \in \{1,2, \cdots, l_d\}, \\
        \boldsymbol{\Sigma}^{-1}, & \mbox{otherwise},
    \end{cases}
\end{equation}
\begin{equation}
    \mathbf{F} = 
    \begin{cases}
        \mathbf{0}, & \mbox{if } t \in \{1,2, \cdots, l_d\}, \\
        (\boldsymbol{\Sigma}^{-1}) \sum_{p=1}^{d} \mathbf{A}_p \mathbf{x}_{t-l_p}, & \mbox{otherwise}.
    \end{cases}  
\end{equation}

\subsubsection{Sampling precision}\label{S:2.3.3}
With the combination of the likelihood in Eq. \eref{y_it} and the prior in Eq. \eref{gamma}, the posterior distribution of precision $\tau_\epsilon$ can be represented by a Gamma distribution \cite{sun2019bayesian}, namely,
\begin{equation}
    \label{sample_gamma}
    \begin{split}
        p(\tau_\epsilon|\mathbf{Y}, \mathbf{U}, \mathbf{X}, \boldsymbol{\Theta}_\tau) &= \text{Gamma} (a_0^{*}, b_0^{*}) \\
        & \propto \prod_{i=1}^{M} \prod_{t=1}^{T} \mathcal{N} (y_{i,t}|\mathbf{u}_i^\top \mathbf{x}_t, \tau_\epsilon) \times \text{Gamma}(\tau_\epsilon|a_0, b_0),
    \end{split}
\end{equation}
where the hyper-parameters $a_0^{*}$ and $b_0^{*}$ can be expressed as
\begin{equation}
    \label{a_0_star}
    \begin{split}
    a_0^{*} &= \frac{1}{2} \sum_{(i,t) \in \Omega} s_{i,t} + a_0, \\
    b_0^{*} &= \frac{1}{2} \sum_{(i,t) \in \Omega} (y_{i,t} - \mathbf{u}_i^\top \mathbf{x}_t)^{2} + b_0.
    \end{split}
\end{equation}
Note that $s_{i,t}$ is 1 if $(i,t) \in \Omega$ and 0 otherwise.

\subsection{Structural response forecasting}\label{S:2.4}

We predict the future structural response $y_{i,t+1}$ based on both incrementally updated spatial feature $\mathbf{U}$ (see Figure \ref{fig:batch_window}) and temporal feature $\mathbf{x}_{t+1}$ (see Figure \ref{fig:MF}), but set a periodical updating constraint on forecasting the spatial attribute considering computational efficiency. Namely, after getting well-trained parameters from the imputation process, we keep $\{\mathbf{X}, \mathbf{A}\}$ unchanged for the forecasting step and only view $\{\mathbf{U}, \boldsymbol{\Sigma}, \mathbf{x}_{t+1}, \tau_\epsilon\}$ as the updated targets. Moreover, to predict $y_{i,t+2}$ sequentially, we provide the observed $y_{i,t+1}$ as an input and conduct the above procedure iteratively. The general philosophy of Bayesian forecasting can be illustrated by two steps as follows.

The first step is to learn $\{\mathbf{U}, \boldsymbol{\Sigma}, \mathbf{x}_{t}, \tau_\epsilon\}$ from the historical observation $\mathbf{y}_{:,t}$. The model generative formulations are expressed as:
\begin{equation}
    \label{eqn:forecast}
    \begin{split}
        y_{i,t} &\sim \mathcal{N}(\mathbf{u}_i^\top \mathbf{x}_{t}, \tau_\epsilon^{-1}),\\
        \mathbf{x}_{t} &\sim \mathcal{N}({\widetilde{\boldsymbol{\upmu}}_x}, \widetilde{\boldsymbol{\Sigma}}_x),\\
        \widetilde{\boldsymbol{\Sigma}}_x &\sim \mathcal{IW}(\boldsymbol{\Psi}_0, v_0),\\
        \tau_\epsilon &\sim \text{Gamma}(a_0, b_0)
    \end{split}
\end{equation}
\noindent where $\widetilde{\boldsymbol{\upmu}}_x$ is a known parameter denoted as $\mathbf{A}^\top \mathbf{z}_{t}$. Note that there is possible missing values in $\mathbf{y}_{:,t}$. The model inference using Gibbs sampling for this step is divided into three parts. To begin with, if $t$ is at the end of batch window, we need to sample the spatial factor and its hyper-paramters referring to Eq. \eref{sample_mu_u} and \eref{sample_u_i}. Secondly, for the temporal feature, we do sampling on the hyper-parameter $(\widetilde{\boldsymbol{\Sigma}}_x)^{-1} \sim \mathcal{W}((\boldsymbol{\Psi}_0^{*})^{-1}, v_0+1)$ where
\begin{equation}
    \label{eqn:forecast_sigma}
    \boldsymbol{\Psi}_0^{*} = \boldsymbol{\Psi}_0 + (\mathbf{x}_t - \mathbf{A}^\top \mathbf{z}_t)(\mathbf{x}_t - \mathbf{A}^\top \mathbf{z}_t)^\top.
\end{equation}
Then we sample the future temporal factor $\mathbf{x}_t \sim \mathcal{N}((\widetilde{\boldsymbol{\upmu}}_x)^{*}, (\widetilde{\boldsymbol{\Sigma}}_x)^{*})$ with 
\begin{equation}
    \label{eqn:forecast_xt}
    \begin{split}
        \widetilde{\boldsymbol{\Sigma}}_x^{*} &= \left(\tau_\epsilon \sum_{i=1}^{M} \mathbf{u}_i \mathbf{u}_i^\top + \widetilde{\boldsymbol{\Sigma}}_x^{-1} \right)^{-1},\\
        \widetilde{\boldsymbol{\upmu}}_x^{*} &= \widetilde{\boldsymbol{\Sigma}}_x^{*} \left( \tau_\epsilon \sum_{i=1}^{M} \mathbf{u}_i y_{i,t} +  \widetilde{\boldsymbol{\Sigma}}_x^{-1} \mathbf{A}^\top \mathbf{z}_t \right).
    \end{split}
\end{equation}
The third part is the sampling of the precision parameter $\tau_\epsilon$ using Eq. \eref{a_0_star}.

\begin{algorithm}[t!]
\small
\SetAlgoLined
    \KwIn{the SHM data tensor $\mathbf{Y}$, the indicator tensor $\mathbf{\Omega}$, the chain length for imputation $N_1^\text{mc}$, the burn-in period for imputation $N_1^\text{b}$, the chain length for forecasting $N_2^\text{mc}$, the burn-in period for forecasting $N_2^\text{b}$, tensor rank $K$, time lags $\mathcal{L}$, forecasting length $\mathcal{T}$, forward batch length $I$ and critical point $T_1$.}
    \KwOut{the chains of samples for the total imputed tensor $\widehat{\mathbf{Y}}_{1}$ and the forecasted tensor $\widehat{\mathbf{Y}}_{2}$.}
    {\bf Initialize:} $\mathbf{U}$\;
    $N_{\text{s}} = T_1 / I$, $N_{\text{total}} = T_\text{total} / I$\;
    Define a counting tensor for averaging imputations $\mathbf{C} \gets \mathbf{0}$\;
    \For{\upshape $w=1,\cdots,N_\text{total}$}
    {
        \tcp{Short time dynamic batch window}
        \If{$w \leq N_{\text{s}}$}
        {
            $\mathbf{U}^{(w)} \gets$ updated $\mathbf{U}$\;
            $\mathbf{X}^{(w)} \gets$ randomly initialized $\mathbf{X}$\;
            $\mathbf{U}, \widehat{\mathbf{Y}}_{1}^{(w)}, \widehat{\mathbf{Y}}_{2}^{(w)} \gets$ impute and forecast with $\mathbf{U}^{(w)}, \mathbf{X}^{(w)}$ in  $[0,wI]$ (Algorithm \ref{alg:gibbs})\;
            $\mathbf{C}^{(w)} \gets \mathbf{1}$\;
        }
        \tcp{Long-term fixed batch window}
        \Else
        {
            $\mathbf{U}^{(w)} \gets$ updated $\mathbf{U}$\;
            $\mathbf{X}^{(w)} \gets$ updated $\mathbf{X}$\;            
            $\mathbf{U}, \mathbf{X}, \widehat{\mathbf{Y}}_{1}^{(w)}, \widehat{\mathbf{Y}}_{2}^{(w)} \gets$ impute and forecast with $\mathbf{U}^{(w)}, \mathbf{X}^{(w)}$ in $[(w-N_\text{s})I,wI]$ (Algorithm \ref{alg:gibbs})\;
            $\mathbf{C}^{(w)} \gets \mathbf{1}$\;
        }
    }
    Collect and average the imputations $\widehat{\mathbf{Y}}_{1} \gets$ sum($\widehat{\mathbf{Y}}_{1}(1:N_\text{total})$)/sum(${\mathbf{C}}(1:N_\text{total})$)\;
    Collect the forecasting $\widehat{\mathbf{Y}}_{2} \gets$ sum($\widehat{\mathbf{Y}}_{2}(1:N_\text{total})$)\;
    \caption{Incremental tensor learning}
    \label{alg:incre}
\end{algorithm}
\normalsize

\begin{algorithm}[t!]
\small
\SetAlgoLined
    \KwIn{the SHM data tensor $\mathbf{Y}^{(w)}$, the indicator tensor $\mathbf{\Omega}^{(w)}$, the chain length for imputation $N_1^\text{mc}$, the burn-in period for imputation $N_1^\text{b}$, the chain length for forecasting $N_2^\text{mc}$, the burn-in period for forecasting $N_2^\text{b}$, tensor rank $K$, time lags $\mathcal{L}$ and forecasting length $\mathcal{T}$.}
    \KwOut{updated $\mathbf{U}$, updated $\mathbf{X}$, the chains of samples for the estimated tensor $\widehat{\mathbf{Y}}_{1}^{(w)}$ and the predicted tensor $\widehat{\mathbf{Y}}_{2}^{(w)}$.}
    {\bf Initialize:} $\mathbf{U}^{(w)}$, $\mathbf{X}^{(w)}$, $\mathbf{A}$, $\boldsymbol{\Theta}_0^{u}$, $\boldsymbol{\Theta}_0^{x}$ and $\boldsymbol{\Theta}_{\tau}$\;
    \tcp{The imputation process}
    \For{\upshape $n_{1}=1,\cdots,N_1^\text{mc}$}
    {
        Sample the hyperparameter $\boldsymbol{\Theta}_u$ (Eq. \ref{sample_mu_u})\;
        $\boldsymbol{\Theta}_u \sim p(\boldsymbol{\Theta}_u|\mathbf{U}^{(w)}, \boldsymbol{\Theta}_0^{u})$\;
        \For{$i=1,\cdots,M$}
        {
            Sample the spatial feature $\mathbf{u}_i$ (Eq. \ref{sample_u_i})\;
            $\mathbf{u}_i \sim p(\mathbf{u}_i|\mathbf{Y}^{(w)}, \mathbf{X}^{(w)}, \boldsymbol{\Theta}_u, \tau_\epsilon)$\;
        }
        Sample the hyperparameter $\boldsymbol{\Theta}_x$ (Eq. \ref{sample_A})\;
        $\boldsymbol{\Theta}_x \sim p(\boldsymbol{\Theta}_x|\mathbf{X}^{(w)}, \boldsymbol{\Theta}_0^{x})$\;
        \For{$t=1,\cdots,T$}
        {
            Sample the temporal feature $\mathbf{x}_t$ (Eq. \ref{sample_x_t})\;
            $\mathbf{x}_t \sim p(\mathbf{x}_t|\mathbf{Y}^{(w)}, \mathbf{U}^{(w)}, \boldsymbol{\Theta}_x, \tau_\epsilon)$\;
        } 
        Sample the precision parameter $\tau_\epsilon$ (Eq. \ref{sample_gamma})\;
        $\tau_\epsilon \sim p(\tau_\epsilon|\mathbf{Y}^{(w)}, \mathbf{U}^{(w)}, \mathbf{X}^{(w)}, \boldsymbol{\Theta}_\tau)$\;
        \If{\upshape $n_{1} \ge N_1^\text{b}$}
        {
            Compute and collect the sample $\widehat{\mathbf{Y}}_{1}^{(w)} = [\mathbf{U}^{(w)}]^\top \mathbf{X}^{(w)}$\;
        }
    }
    \tcp{The forecasting process}
    \For{\upshape $s=1,\cdots,\mathcal{T}$}
    {
        \For{\upshape $n_{2}=1,\cdots,N_2^\text{mc}$}
        {
            \If{\upshape $s \text{ mod } I==0$}
            {
                Sample the hyperparameter $\boldsymbol{\Theta}_u$ (Eq. \ref{sample_mu_u})\;
                Sample the spatial feature $\mathbf{U}$ (Eq. \ref{sample_u_i})\;
            }
            Sample the hyperparameter $\widetilde{\boldsymbol{\Sigma}}_x$ (Eq. \ref{eqn:forecast_sigma})\;
            Sample the temporal feature $\mathbf{x}_t$ (Eq. \ref{eqn:forecast_xt})\;
            Sample the precision parameter $\tau_\epsilon$ (Eq. \ref{sample_gamma})\;
            \If{\upshape $n_{2} \ge N_2^\text{b}$}
            {
                Compute and collect the sample $\widehat{\mathbf{Y}}_{2} = [\mathbf{U}^{(w)}]^\top (\mathbf{A}\mathbf{z}_{t+1})$\; 
            }
        }    
    }
    \caption{Gibbs sampling for Bayesian tensor learning}
    \label{alg:gibbs}
\end{algorithm}
\normalsize

After generating samples of $\mathbf{x}_t$, the second step for forecasting is that we run Gibbs sampling on the prediction for multiple iterations based on $\mathbf{y}_{:,t+1} \approx \mathbf{U}^\top (\mathbf{A}^\top \mathbf{z}_{t+1})$, and get the average of these samples in the burn-in period as output. This is an efficient strategy for forecasting, especially for large-scale problems. The pseudo code for the proposed incremental Bayesian tensor learning for missing SHM data imputation and structural response forecasting is summarized in  Algorithm \ref{alg:incre} and Algorithm \ref{alg:gibbs}.

\section{Experimental Validation}
    \label{S:3}
In this section, we test the imputation and forecasting performance of the proposed Bayesian tensor learning method under data missing scenarios, using long-term field-monitoring data of a concrete bridge (e.g., strain and temperature records). In particular, we impute and forecast the strain time histories of the bridge. Inspired by the strong correlation between strain and temperature, we formulate the tensor data structure by combining both strain and temperature along the sensor dimension. We also conduct a series of analyses of uncertainty quantification and rank selection for tensor factorization. The numerical analyses are performed on a standard PC with 28 Intel Core i9-7940X CPUs and 2 NVIDIA GTX 1080 Ti GPU.

\begin{figure}[b!]
	\centering
	    \subfigure[Elevation view]{\includegraphics[width=0.42\linewidth]{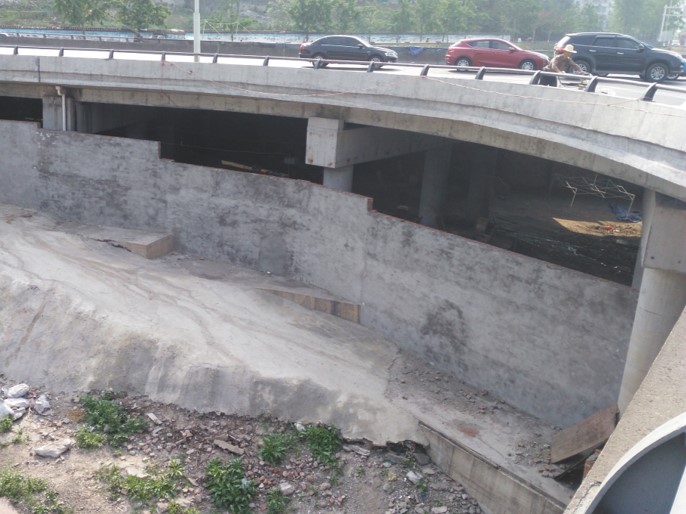}
	        \label{fig:elevation}} 
	    \hspace{1em}
	    \subfigure[Vibrational chord strain gauge]{\includegraphics[width=0.42\linewidth]{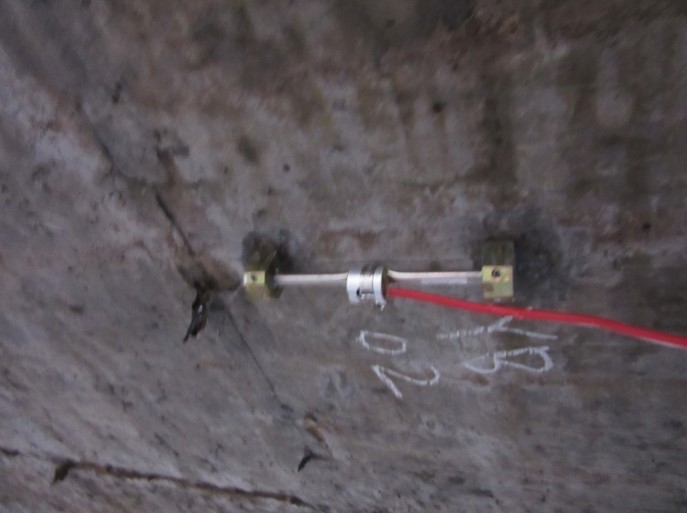}
	        \label{fig:strain_sensor}} 
	    \subfigure[Monitoring sections of strain sensors]{\includegraphics[width=0.42\linewidth]{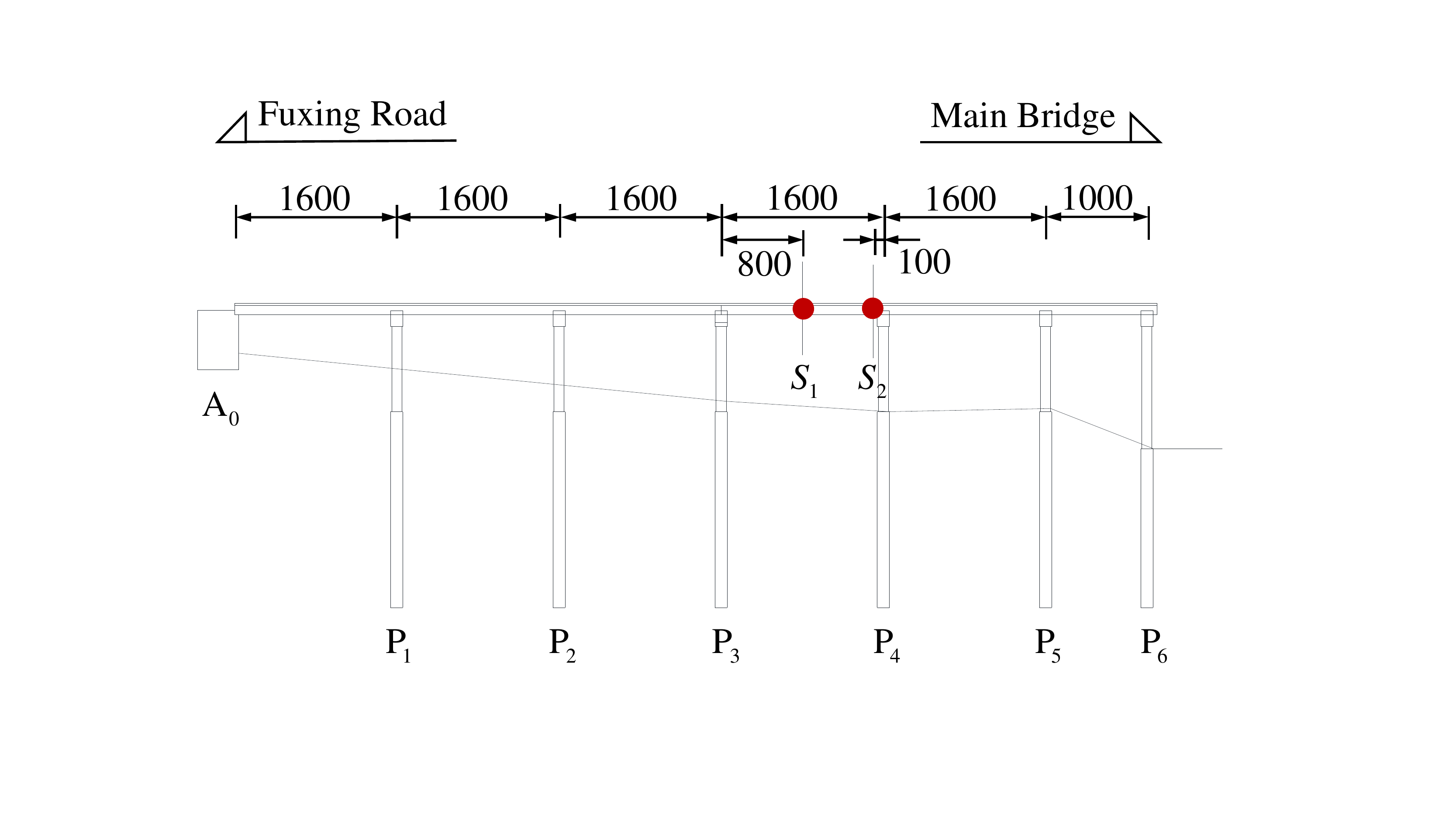}
	        \label{fig:strain_section}} 
	    \hspace{1em}
	    \subfigure[Strain sensor locations at a typical section]{\includegraphics[width=0.42\linewidth]{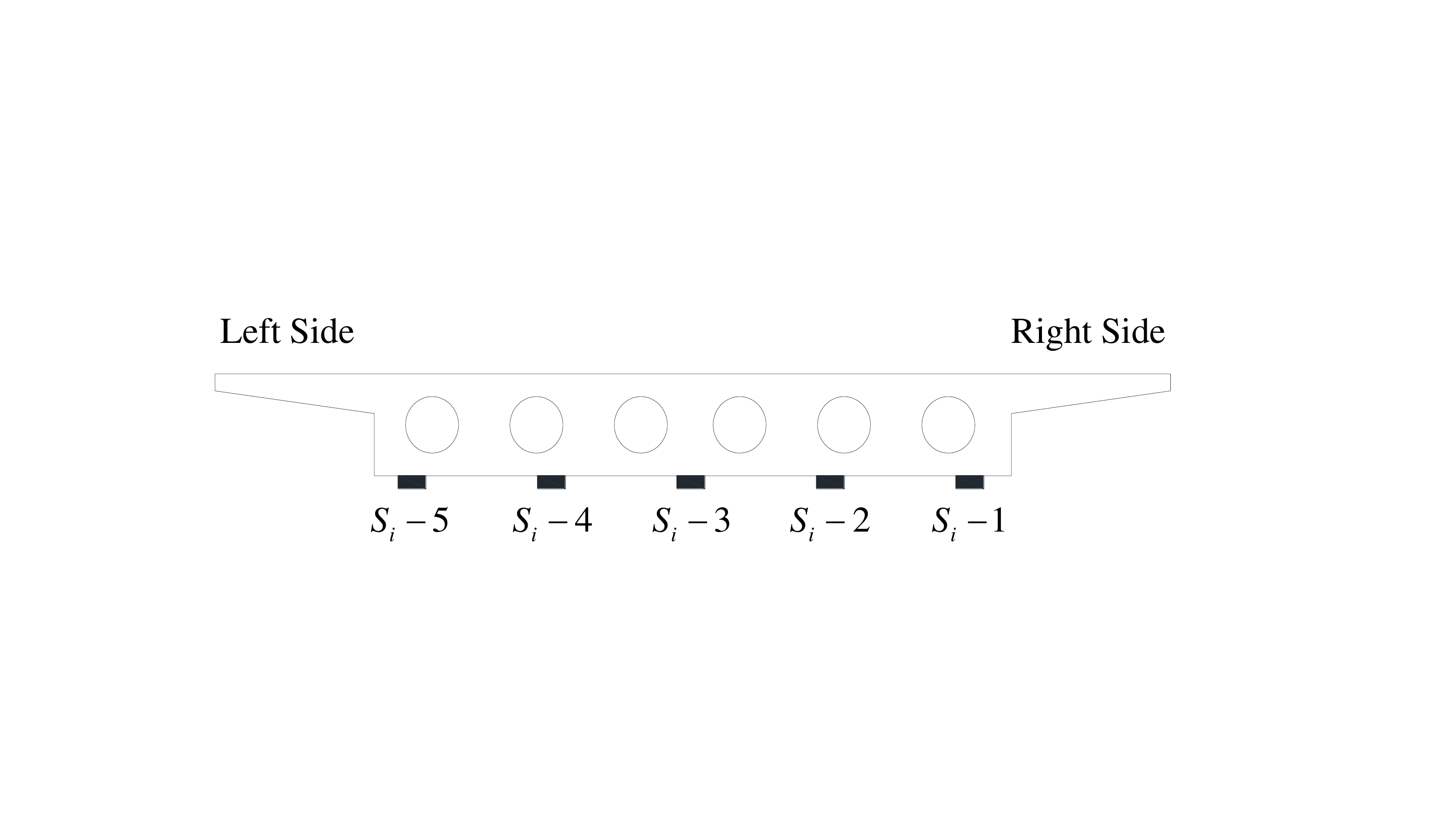} 
	        \label{fig:strain_station}}	        
	\vspace{0pt}
	\caption{The instrumented concrete bridge.}
	\label{fig:Fuxing_bridge}
\end{figure}

\subsection{Bridge Description}\label{S:3.1}
The instrumented concrete bridge (see Figure \ref{fig:Fuxing_bridge}\subref{fig:elevation}) considered herein is a connection bridge located in the old section of Wanzhou District, Chongqing, China. It has the total length of 94.015 m, whose span composition is $5  \times 16 \text{~m} + 10 \text{~m}$ (see Figure \ref{fig:Fuxing_bridge}\subref{fig:strain_section}). The superstructure of this bridge consists of continuous hollow slab beams constructed of reinforced concrete. As shown in Figure \ref{fig:Fuxing_bridge}\subref{fig:strain_section}, we name the bridge abutment $\text{A}_0$ and the piers $\{\text{P}_1, \text{P}_2, \text{P}_3, \text{P}_4, \text{P}_5, \text{P}_6\}$ orderly according to the vehicle moving direction to the main bridge. Two sections are monitored, marked as $\text{S}_1$ and $\text{S}_2$ in the mid-span of the fourth span and near the top of pier $\text{P}_4$, respectively. Figure \ref{fig:Fuxing_bridge}\subref{fig:strain_station} shows that each monitoring section $\text{S}_i (i \in \{1,2\})$ has five strain sensors installed on the bottom of the hollow slab beam. Vibrational chord strain gauges are installed which facilitate monitoring of both strain response of the bridge and the corresponding operation temperature (see Figure \ref{fig:Fuxing_bridge}\subref{fig:strain_sensor}).

\begin{figure}[b!]
	\centering
	    \subfigure[Measured strain data]{\includegraphics[width=0.85\linewidth]{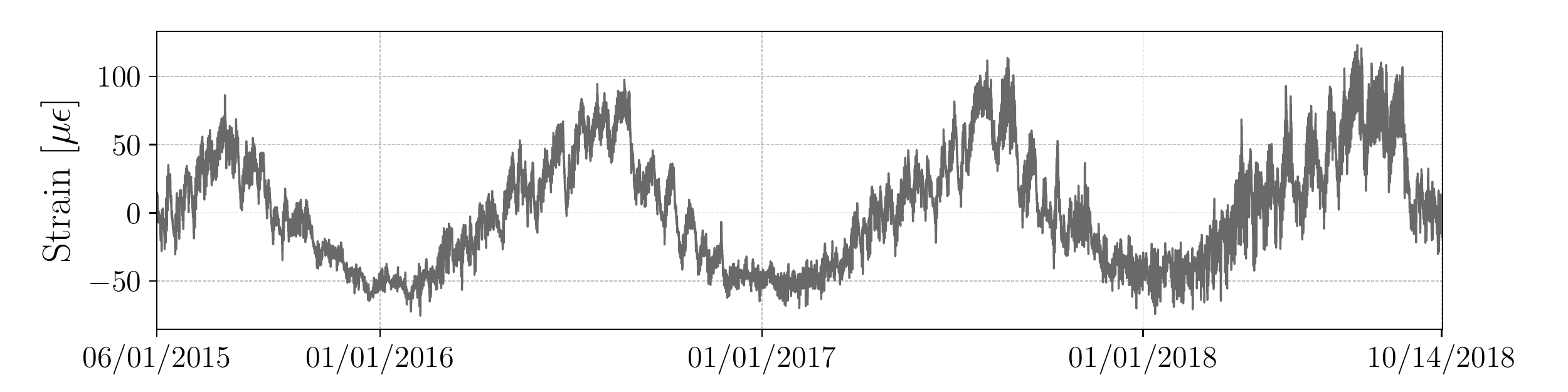}
	    \label{fig:strain}} 
	    \subfigure[Measured temperature data]{\includegraphics[width=0.85\linewidth]{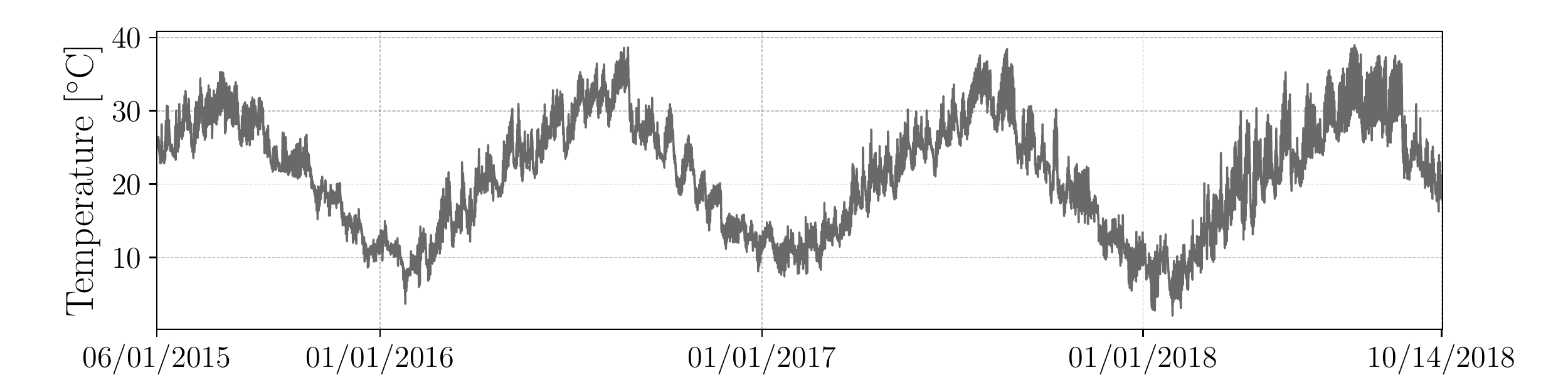}
	    \label{fig:temp}}
	\vspace{0pt}
	\caption{Time series of measurements at sensor $\text{S}_{1}\text{-}1$}
	\label{fig:observations} 
\end{figure}

The dataset collected from the above SHM system contains strain and temperature time histories recorded from June 1, 2015 to October 14, 2018. We resample the data at the rate of 10 min interval. Thus, it can be organized as a two-dimensional tensor with both strain data and temperature data (with a size of 20 $\times$ 177,408, representing sensors $\times$ time stamps). The salient feature behind this data arrangement is that the tensor structure with both strain and temperature can capture a lower rank compared to the tensor structure with only strain data, due to the strong correlation between strain and temperature. Figure \ref{fig:observations} illustrates the recorded strain and temperature time series for over three years from a typical sensor (e.g., $\text{S}_{1}\text{-}1$ as shown in Figure \ref{fig:strain_section} and \ref{fig:strain_station}).

\subsection{Scenarios of Missing Data}\label{S:3.2}
First of all, to evaluate the proposed model for imputation and forecasting, we only set data missing on the strain recordings while keeping the temperature data fixed/known. Namely, the missing rate $\eta$ (e.g., 20\%) is introduced for the strain data in the validation experiments, which is computed as the ratio of the amount of missing data to the total amount of measurements. Secondly, to simulate the real-world missing conditions during monitoring period, we define three primary missing scenarios for the two-dimensional tensor data considered herein, following a similar experimental design procedure for higher-dimensional tensors discussed in \cite{chen2018spatial}. The first scenario is called ``random missing'' (RM) which presents discrete and arbitrary lack of data in the time histories. Each strain entry in the data matrix is dropped randomly (e.g., following a uniform random distribution). The second scenario is termed as ``structured missing'' (SM) where there is data missing occurs continuously for certain periods (e.g., one day or consecutive days). It is a common scenario in practical SHM applications due to sensor malfunctioning, but more challenging and less investigated in literature. In particular, we structurally remove the strain data by selecting multiple days randomly and dropping the corresponding data to simulate a practical missing condition. The last scenario is named ``mixed missing'' (MM) which combines random missing and structured missing at different rates.

After setting the different missing scenarios, we define a sparse binary matrix $\mathbf{B} \in \mathbb{R}^{M \times T} (b_{i,t}=1 \text{ if } (i,t) \in \Omega \text{ and 0 otherwise})$ to record the missing positions for the subsequent comparison between imputation results and the ground truth. The target dense tensor without data missing is named $\mathbf{Y}_{\text{d}}$, and the partially observed tensor $\mathbf{Y}$ can be calculated element-wisely by $\mathbf{B} \odot \mathbf{Y}_{\text{d}}$, where $\odot$ denotes the Hadamard product. The imputation/forecasting accuracy $\rho$ is defined as the root mean square error (RMSE) between the reconstructed/predicted data and the corresponding ground truth, normalized by the root mean square (RMS) of the target values:
\begin{equation}
    \label{accuracy}
    \rho = \left( 1 - \frac{\sqrt{\frac{1}{n} \sum_{i=1}^{n}(y_i-y_i^{*})^2}}{\sqrt{\frac{1}{n} \sum_{i=1}^{n}y_i^2}} \right) \times 100\%.
\end{equation}
where $y_i$ and $y_i^{*}$ denote the ground truth and the estimated value at the same missing position $i$, and $n$ is the total number of missing entries.

\begin{figure}[t!]
	\centering
	    \subfigure[Random missing scenario]{\includegraphics[width=0.75\linewidth]{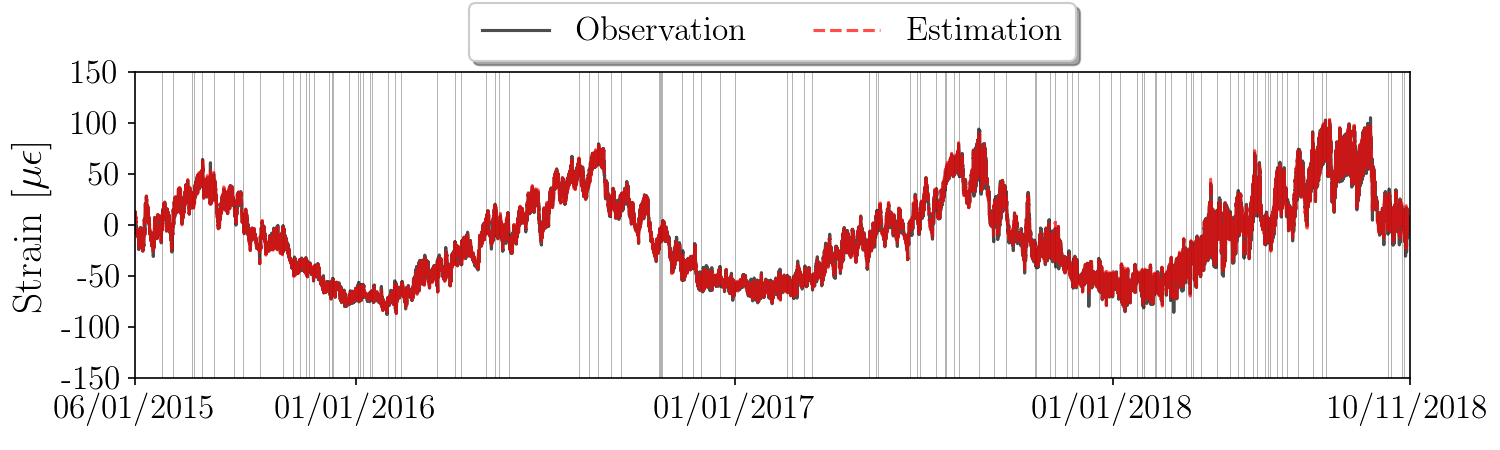}
        \label{fig:rand_impu}} 
	   	\subfigure[Structured missing scenario]{\includegraphics[width=0.75\linewidth]{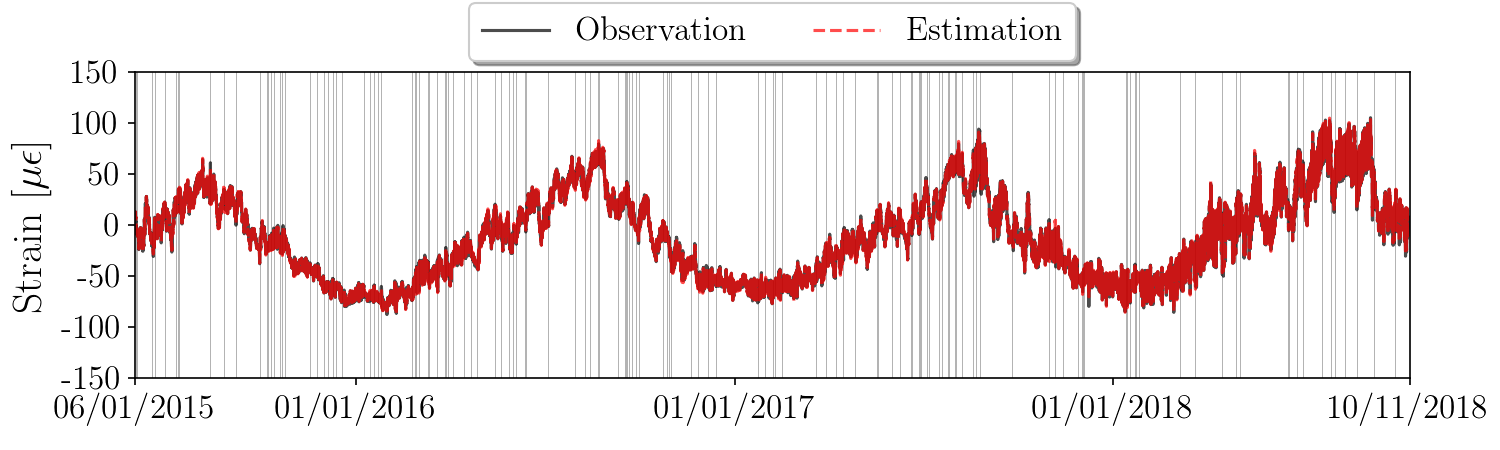}
        \label{fig:fiber_impu}} 
	 	\subfigure[Mixed missing scenario (Case 1)]{\includegraphics[width=0.75\linewidth]{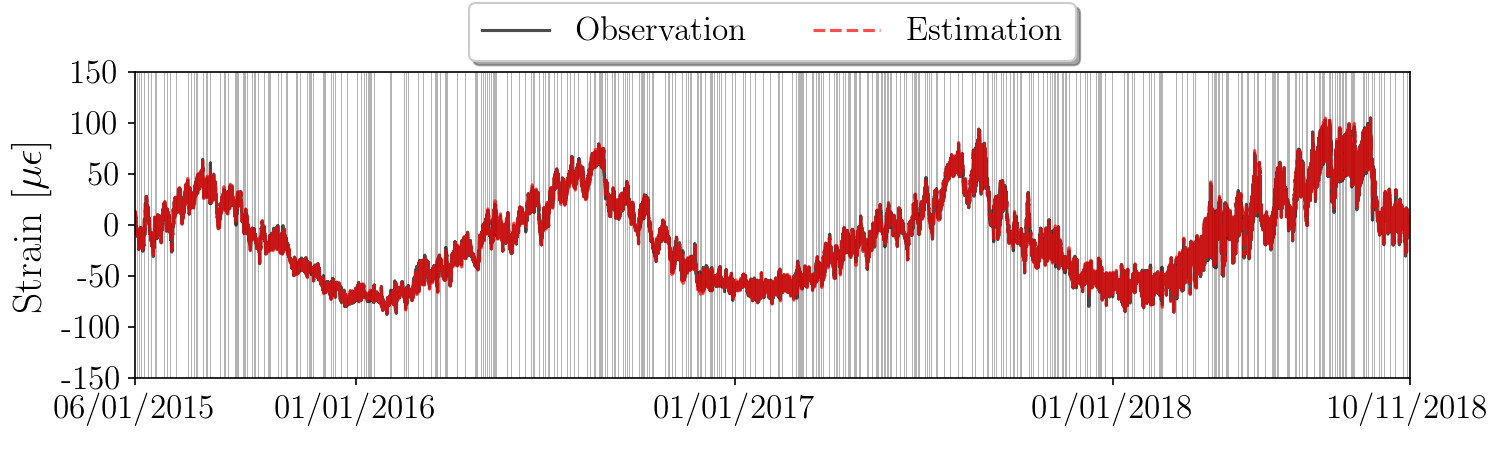}
        \label{fig:mix1_impu}} 
	   	\subfigure[Mixed missing scenario (Case 2)]{\includegraphics[width=0.75\linewidth]{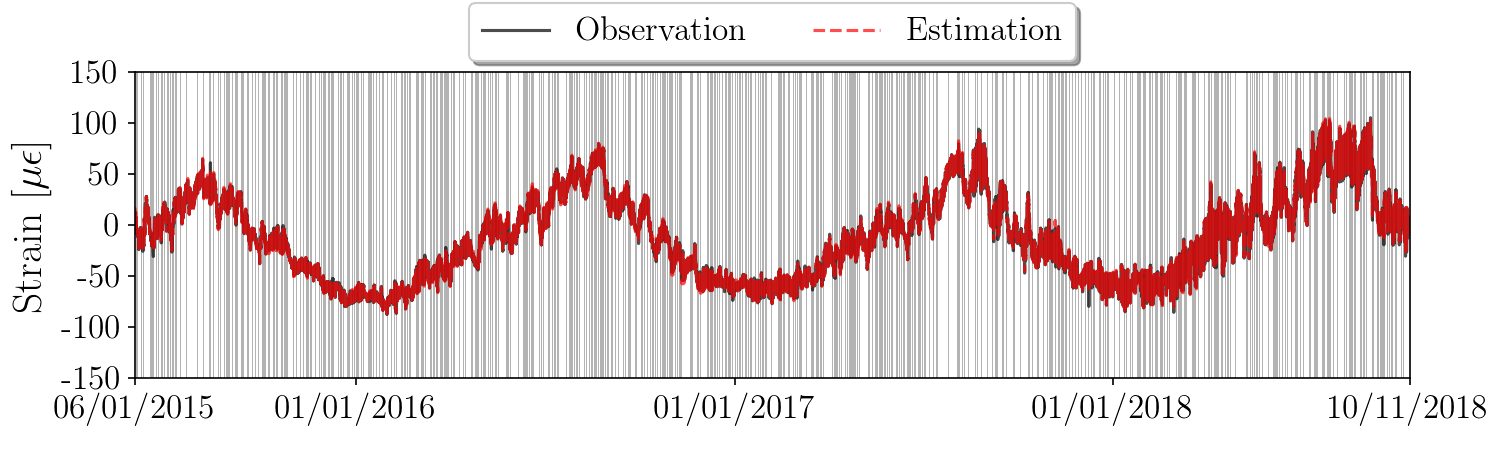}
        \label{fig:mix2_impu}} 
	\vspace{0pt}
	\caption{The imputation result for four missing cases of Sensor  $\text{S}_{2}\text{-}4$. Note that the shading areas represent the time periods where data missing occurs, while the white box areas denote that the strain time series are successfully recorded. The dataset ranges from June 1, 2015 to October 11, 2018 including 41 months.}
	\label{fig:impu_eval}
\end{figure}

\begin{figure}[t!]
	\centering
	    \subfigure[Random missing scenario]{\includegraphics[width=0.75\linewidth]{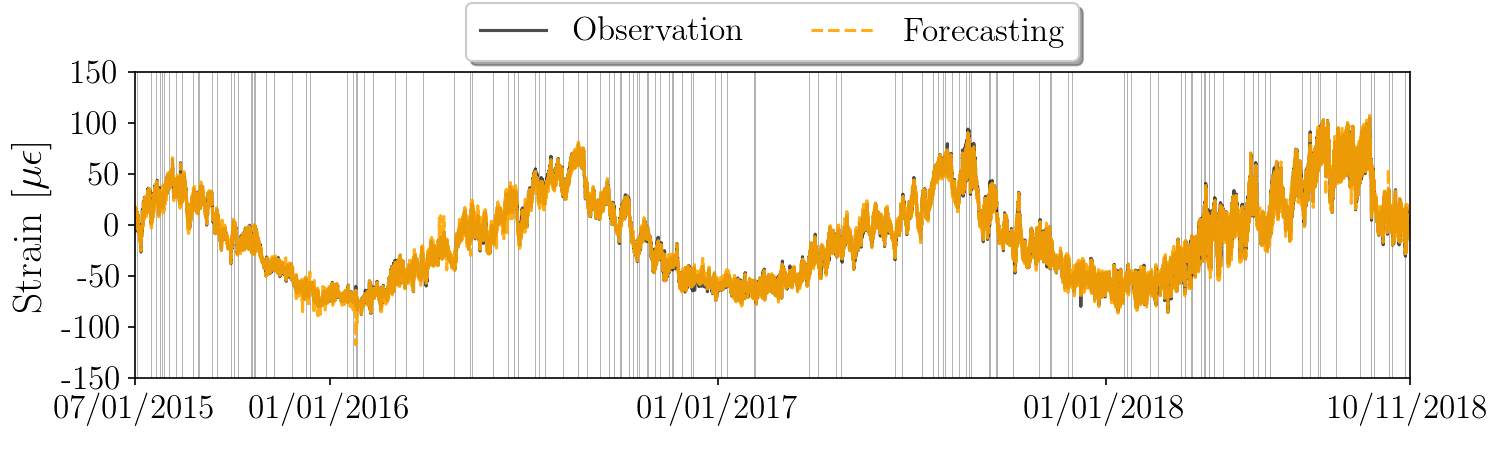}
        \label{fig:rand_pred}} 
	   	\subfigure[Structured missing scenario]{\includegraphics[width=0.75\linewidth]{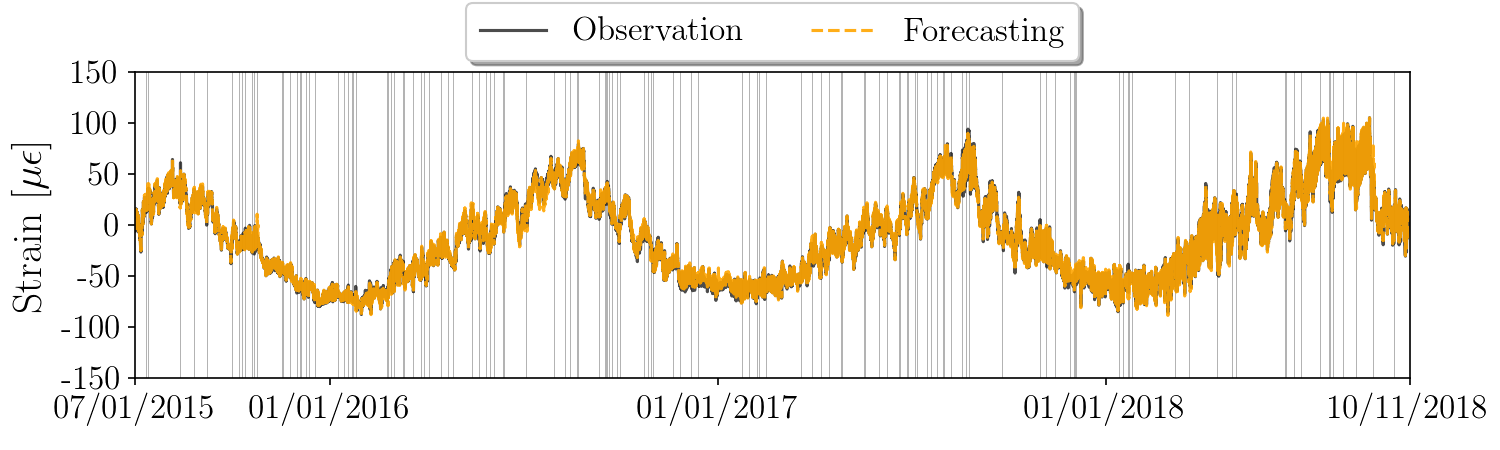}
        \label{fig:fiber_pred}} 
	 	\subfigure[Mixed missing scenario (Case 1)]{\includegraphics[width=0.75\linewidth]{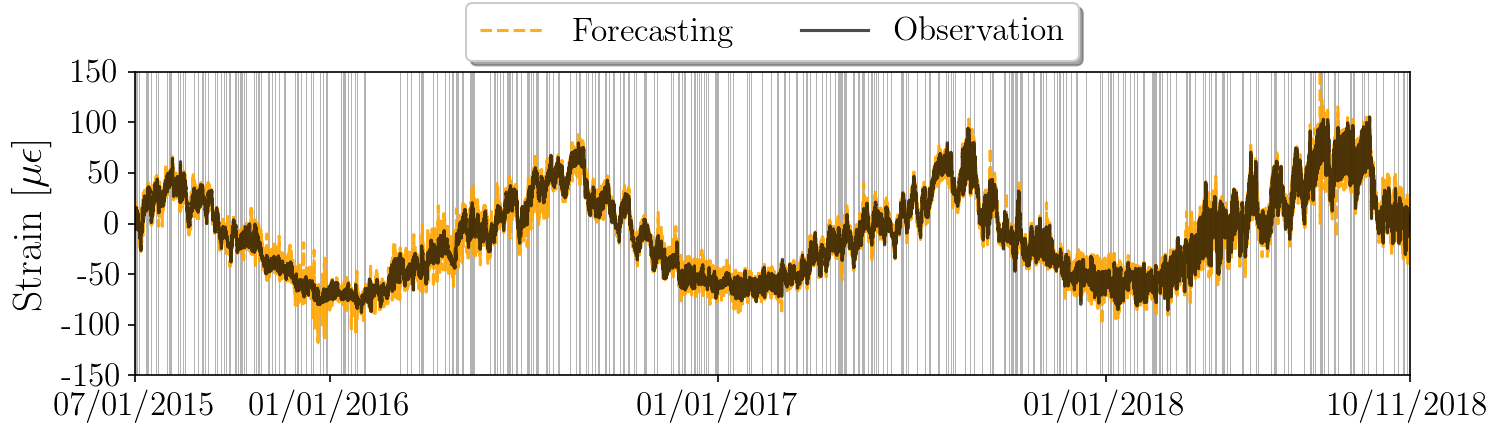}
        \label{fig:mix1_pred}} 
	   	\subfigure[Mixed missing scenario (Case 2)]{\includegraphics[width=0.75\linewidth]{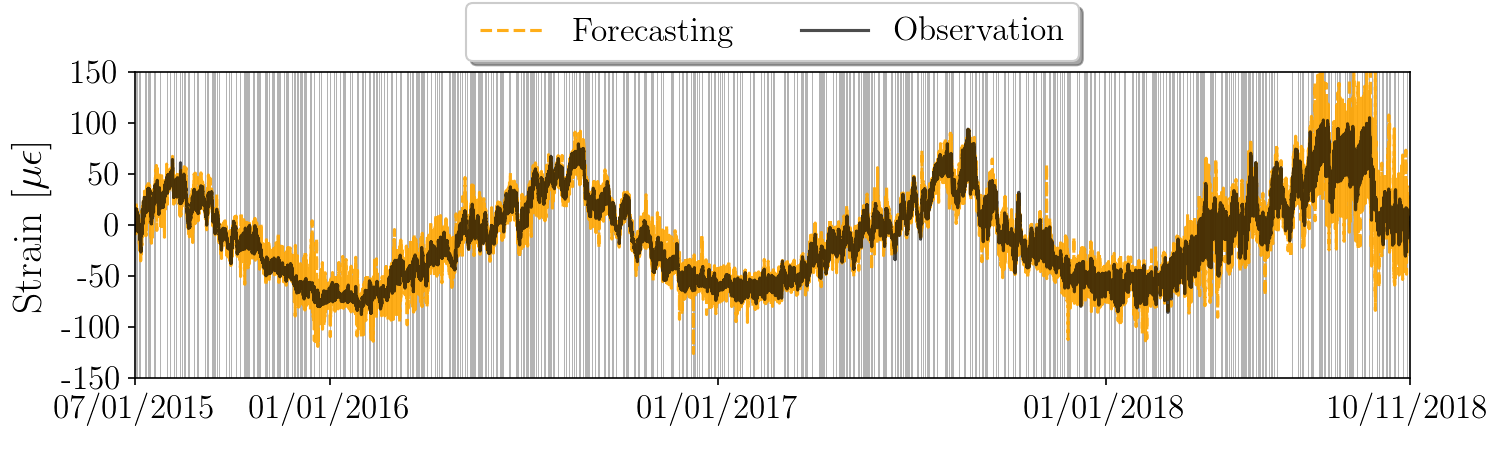}
        \label{fig:mix2_pred}} 
	\vspace{0pt}
	\caption{The forecasting result for four missing cases of Sensor  $\text{S}_{2}\text{-}4$. Note that the shading areas represent the time periods where data missing occurs, while the white box areas denote that the strain time series are successfully recorded. The forecasting dataset is from July 1, 2015 to October 11, 2018 with 40 months.}
	\label{fig:pred_eval}
\end{figure}

\begin{figure}[t!]
	\centering
	    \subfigure[RM (year 2016)]{\includegraphics[width=0.3\linewidth]{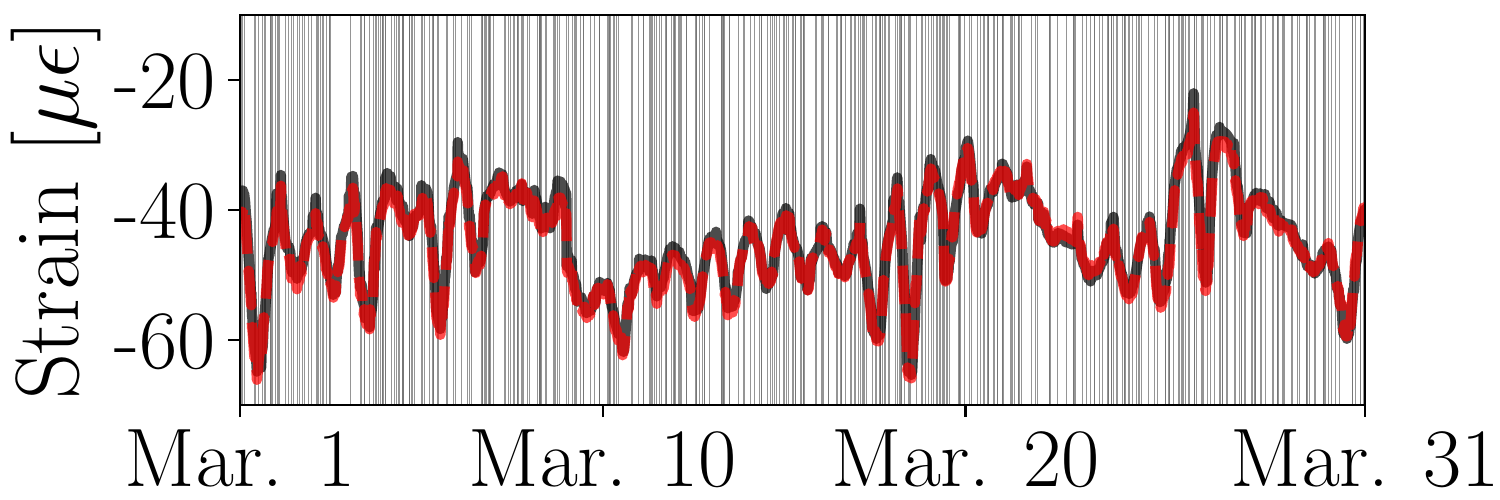}
	    \label{fig:rm_impu_1}} 
	    \subfigure[RM (year 2017)]{\includegraphics[width=0.3\linewidth]{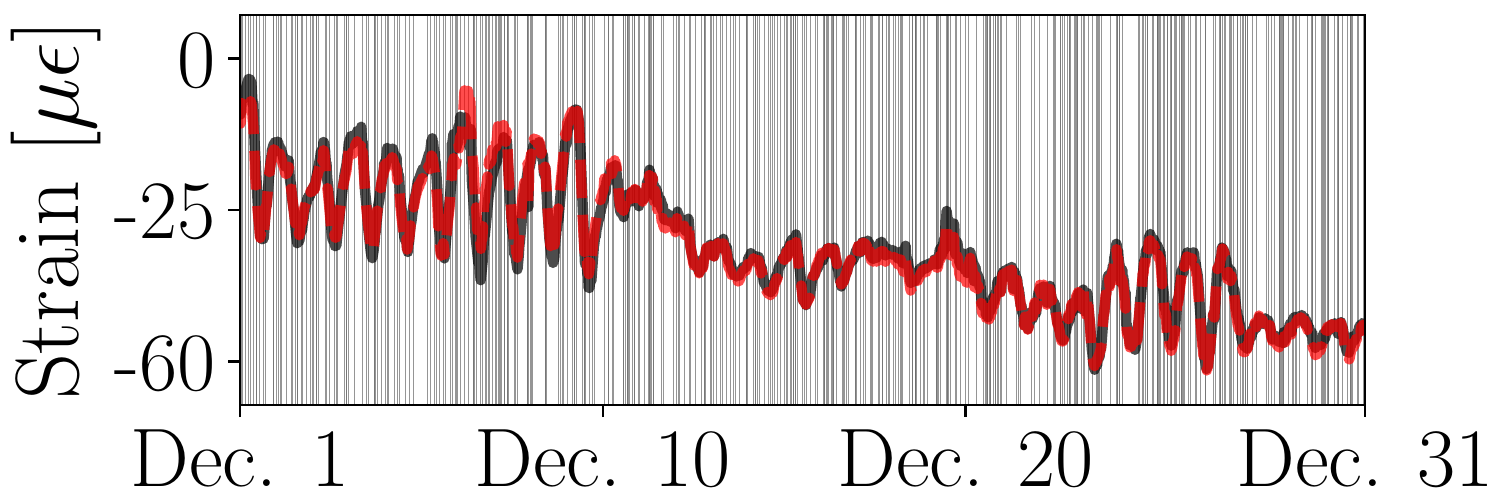}
	    \label{fig:rm_impu_2}} 
	    \subfigure[RM (year 2018)]{\includegraphics[width=0.3\linewidth]{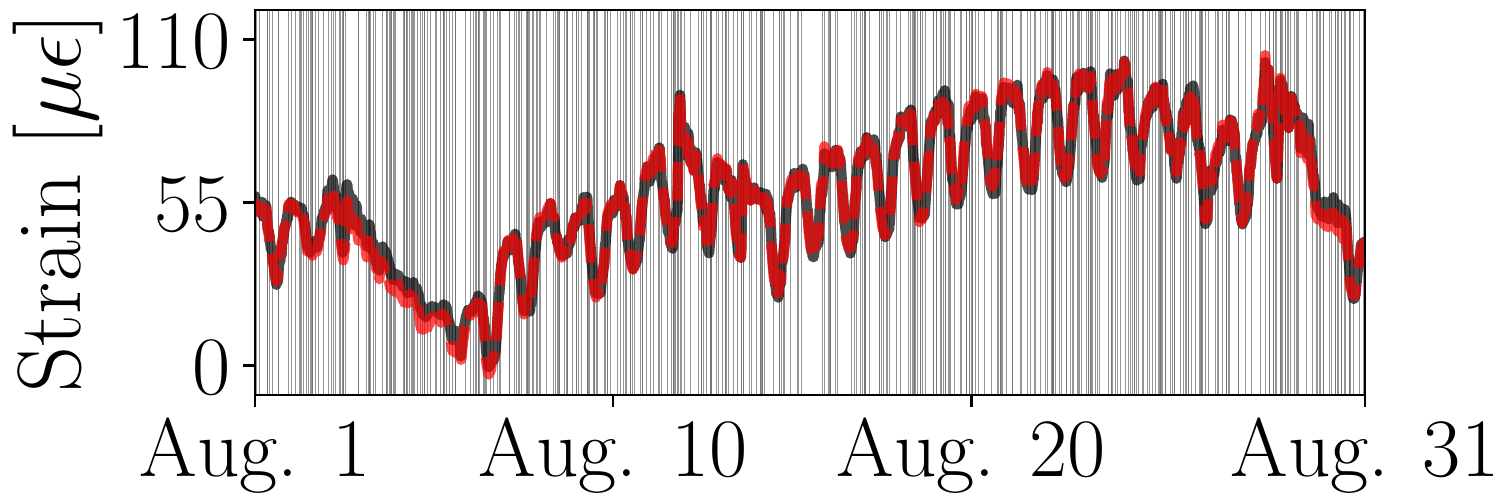}
	    \label{fig:rm_impu_3}} 
	    \subfigure[SM (year 2016)]{\includegraphics[width=0.3\linewidth]{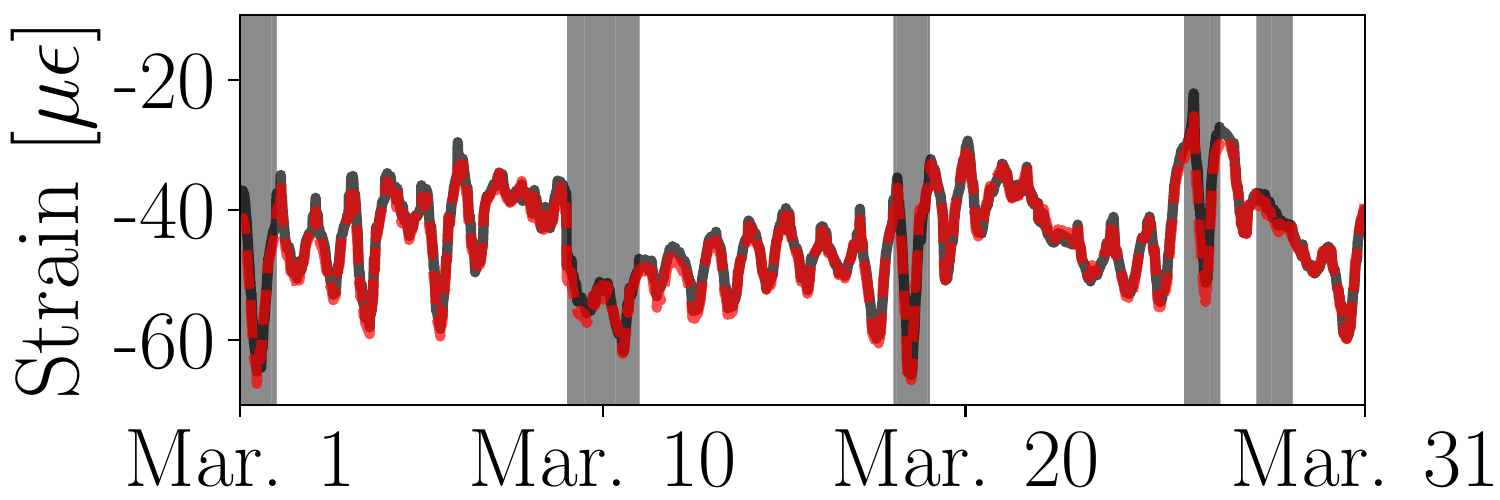}
	    \label{fig:sm_impu_1}} 
	    \subfigure[SM (year 2017)]{\includegraphics[width=0.3\linewidth]{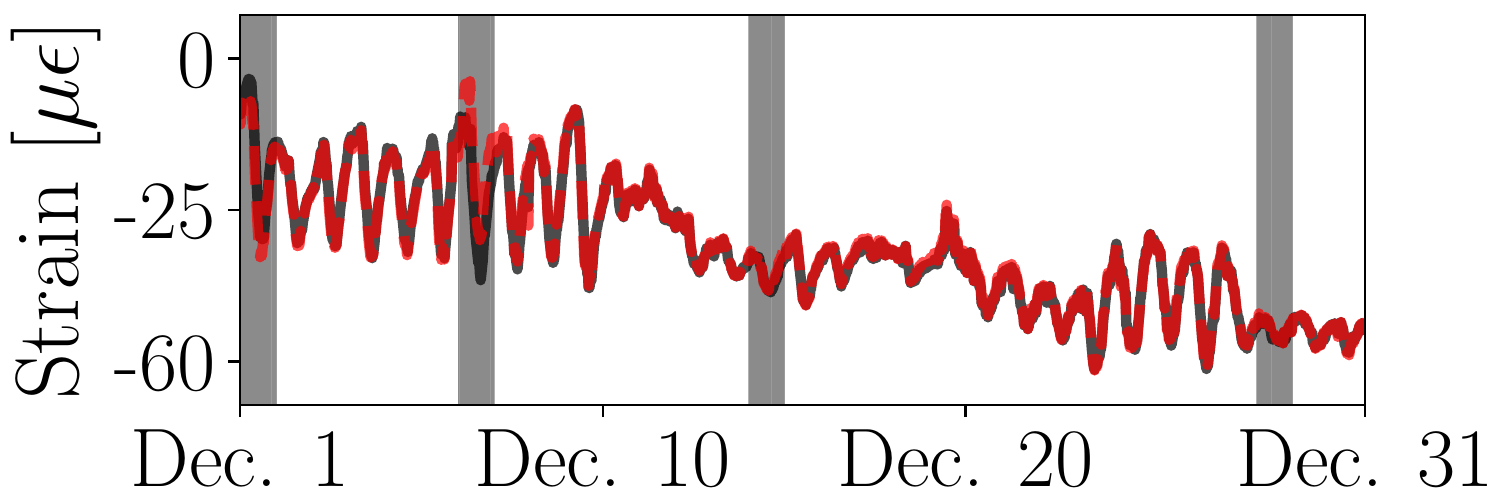}
	    \label{fig:sm_impu_2}} 
	    \subfigure[SM (year 2018)]{\includegraphics[width=0.3\linewidth]{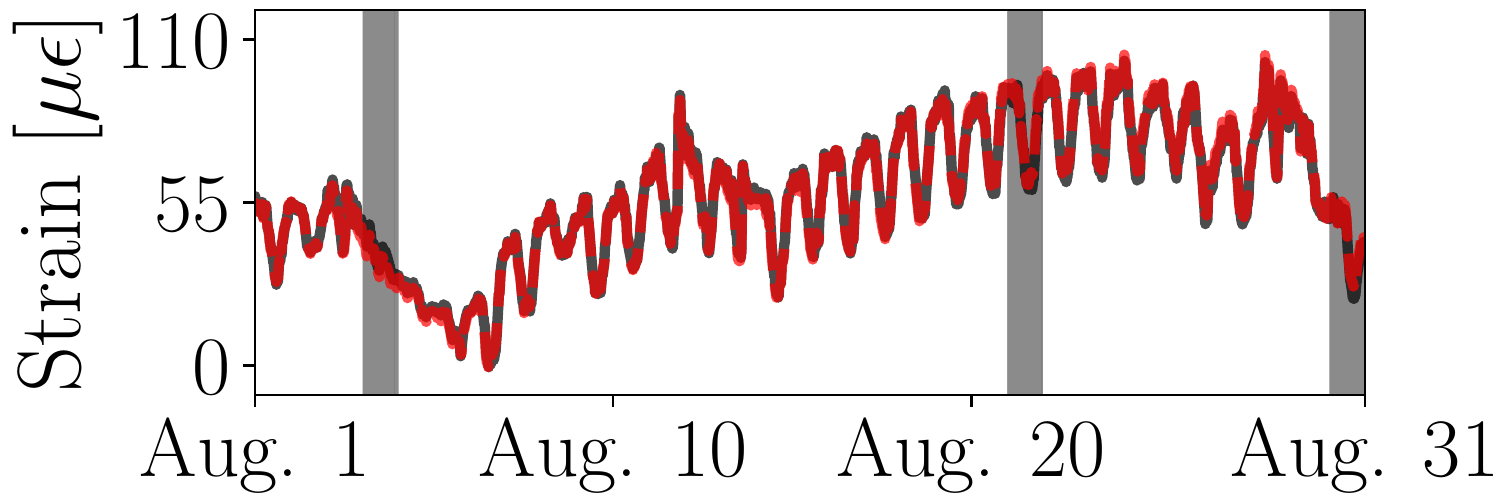}
	    \label{fig:sm_impu_3}} 
	    \subfigure[MM for Case 1 (year 2016)]{\includegraphics[width=0.3\linewidth]{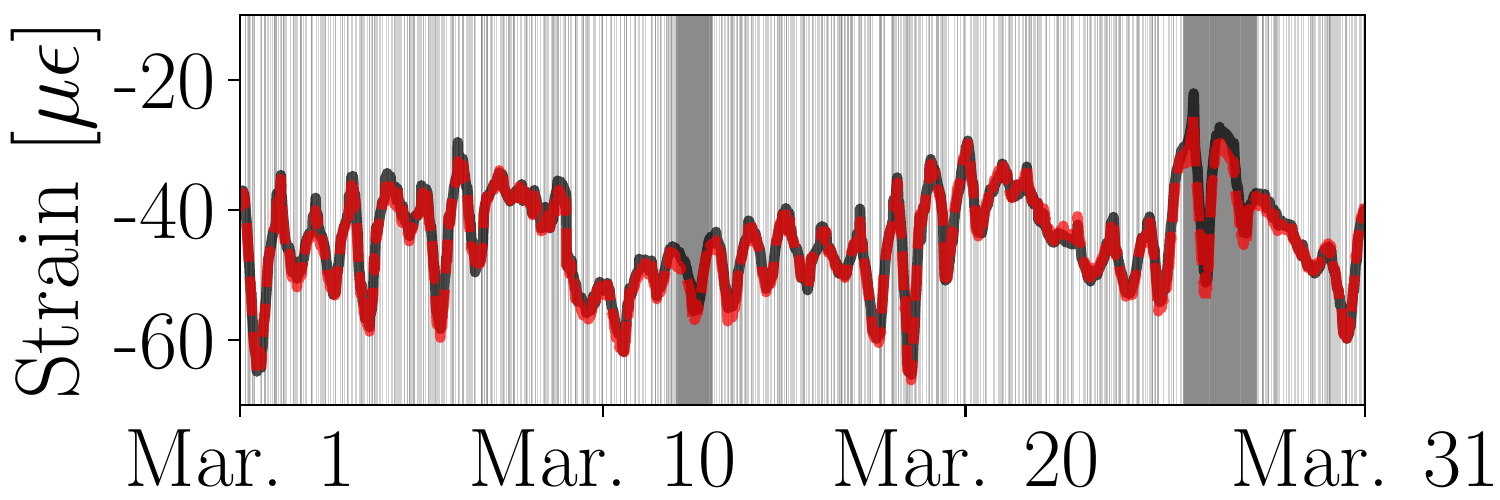}
	    \label{fig:mix1_impu_1}} 
	    \subfigure[MM for Case 1 (year 2017)]{\includegraphics[width=0.3\linewidth]{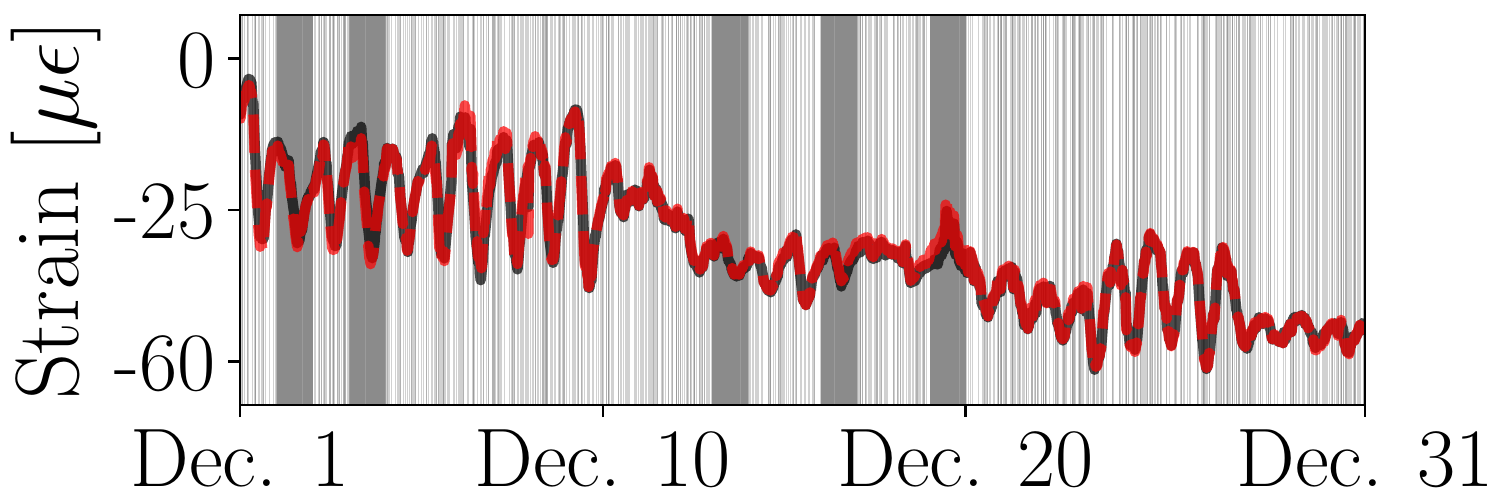}
	    \label{fig:mix1_impu_2}} 
	    \subfigure[MM for Case 1 (year 2018)]{\includegraphics[width=0.3\linewidth]{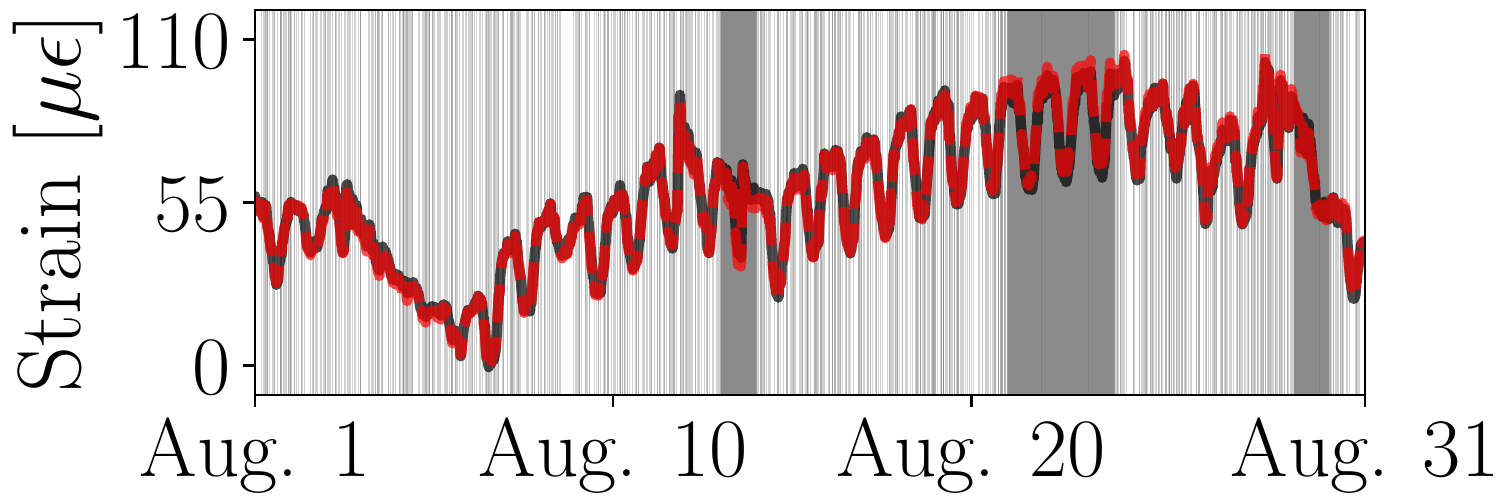}
	    \label{fig:mix1_impu_3}} 
	    \subfigure[MM for Case 2 (year 2016)]{\includegraphics[width=0.3\linewidth]{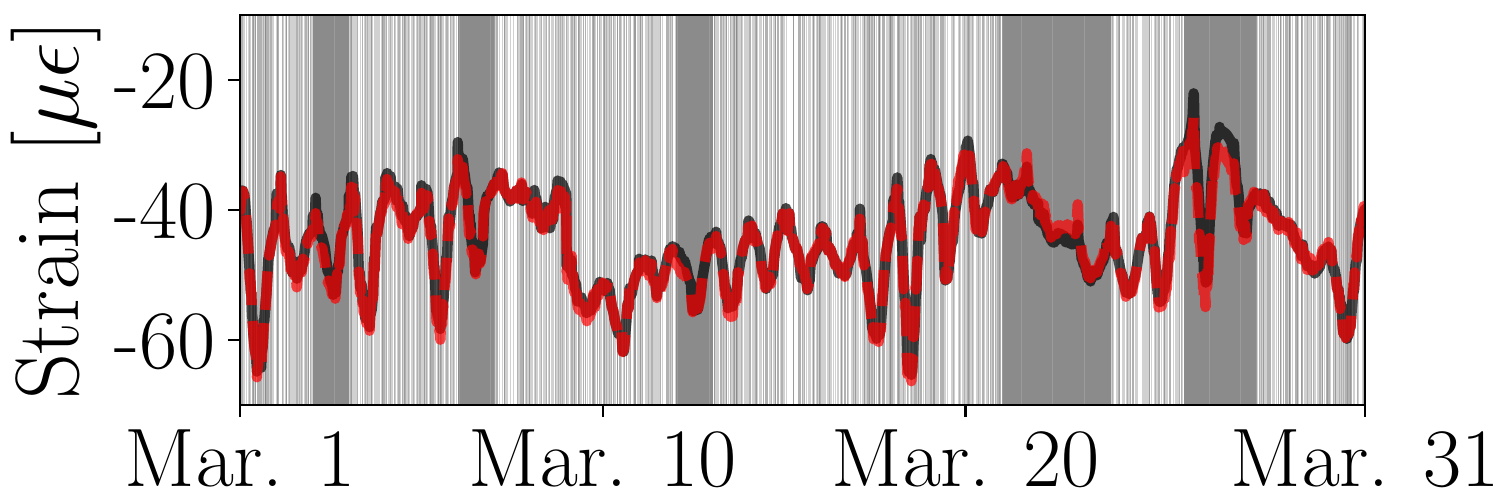}
	    \label{fig:mix2_impu_1}} 
	    \subfigure[MM for Case 2 (year 2017)]{\includegraphics[width=0.3\linewidth]{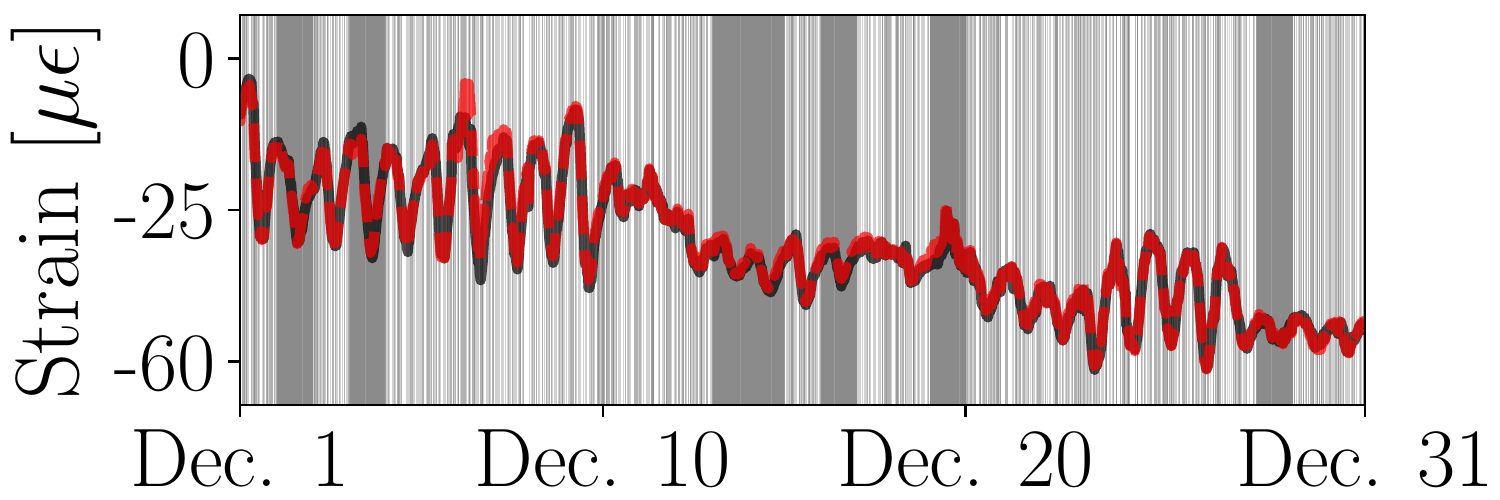}
	    \label{fig:mix2_impu_2}} 
	    \subfigure[MM for Case 2 (year 2018)]{\includegraphics[width=0.3\linewidth]{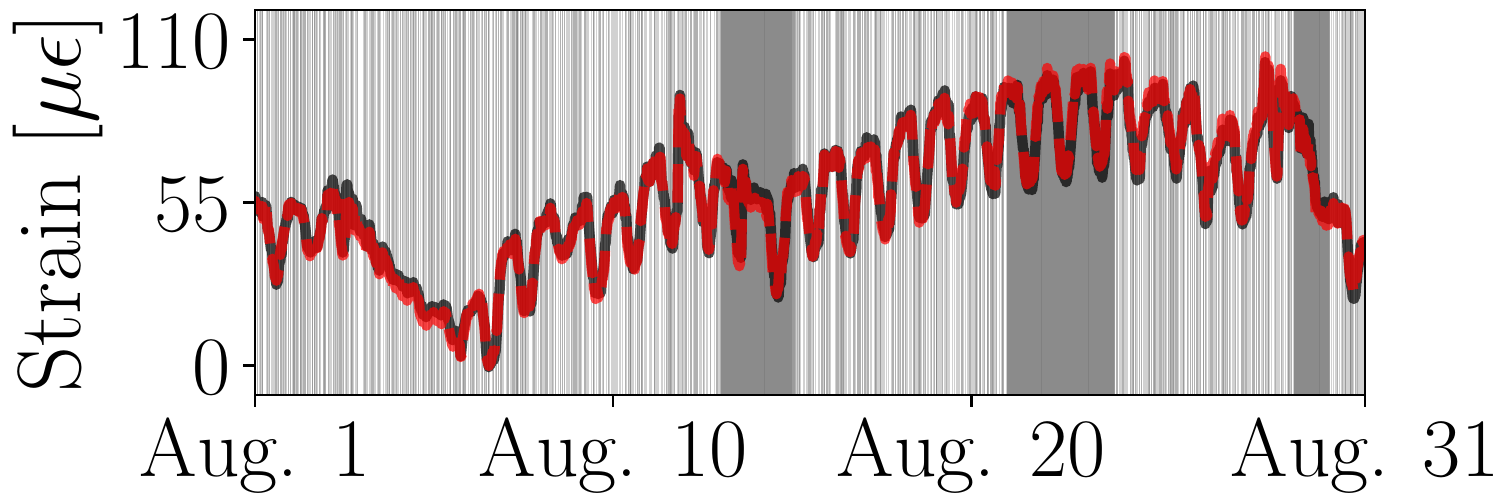}
	    \label{fig:mix3_impu_3}} 
	\vspace{0pt}
	\caption{The zoomed view of the imputed strain time series in Figure \ref{fig:impu_eval}. Note that the shading areas represent the time periods where data missing occurs, while the white box areas denote that the strain time series are successfully recorded. The black lines and the red dashed lines depict the one-month field measurement and the imputed time histories, respectively.}
	\label{fig:impu_one_month} 
\end{figure}

\begin{figure}[t!]
	\centering
	    \subfigure[RM (year 2016)]{\includegraphics[width=0.3\linewidth]{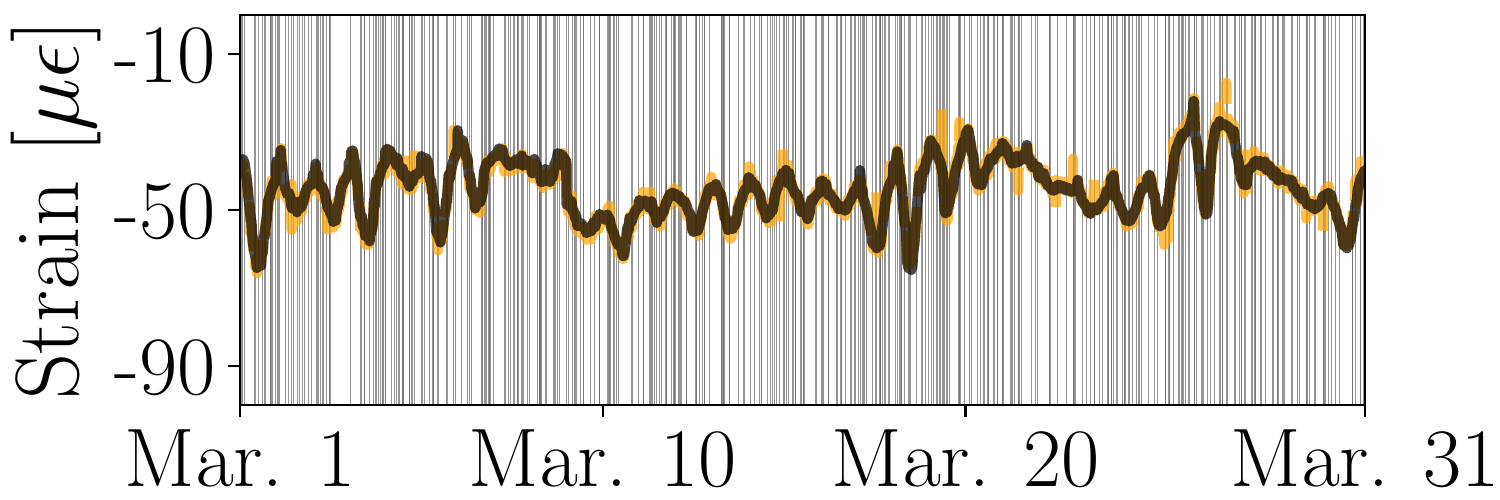}
	    \label{fig:rm_pred_1}} 
	    \subfigure[RM (year 2017)]{\includegraphics[width=0.3\linewidth]{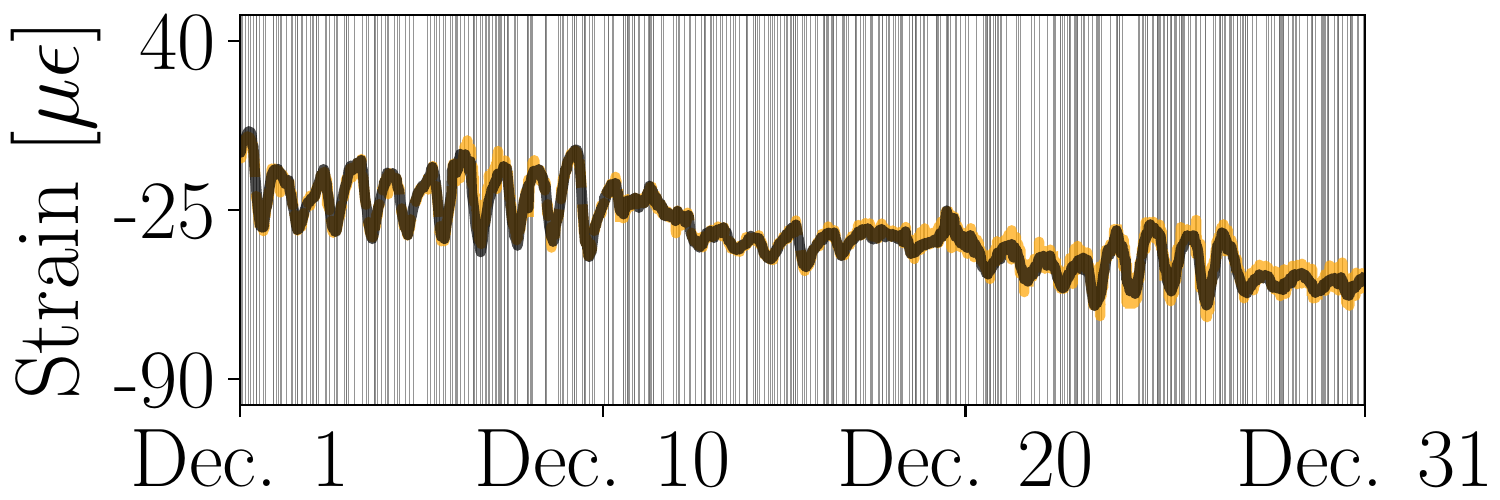}
	    \label{fig:rm_pred_2}} 
	    \subfigure[RM (year 2018)]{\includegraphics[width=0.3\linewidth]{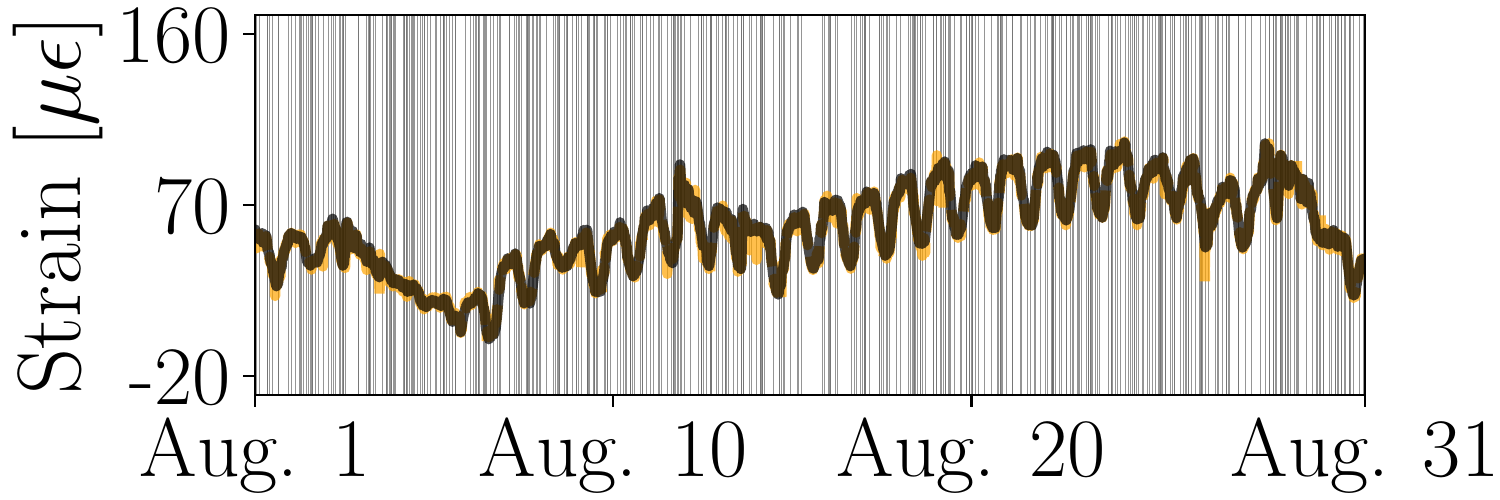}
	    \label{fig:rm_pred_3}} 
	    \subfigure[SM (year 2016)]{\includegraphics[width=0.3\linewidth]{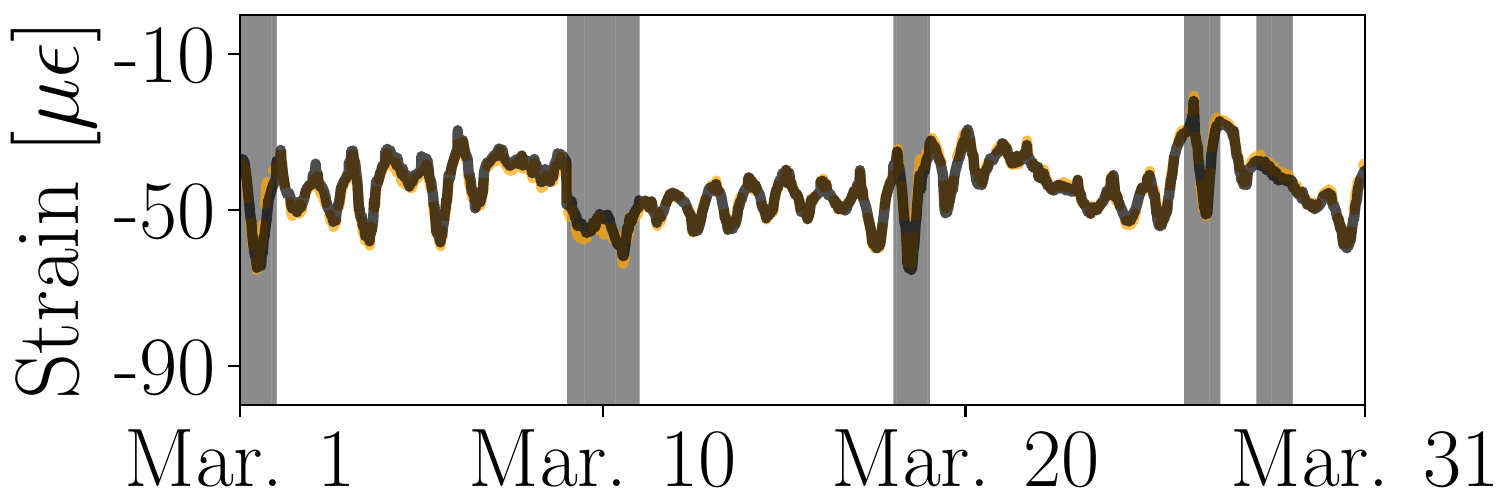}
	    \label{fig:sm_pred_1}} 
	    \subfigure[SM (year 2017)]{\includegraphics[width=0.3\linewidth]{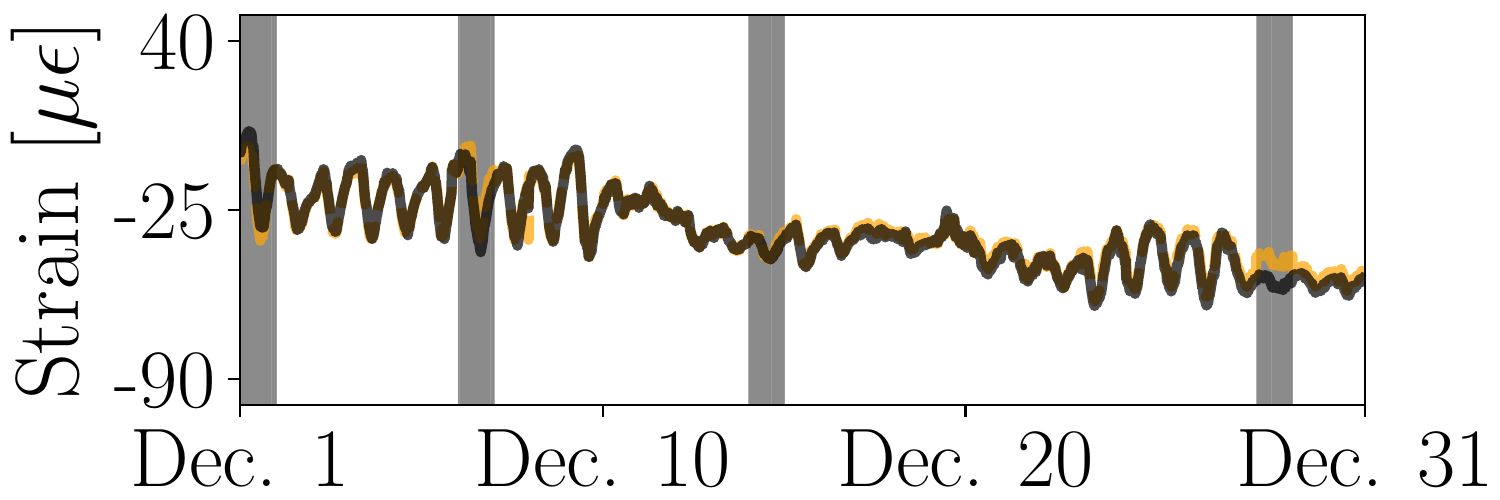}
	    \label{fig:sm_pred_2}} 
	    \subfigure[SM (year 2018)]{\includegraphics[width=0.3\linewidth]{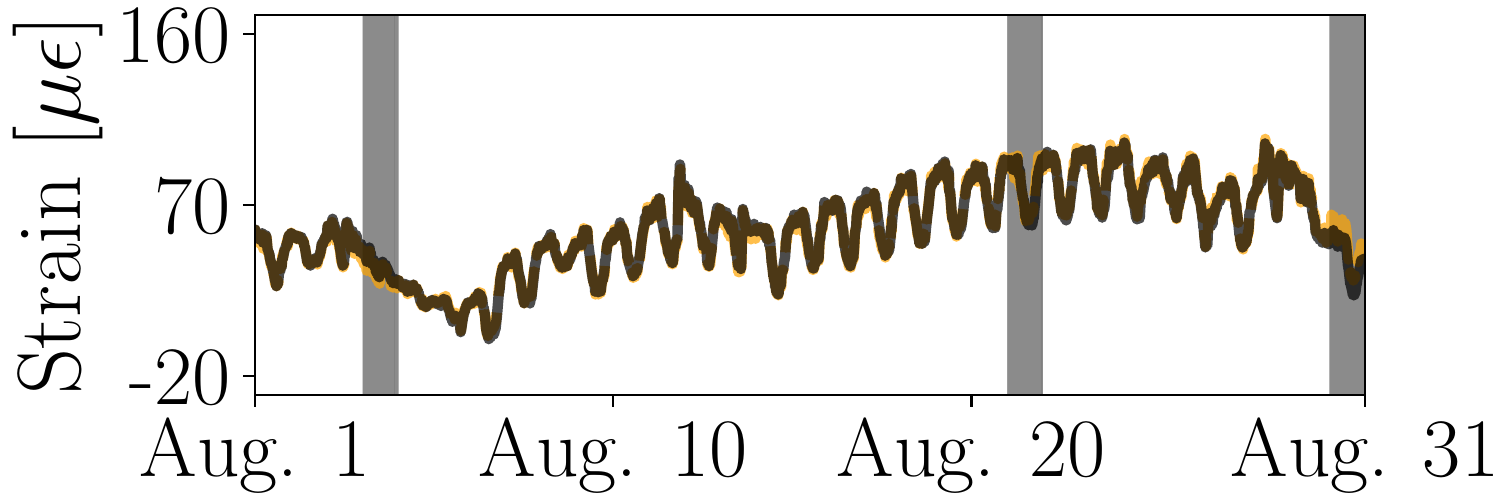}
	    \label{fig:sm_pred_3}} 
	    \subfigure[MM for Case 1 (year 2016)]{\includegraphics[width=0.3\linewidth]{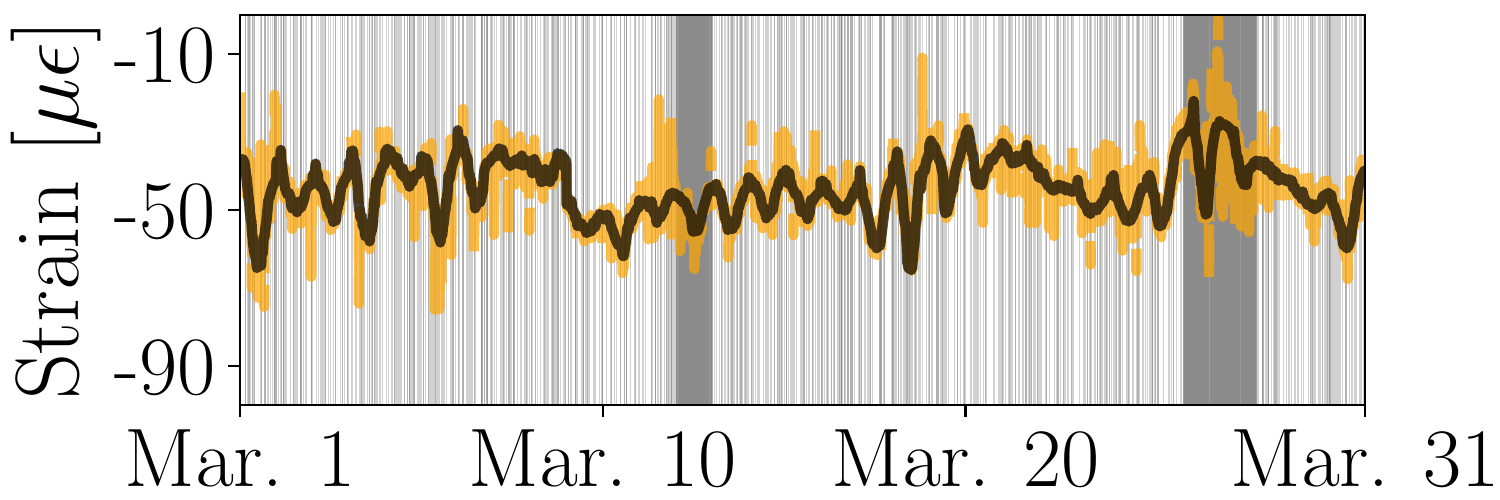}
	    \label{fig:mix1_pred_1}} 
	    \subfigure[MM for Case 1 (year 2017)]{\includegraphics[width=0.3\linewidth]{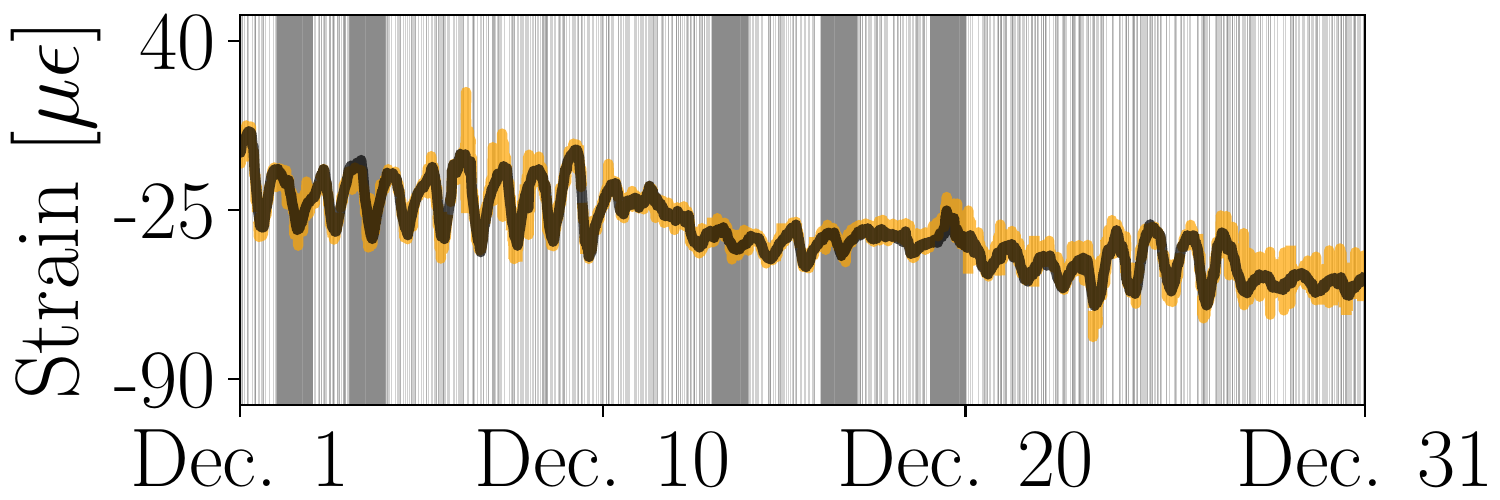}
	    \label{fig:mix1_pred_2}} 
	    \subfigure[MM for Case 1 (year 2018)]{\includegraphics[width=0.3\linewidth]{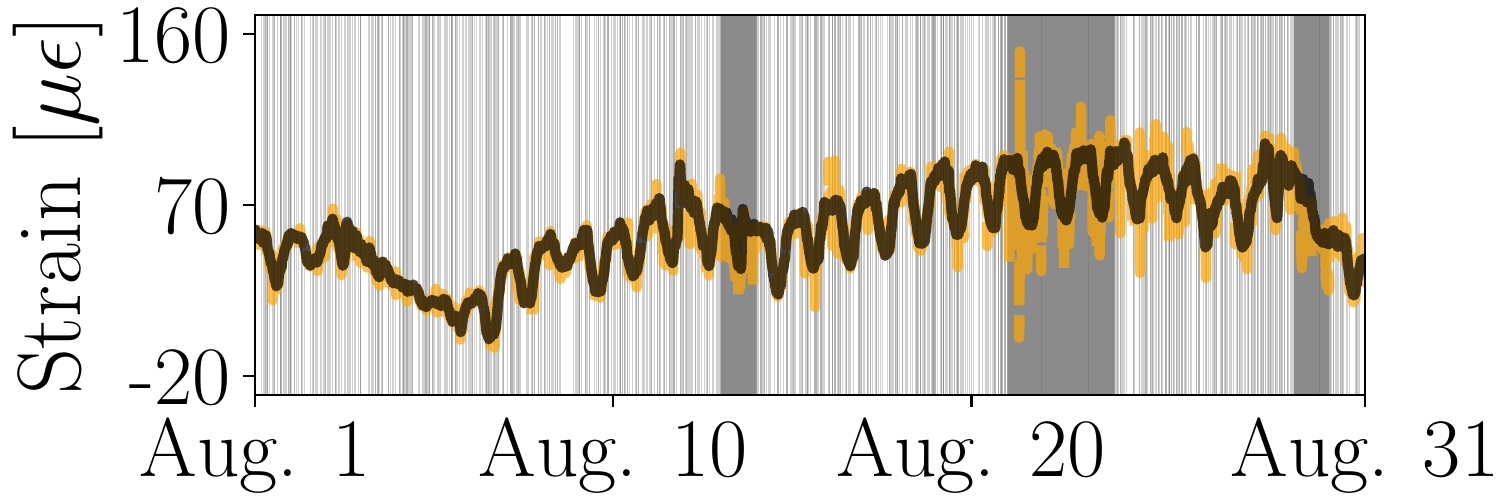}
	    \label{fig:mix1_pred_3}} 
	    \subfigure[MM for Case 2 (year 2016)]{\includegraphics[width=0.3\linewidth]{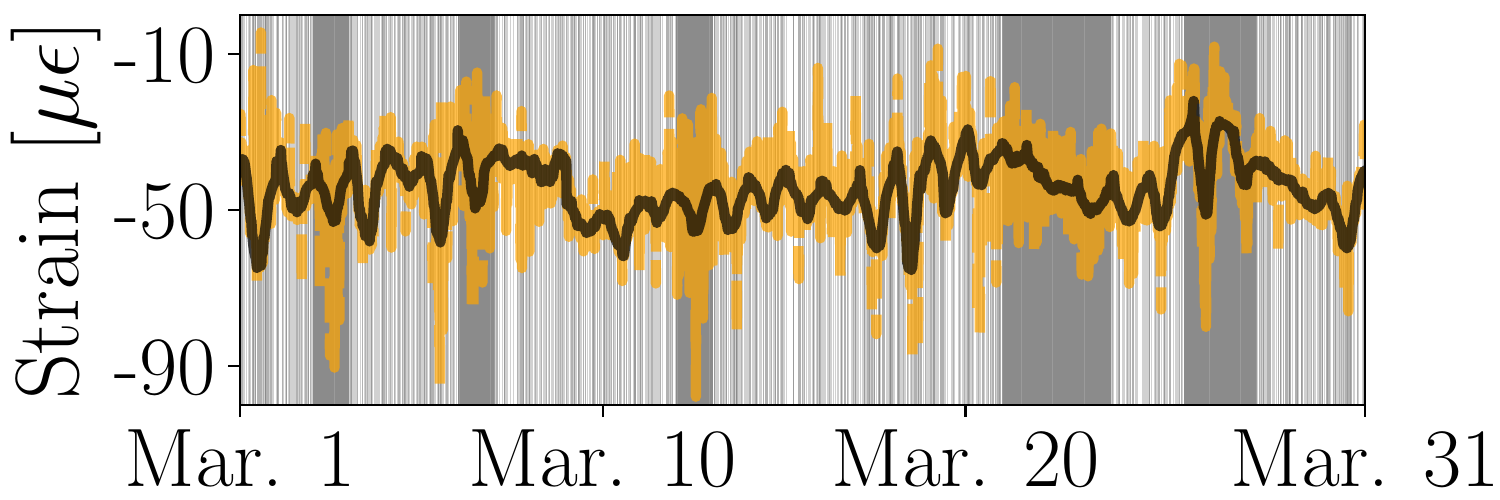}
	    \label{fig:mix2_pred_1}} 
	    \subfigure[MM for Case 2 (year 2017)]{\includegraphics[width=0.3\linewidth]{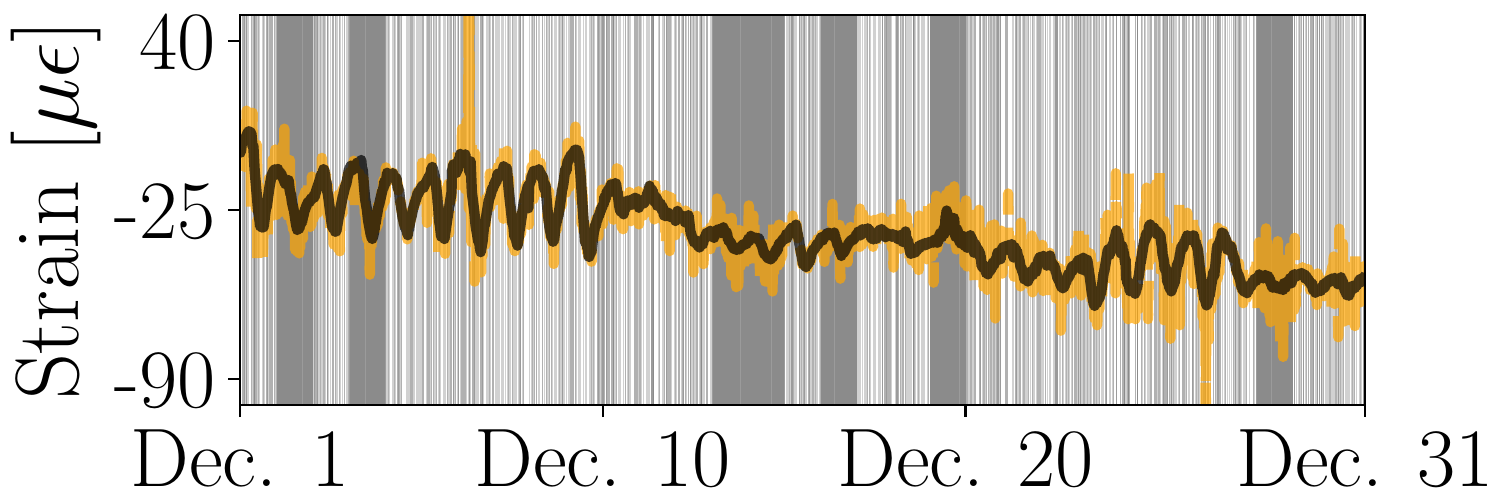}
	    \label{fig:mix2_pred_2}} 
	    \subfigure[MM for Case 2 (year 2018)]{\includegraphics[width=0.3\linewidth]{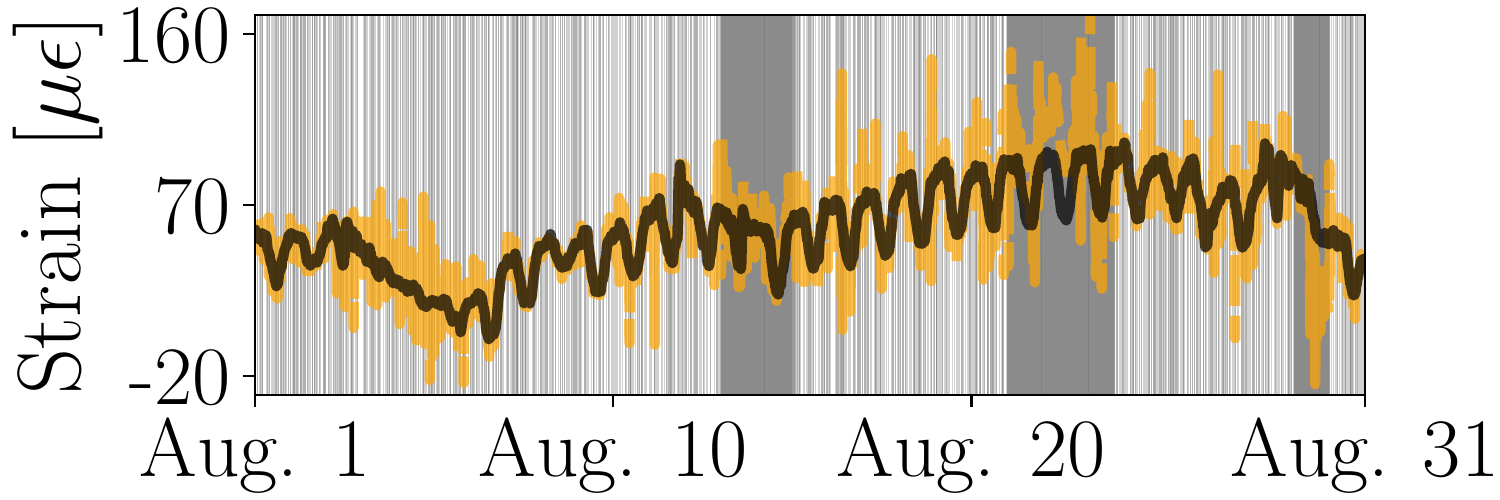}
	    \label{fig:mix3_pred_3}} 
	\vspace{0pt}
	\caption{The zoomed view of the predicted strain time series in Figure \ref{fig:pred_eval}. Note that the shading areas represent the time periods where data missing occurs, while the white box areas denote that the strain time series are successfully recorded. The black lines and the orange dashed lines depict the one-month field measurement and the forecasted time histories, respectively.}
	\label{fig:pred_one_month} 
\end{figure}

\subsection{Results}\label{S:3.3}
We test the overall performance of the proposed method and and identify its limit of capacity for data imputation and response forecasting under various missing settings with different missing rates. In the incremental learning, the forward batch window length $I$ is defined as 30 days with 4,320 data points, while the critical time step $T_1$ is one year (i.e., $12\times30$). The dataset for imputation ranges from June 1, 2015 to October 11, 2018 with 1,230 days in total, and the forecasting data is from July 1, 2015 to October 11, 2018 with thirty-days data ahead of the imputation dataset. To begin with, we first consider the missing rate of 10\%, for both random and structured missing scenarios, while setting the tensor rank of eight. In addition, keeping the tensor rank fixed, we also set two mixed missing cases: Case 1 for 10\% structured and 20\% random missing occurring at the same time, while Case 2 for 20\% structured and 30\% random missing simultaneously. Here, sensor $\text{S}_{2}\text{-}4$ is selected to showcase the result. 

Figure \ref{fig:impu_eval} and \ref{fig:pred_eval} show the corresponding imputation and forecasting result obtained by the proposed incremental Bayesian tensor learning model. It can be seen that the predicted time series match well with the ground truth. In particular, the imputed data possess excellent agreement with the ground truth (see Figure \ref{fig:impu_eval}), while the forecasted response has relatively larger errors especially for the mixed missing cases with overall large missing rates (e.g., Case 1 and Case 2) as shown in Figure \ref{fig:mix1_pred} and \subref{fig:mix2_pred}. In general, the spatiotemporal dependencies of the data are well learned by the proposed model. Besides, we provide three representative segments (zoomed view) of the predicted response by choosing one-month strain (March 2016, December 2017 and August 2018) for showcase of imputation (see Figure \ref{fig:impu_one_month}) and forecasting (see Figure \ref{fig:pred_one_month}).


It is notable that, despite large missing rates, the imputation is very robust and produces excellent estimation as shown in Figure \ref{fig:impu_one_month}. Though the forecasted responses exhibit noisy oscillations depicted in Figure \ref{fig:pred_one_month}, the overall trend is well captured (especially for relatively smaller missing rates, e.g., 10\%).

\begin{figure}[t!]
	\centering
	    \subfigure[RM (Mar. 10, 2016)]{\includegraphics[width=0.3\linewidth]{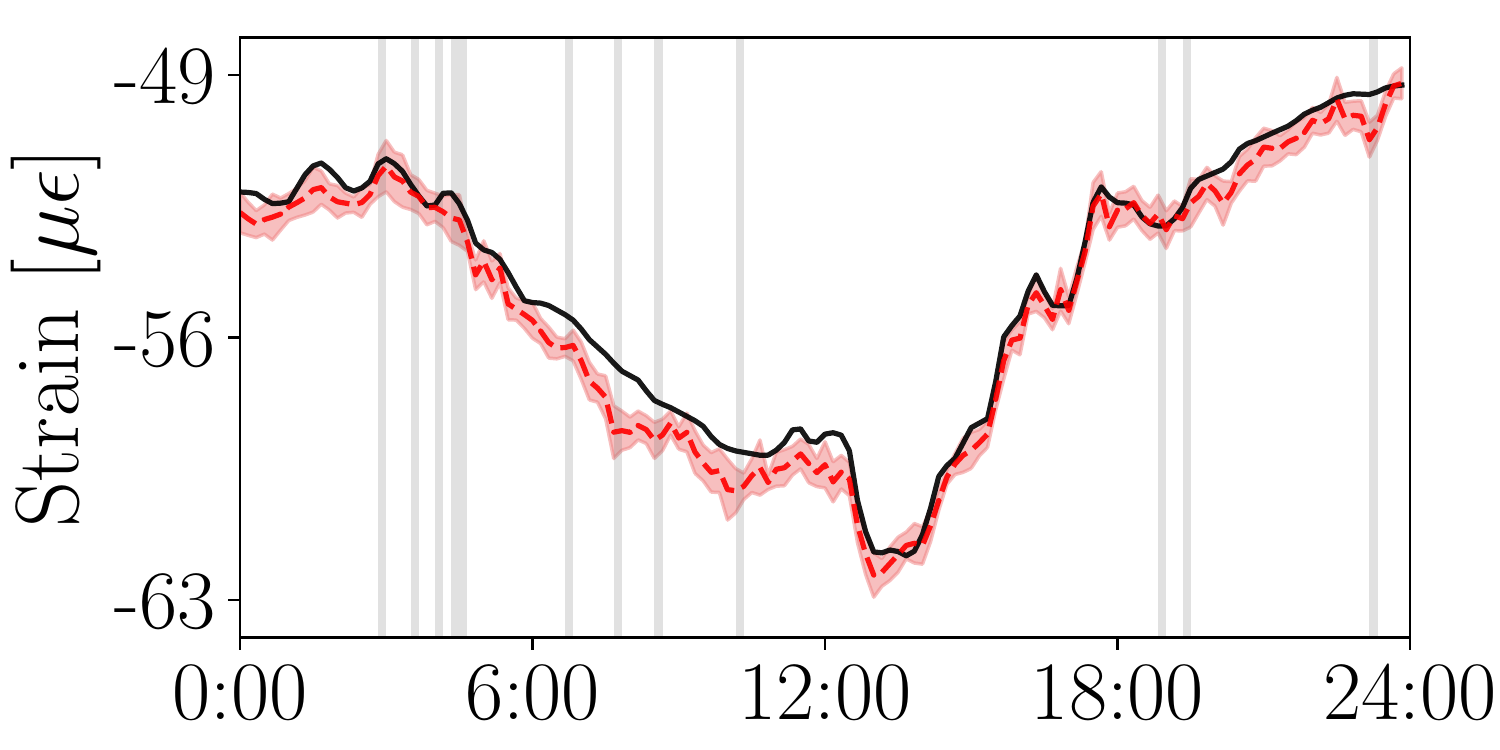}
	    \label{fig:rm_impu_uq_1}} 
	    \subfigure[RM (Dec. 15, 2017)]{\includegraphics[width=0.3\linewidth]{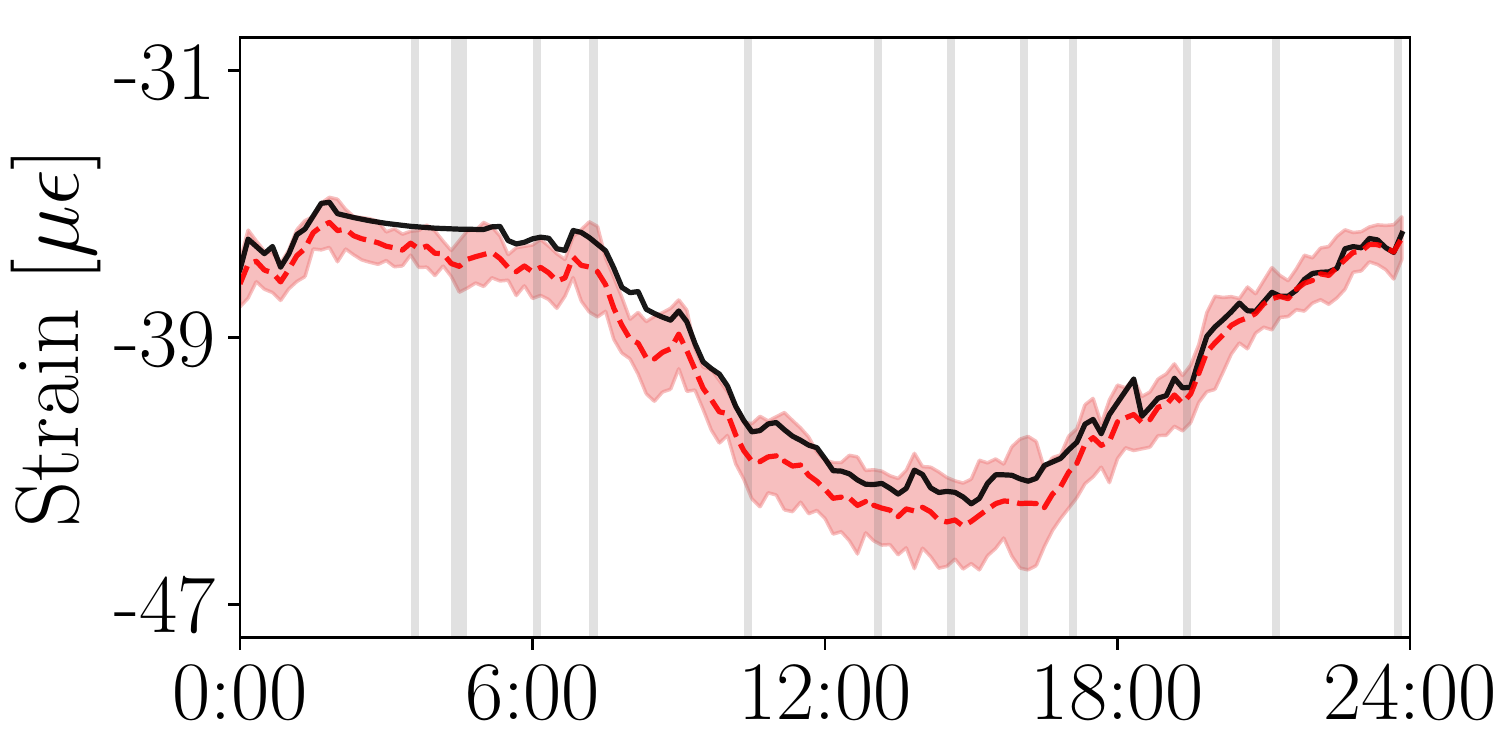}
	    \label{fig:rm_impu_uq_2}} 
	    \subfigure[RM (Aug. 22, 2018)]{\includegraphics[width=0.3\linewidth]{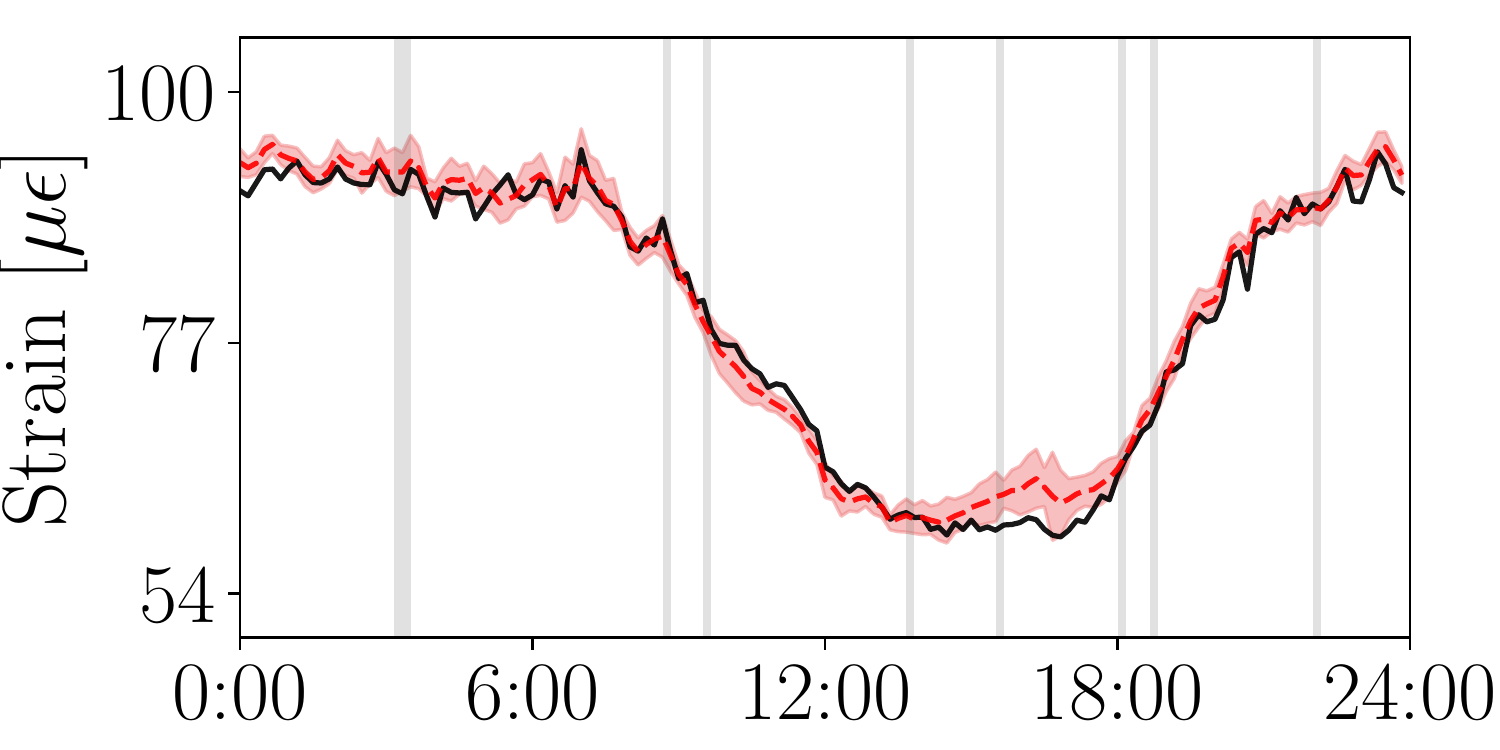}
	    \label{fig:rm_impu_uq_3}} 
	    \subfigure[SM (Mar. 10, 2016)]{\includegraphics[width=0.3\linewidth]{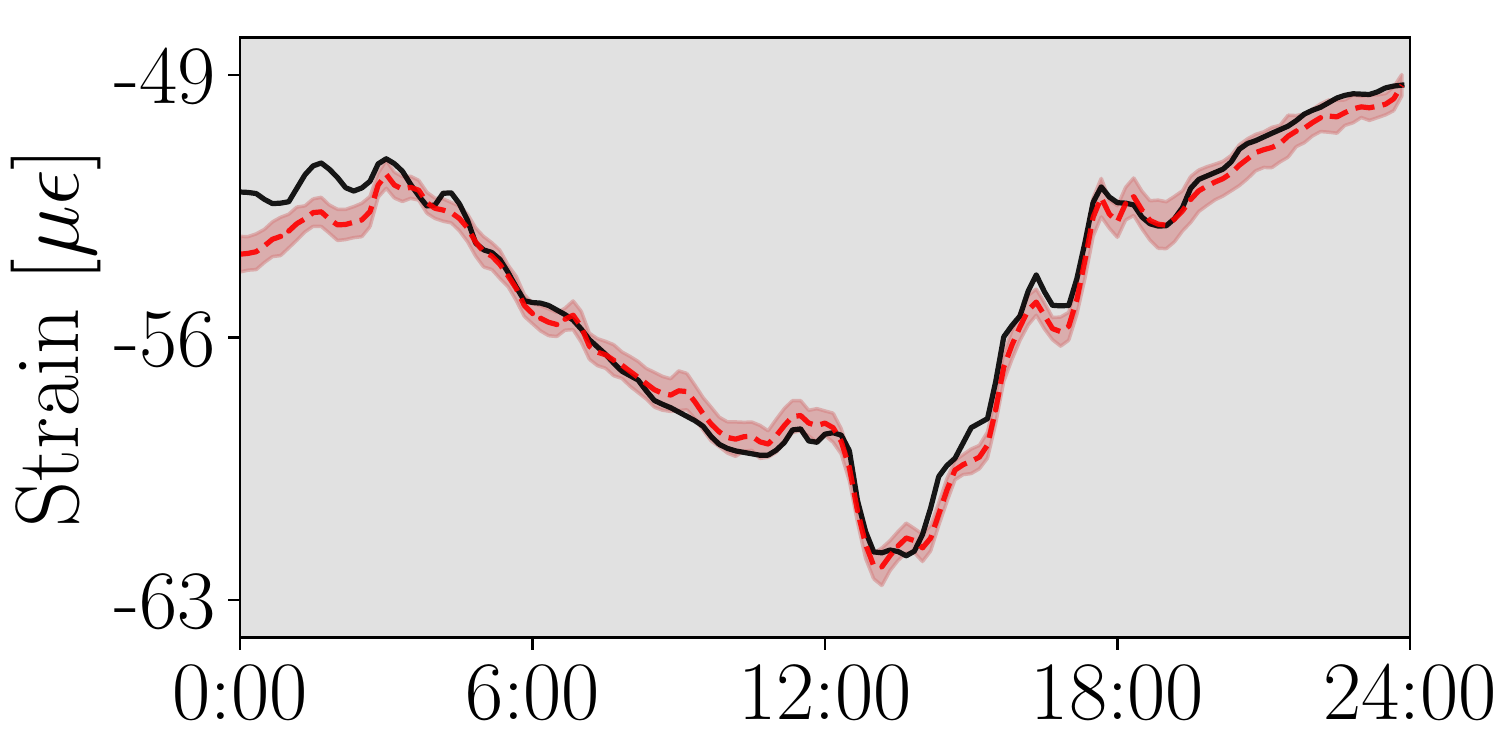}
	    \label{fig:sm_impu_uq_1}} 
	    \subfigure[SM (Dec. 15, 2017)]{\includegraphics[width=0.3\linewidth]{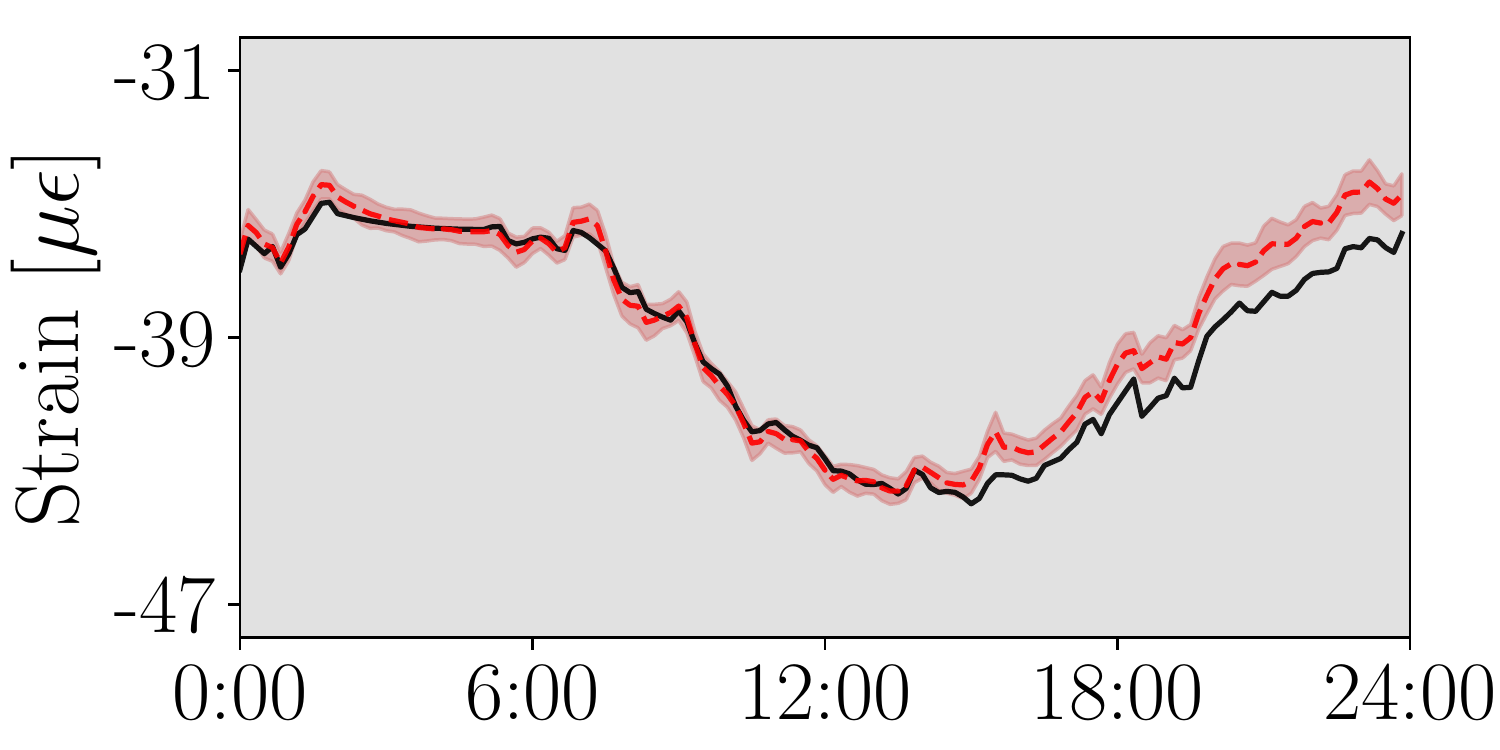}
	    \label{fig:sm_impu_uq_2}} 
	    \subfigure[SM (Aug. 22, 2018)]{\includegraphics[width=0.3\linewidth]{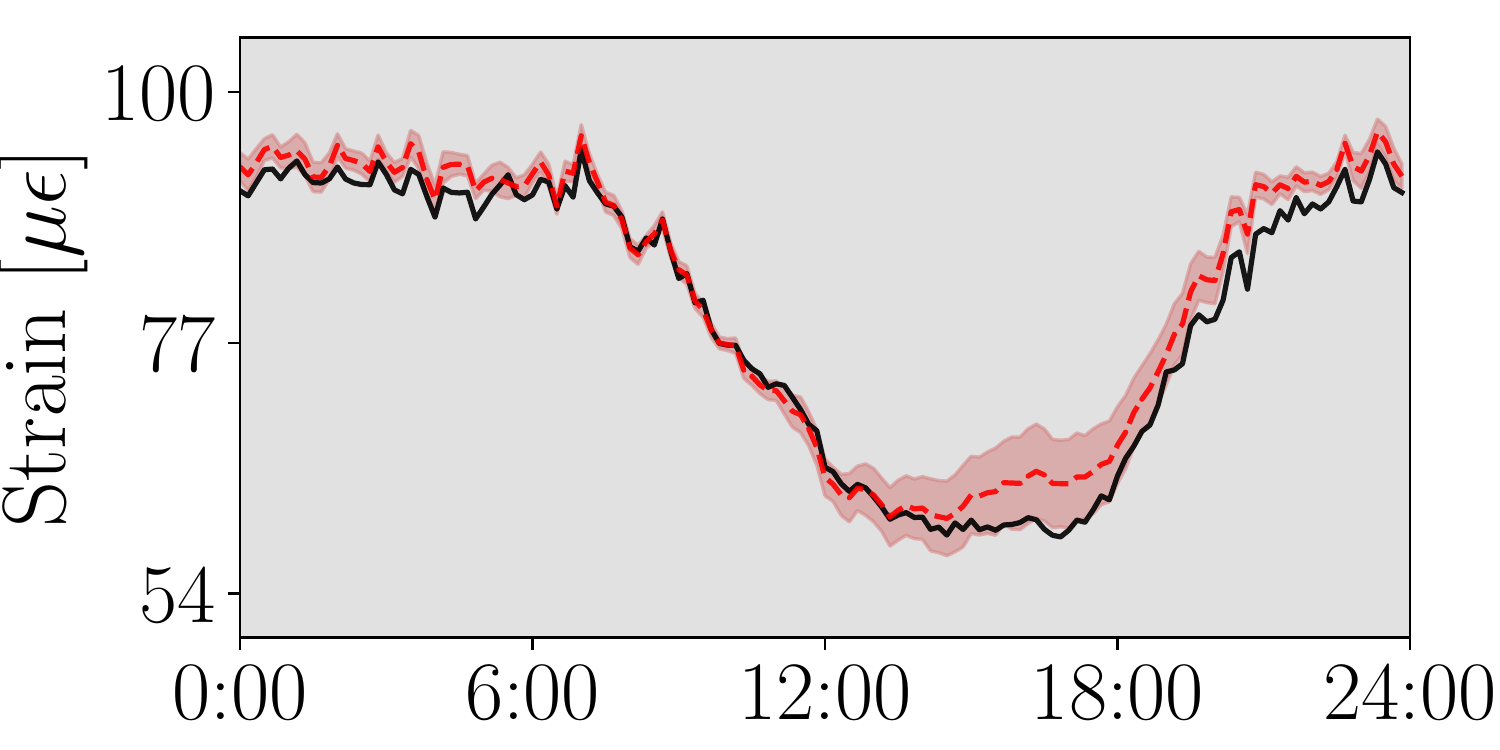}
	    \label{fig:sm_impu_uq_3}} 
	    \subfigure[MM for Case 1 (Mar. 10, 2016)]{\includegraphics[width=0.3\linewidth]{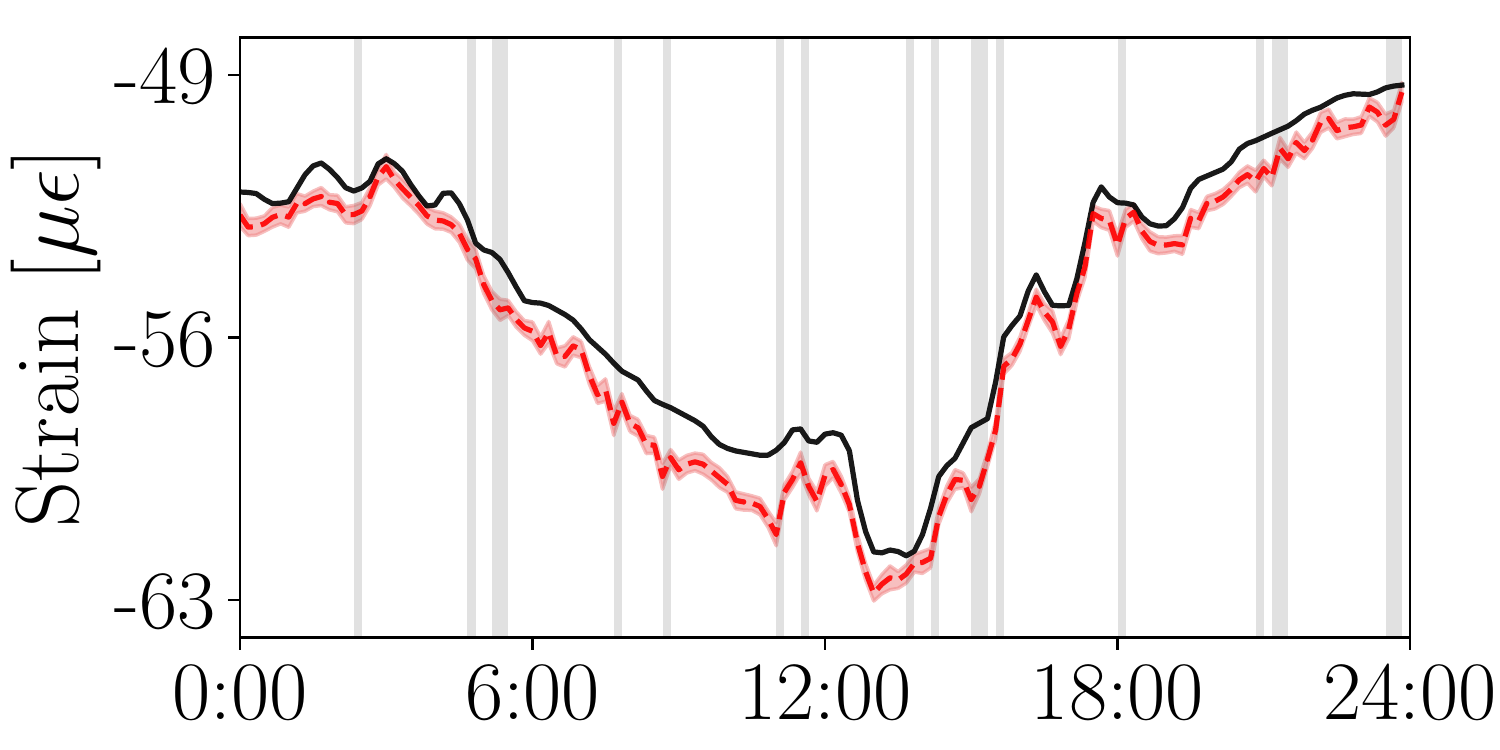}
	    \label{fig:mix1_impu_uq_1}} 
	    \subfigure[MM for Case 1 (Dec. 15, 2017)]{\includegraphics[width=0.3\linewidth]{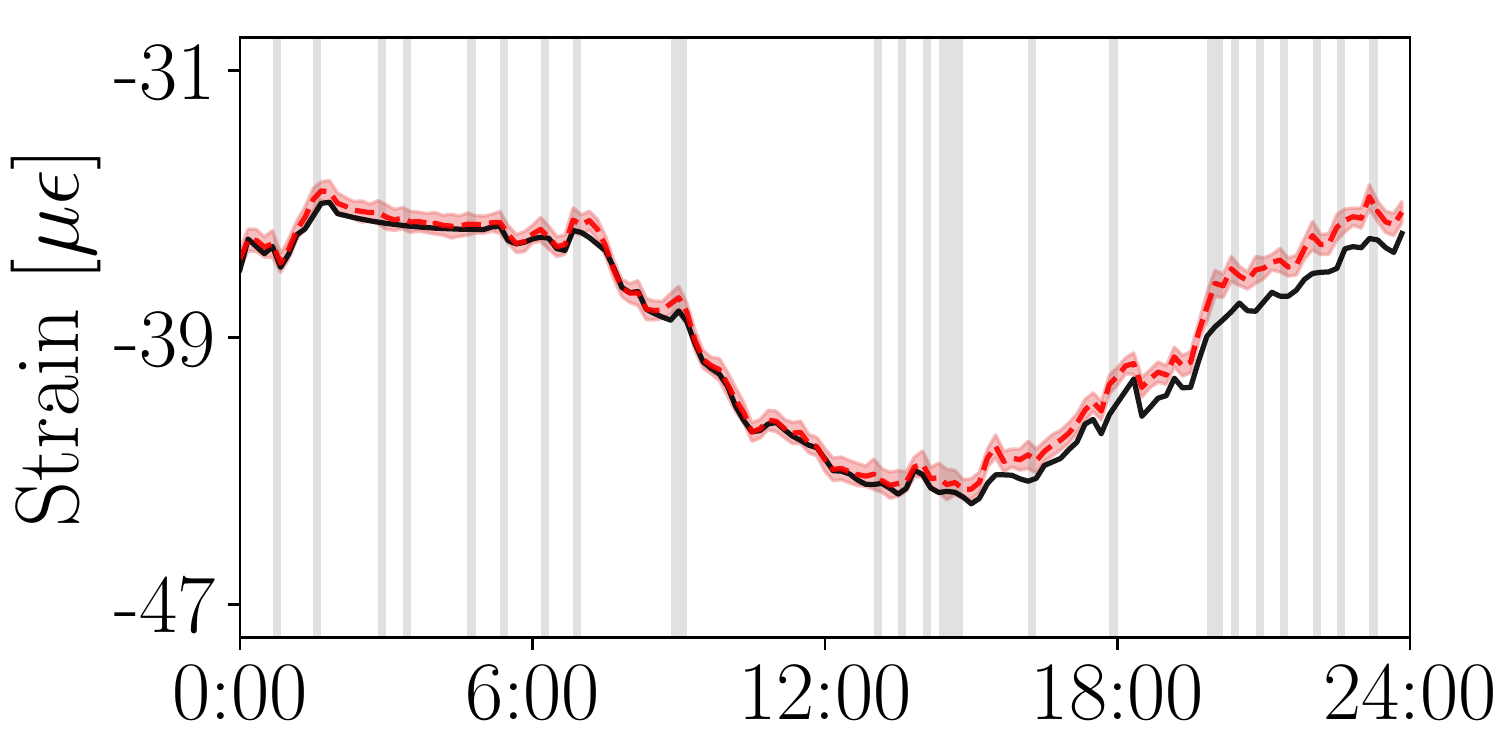}
	    \label{fig:mix1_impu_uq_2}} 
	    \subfigure[MM for Case 1 (Aug. 22, 2018)]{\includegraphics[width=0.3\linewidth]{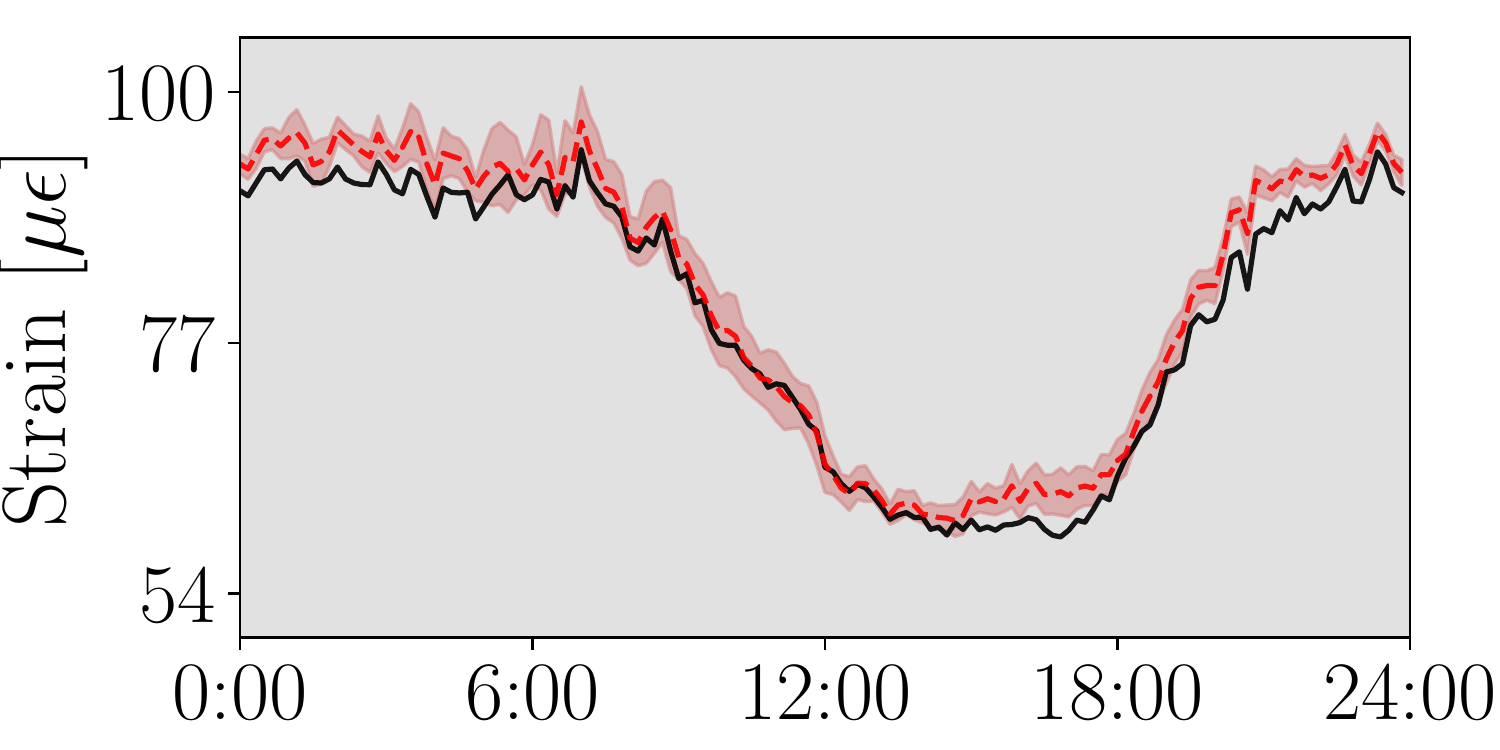}
	    \label{fig:mix1_impu_uq_3}} 
	    \subfigure[MM for Case 2 (Mar. 10, 2016)]{\includegraphics[width=0.3\linewidth]{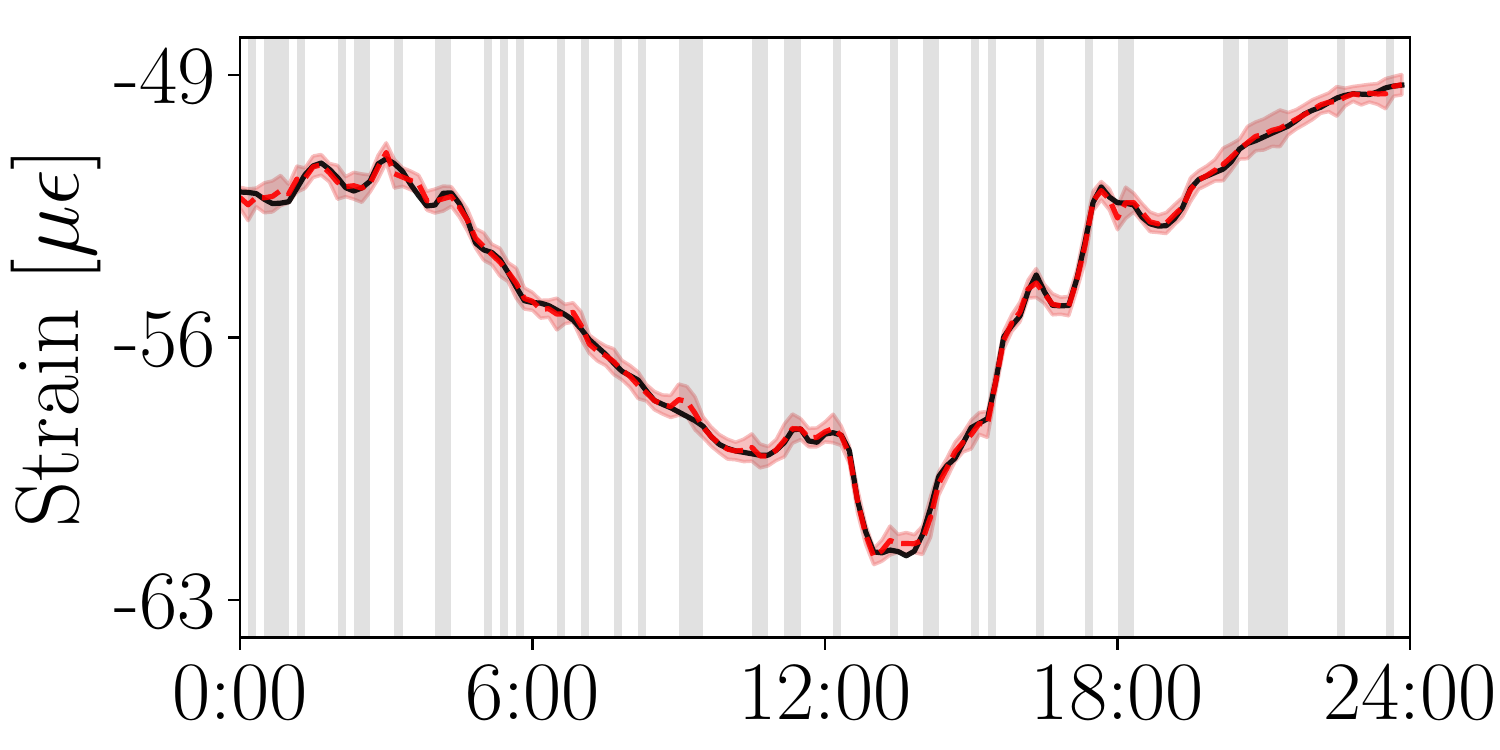}
	    \label{fig:mix2_impu_uq_1}} 
	    \subfigure[MM for Case 2 (Dec. 15, 2017)]{\includegraphics[width=0.3\linewidth]{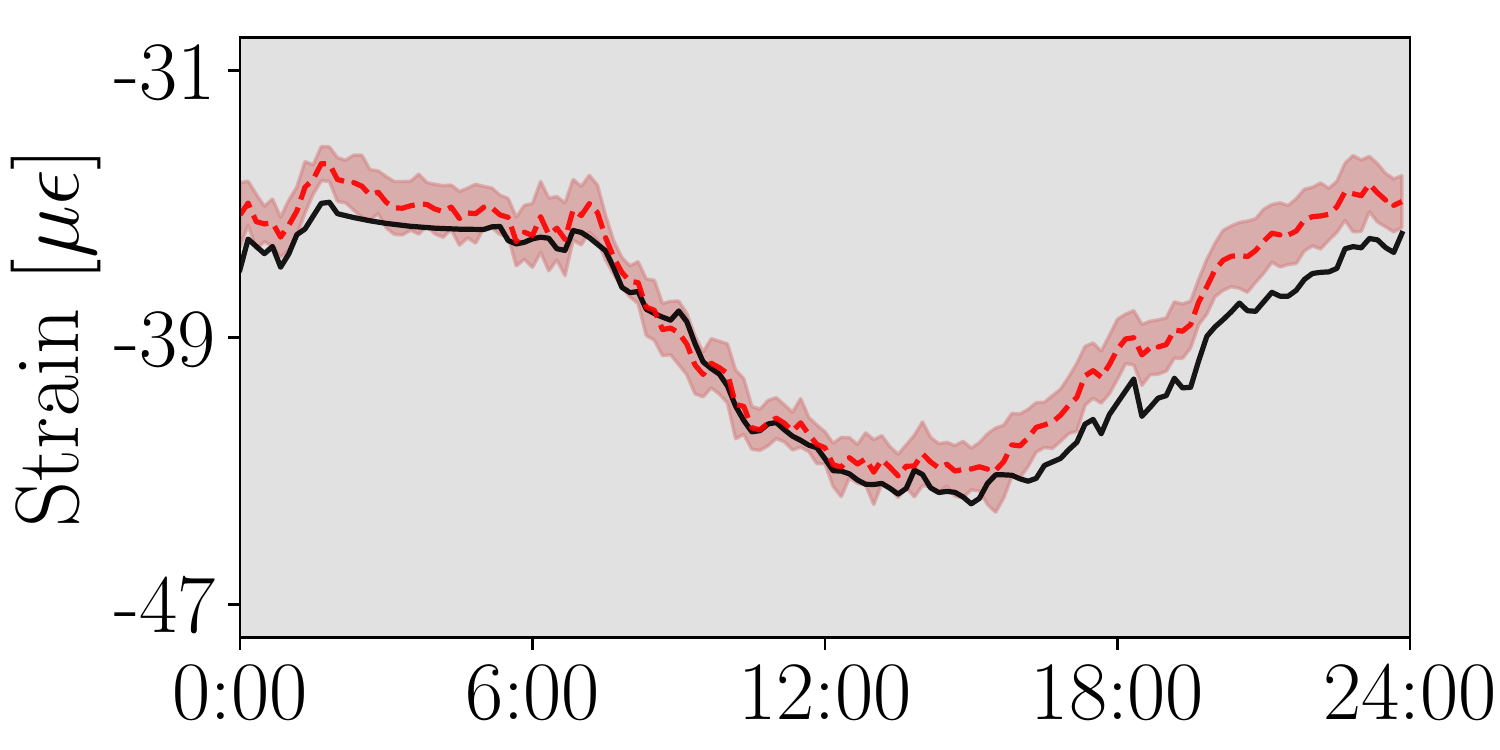}
	    \label{fig:mix2_impu_uq_2}} 
	    \subfigure[MM for Case 2 (Aug. 22, 2018)]{\includegraphics[width=0.3\linewidth]{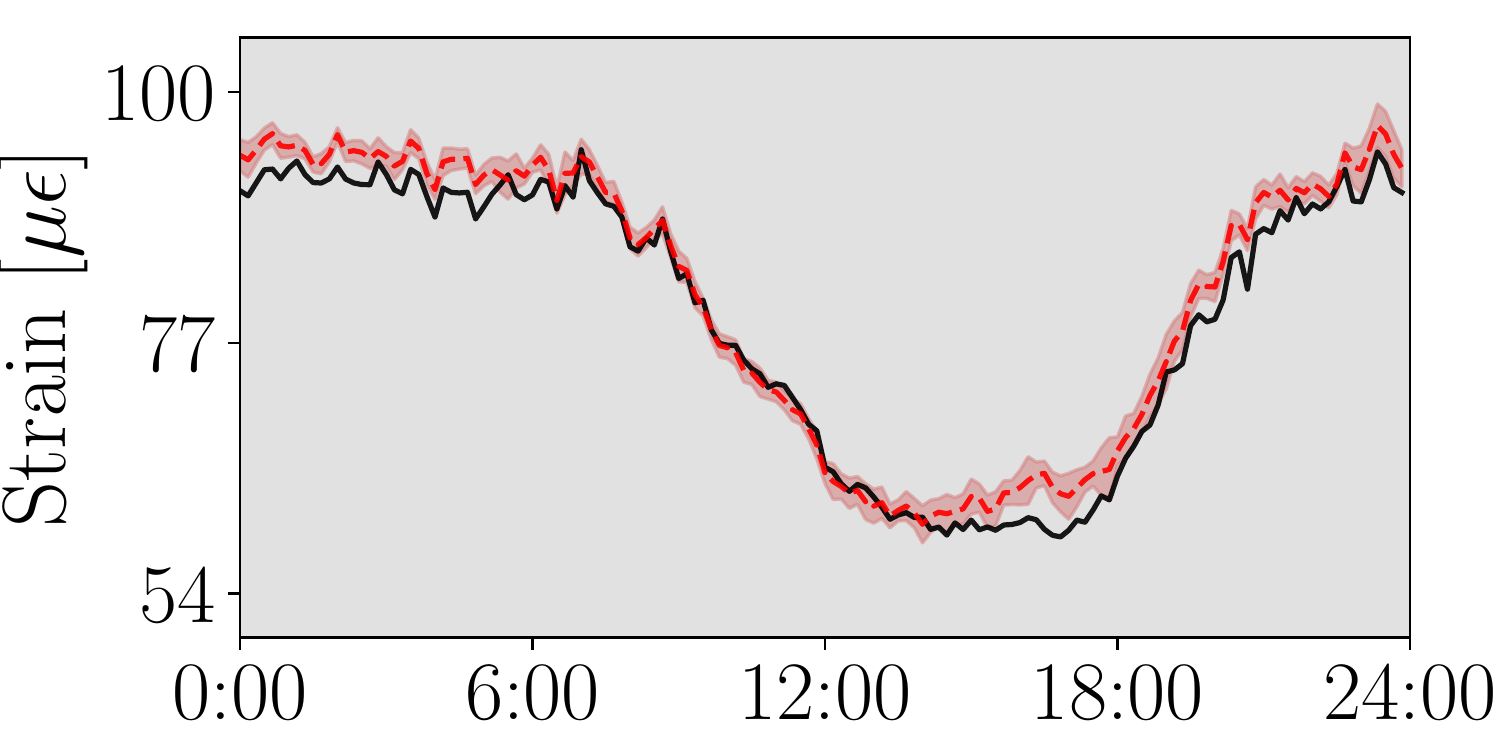}
	    \label{fig:mix3_impu_uq_3}} 
	\vspace{0pt}
	\caption{Uncertainty quantification of imputation for four data missing cases. Note that the shading areas represent the time periods where data missing occurs, while the white box areas denote that the strain time series are successfully recorded. The black lines and the red dashed lines depict the one-day field measurement and the imputed means respectively, and the red band is the area between plus/minus three standard deviations.}
	\label{fig:impu_uq_eval} 
\end{figure}

\begin{figure}[t!]
	\centering
	    \subfigure[RM (Mar. 10, 2016)]{\includegraphics[width=0.3\linewidth]{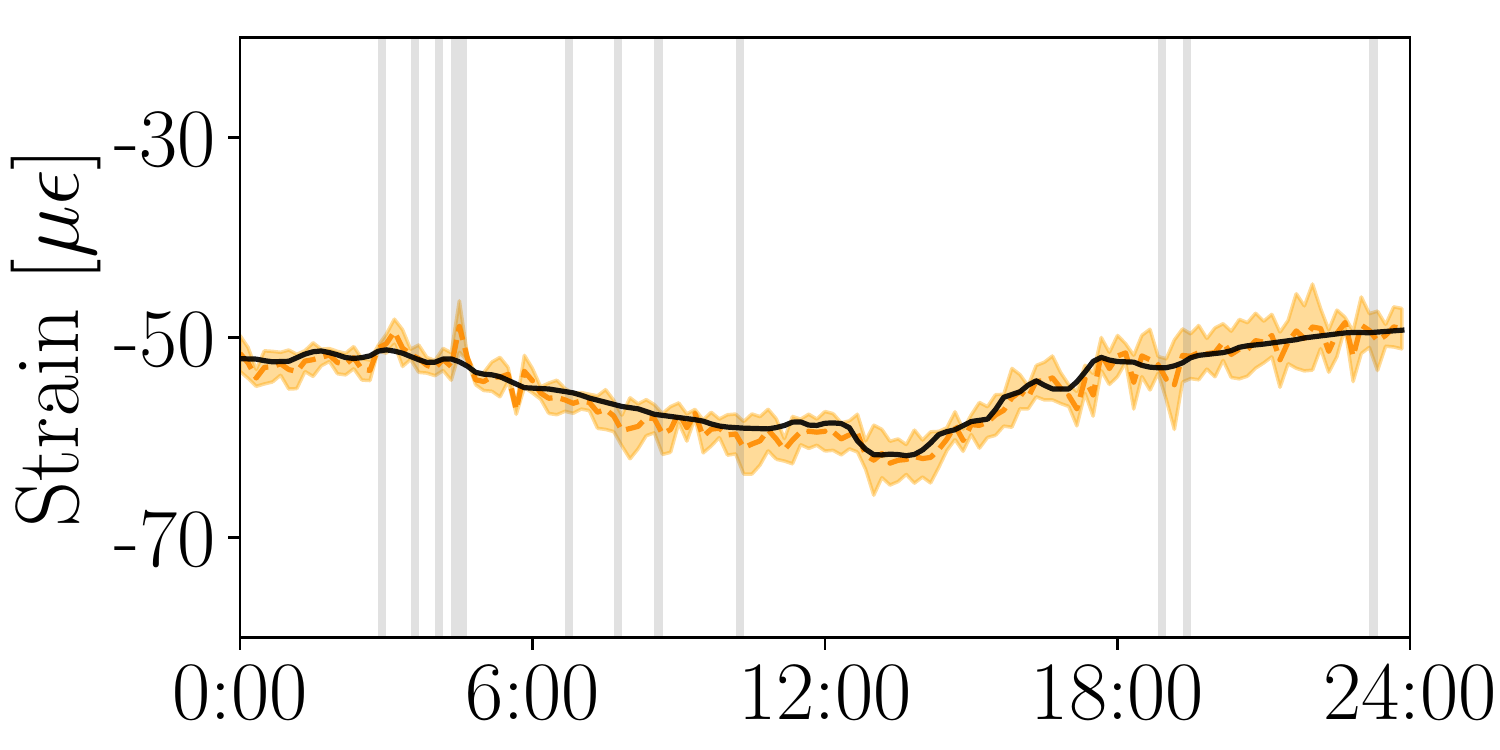}
	    \label{fig:rm_pred_uq_1}} 
	    \subfigure[RM (Dec. 15, 2017)]{\includegraphics[width=0.3\linewidth]{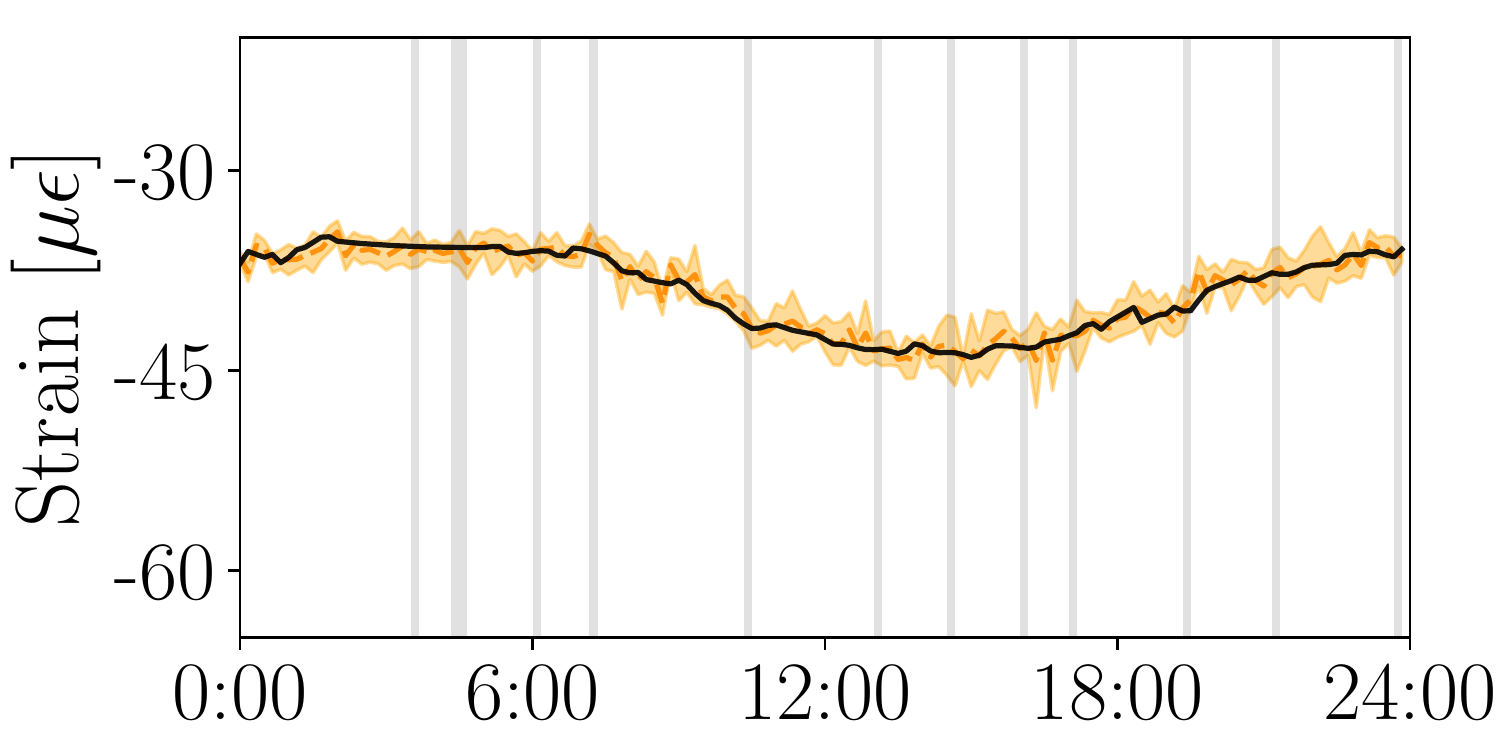}
	    \label{fig:rm_pred_uq_2}} 
	    \subfigure[RM (Aug. 22, 2018)]{\includegraphics[width=0.3\linewidth]{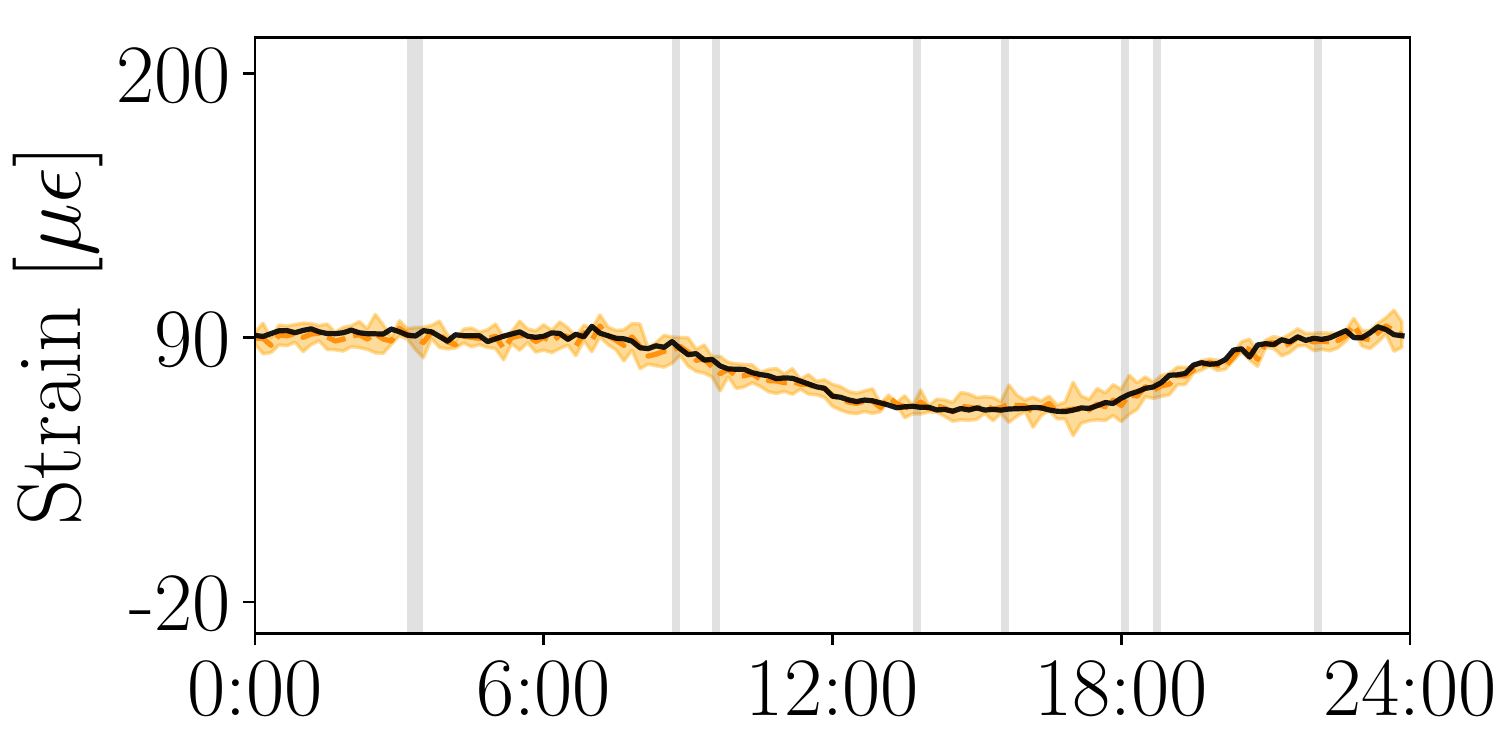}
	    \label{fig:rm_pred_uq_3}} 
	    \subfigure[SM (Mar. 10, 2016)]{\includegraphics[width=0.3\linewidth]{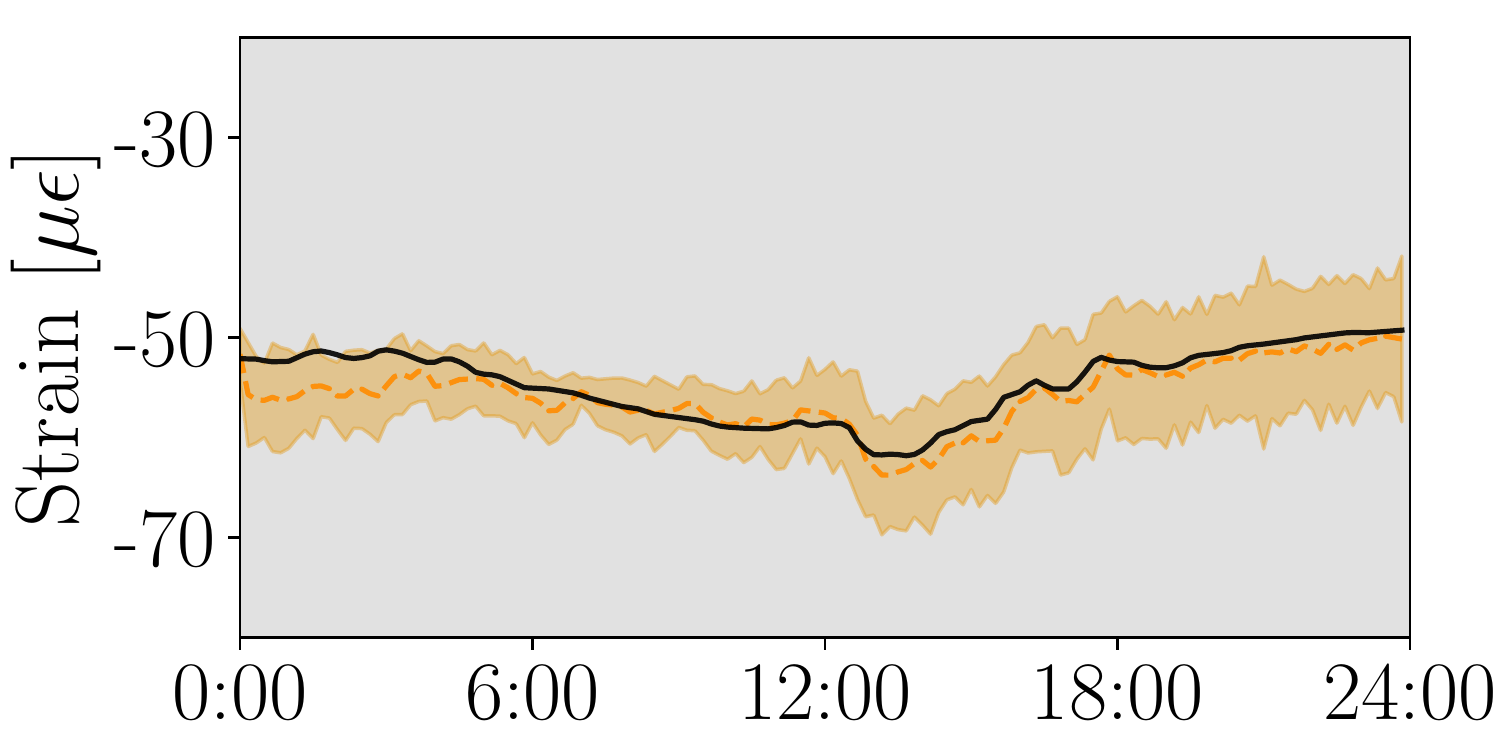}
	    \label{fig:sm_pred_uq_1}} 
	    \subfigure[SM (Dec. 15, 2017)]{\includegraphics[width=0.3\linewidth]{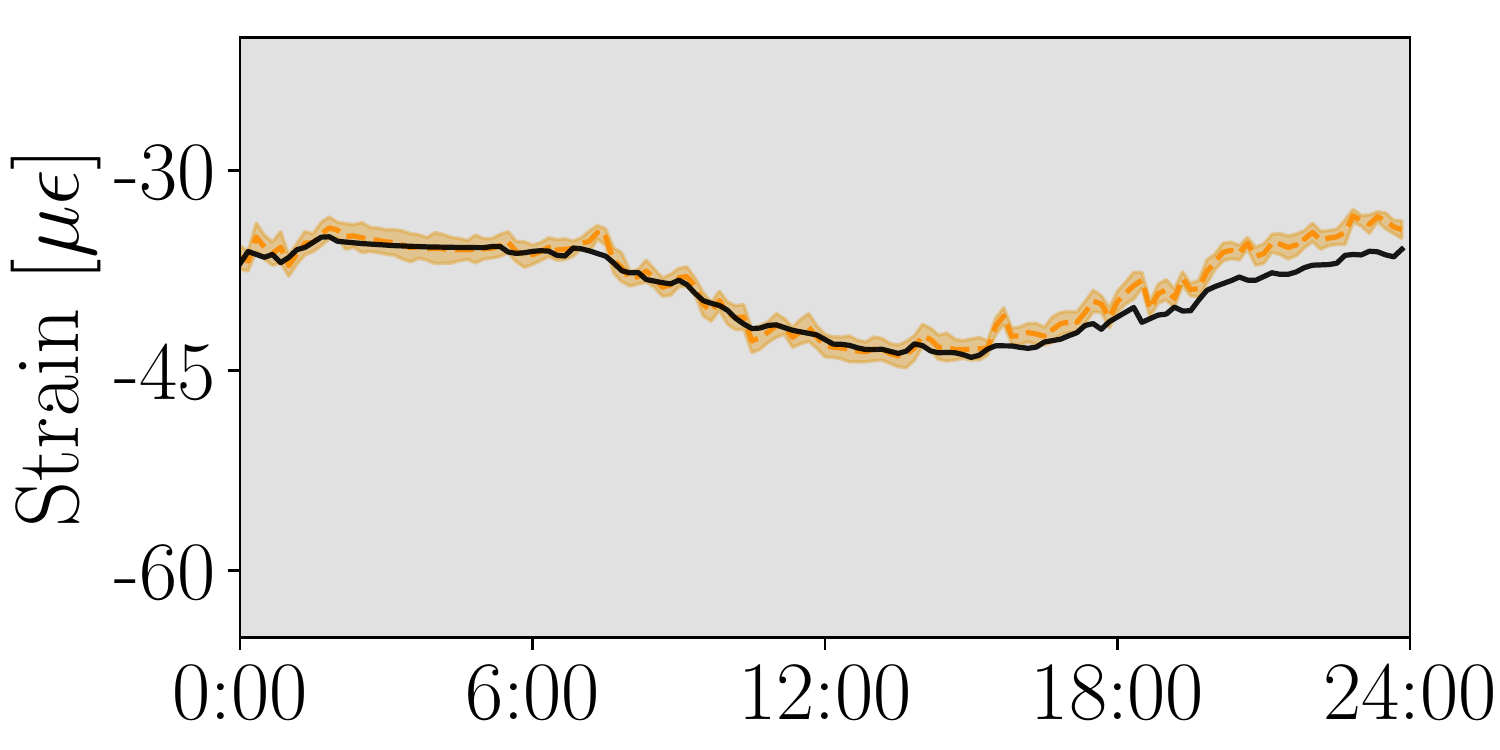}
	    \label{fig:sm_pred_uq_2}} 
	    \subfigure[SM (Aug. 22, 2018)]{\includegraphics[width=0.3\linewidth]{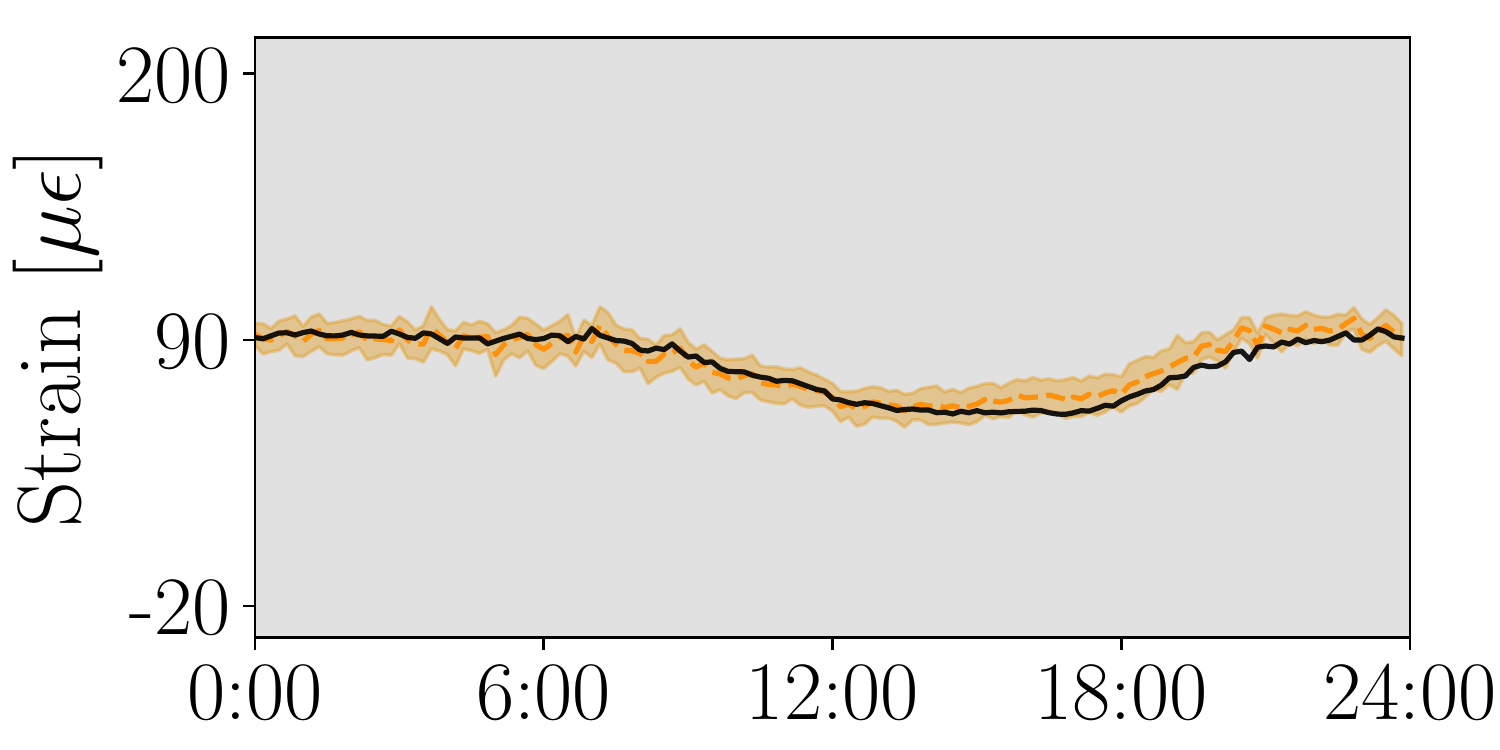}
	    \label{fig:sm_pred_uq_3}} 
	    \subfigure[MM for Case 1 (Mar. 10, 2016)]{\includegraphics[width=0.3\linewidth]{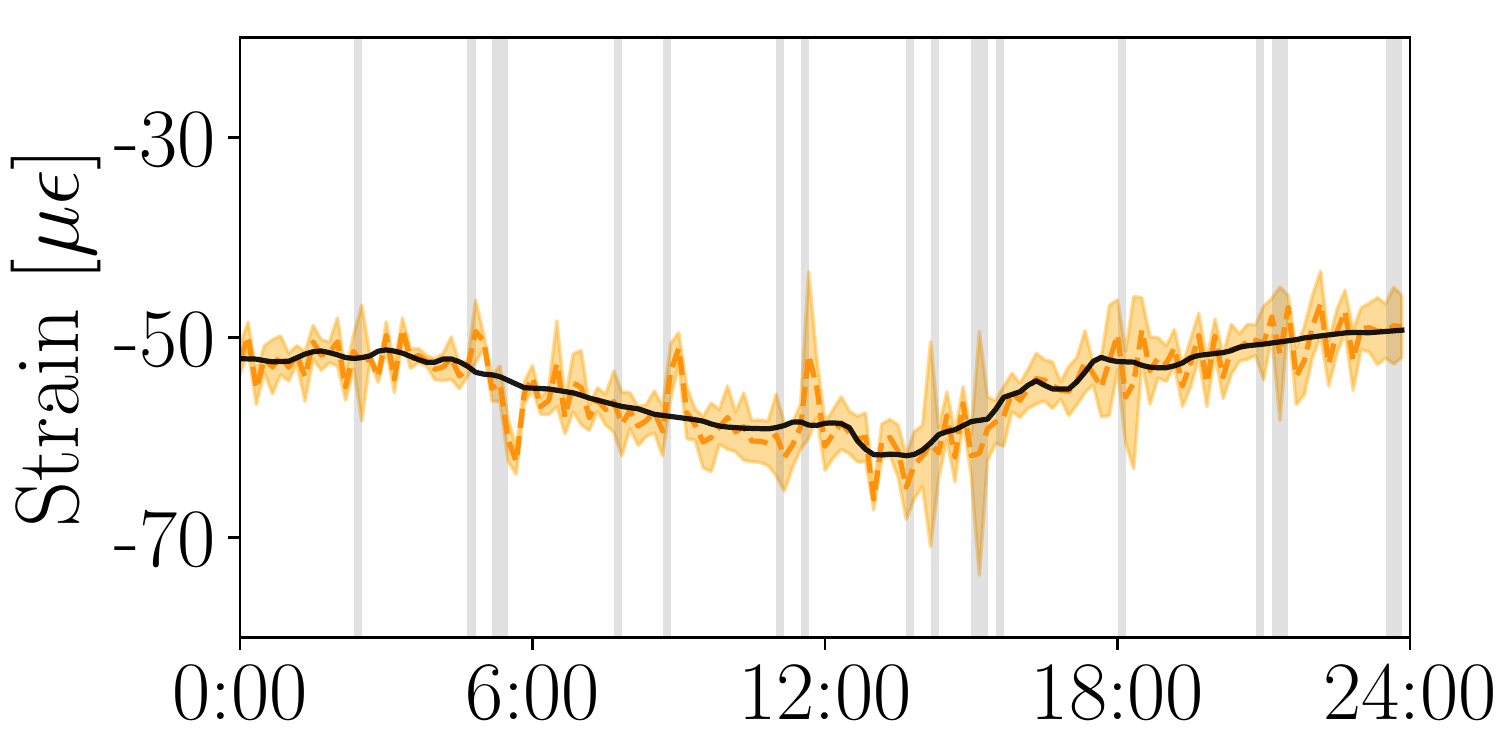}
	    \label{fig:mix1_pred_uq_1}} 
	    \subfigure[MM for Case 1 (Dec. 15, 2017)]{\includegraphics[width=0.3\linewidth]{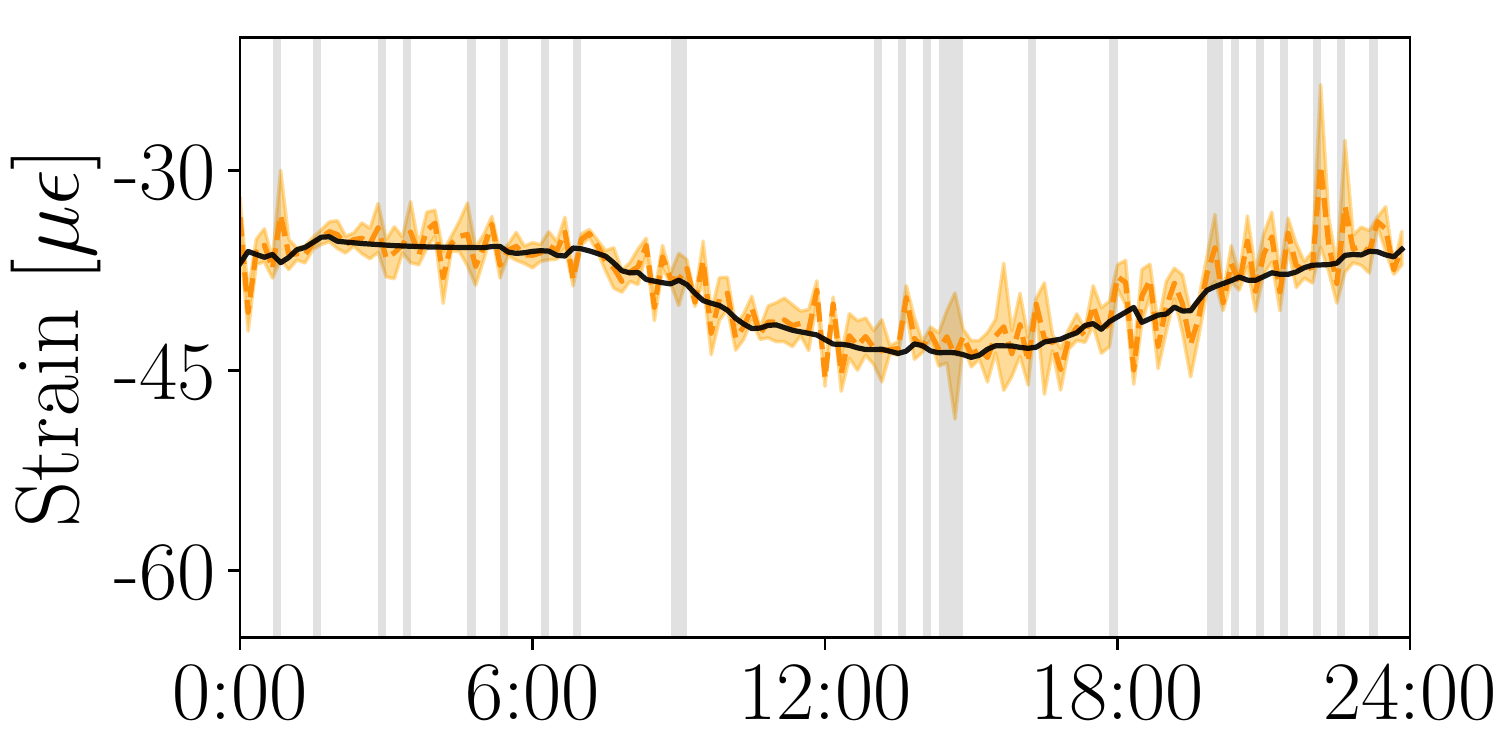}
	    \label{fig:mix1_pred_uq_2}} 
	    \subfigure[MM for Case 1 (Aug. 22, 2018)]{\includegraphics[width=0.3\linewidth]{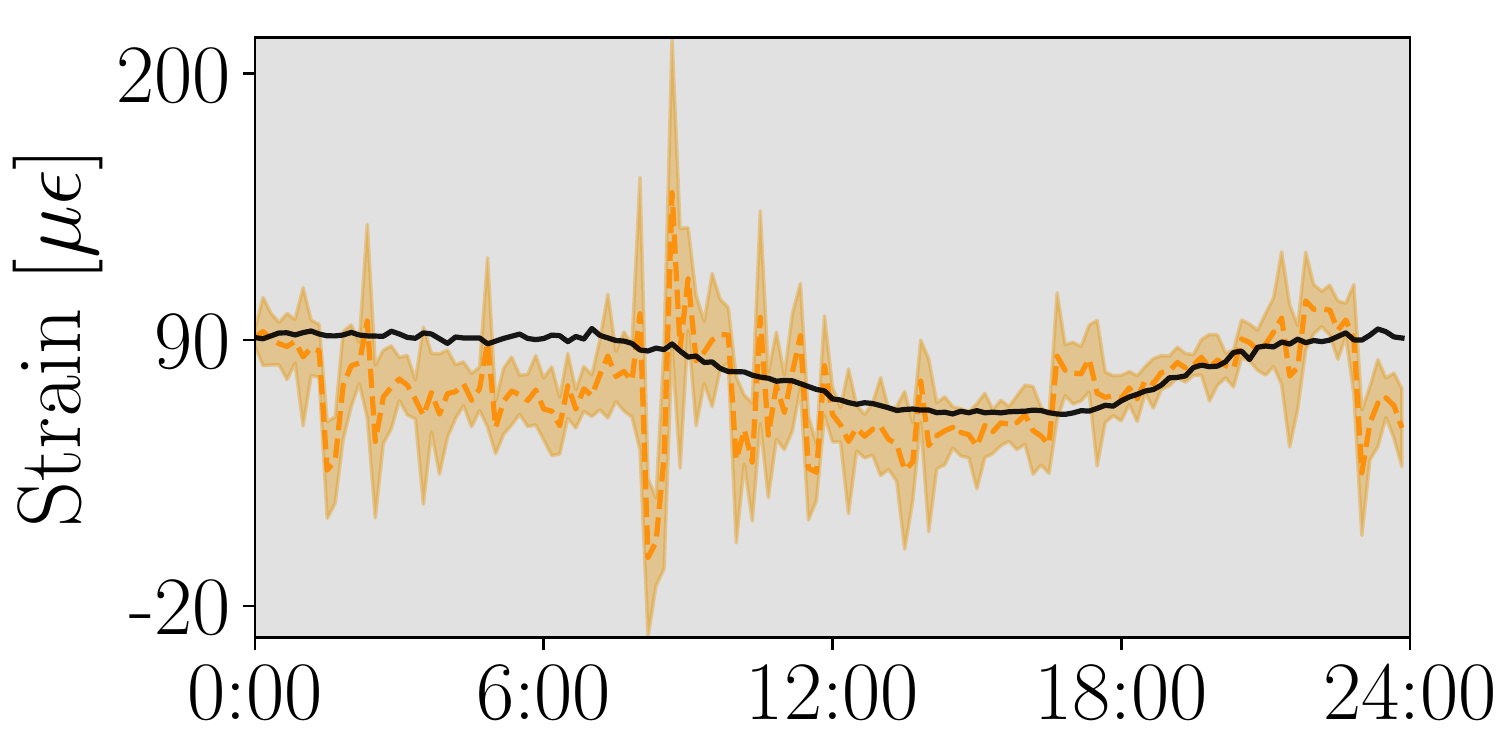}
	    \label{fig:mix1_pred_uq_3}} 
	    \subfigure[MM for Case 2 (Mar. 10, 2016)]{\includegraphics[width=0.3\linewidth]{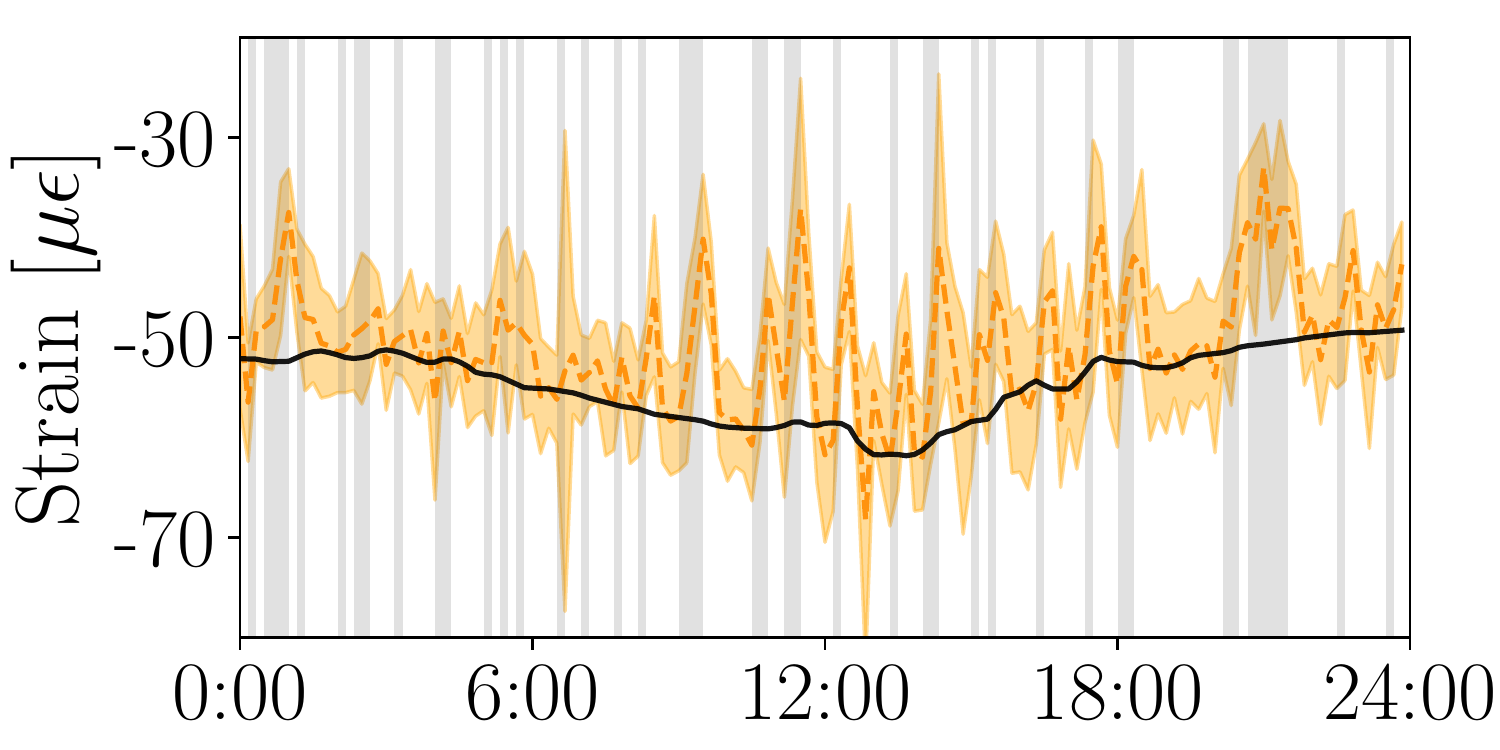}
	    \label{fig:mix2_pred_uq_1}} 
	    \subfigure[MM for Case 2 (Dec. 15, 2017)]{\includegraphics[width=0.3\linewidth]{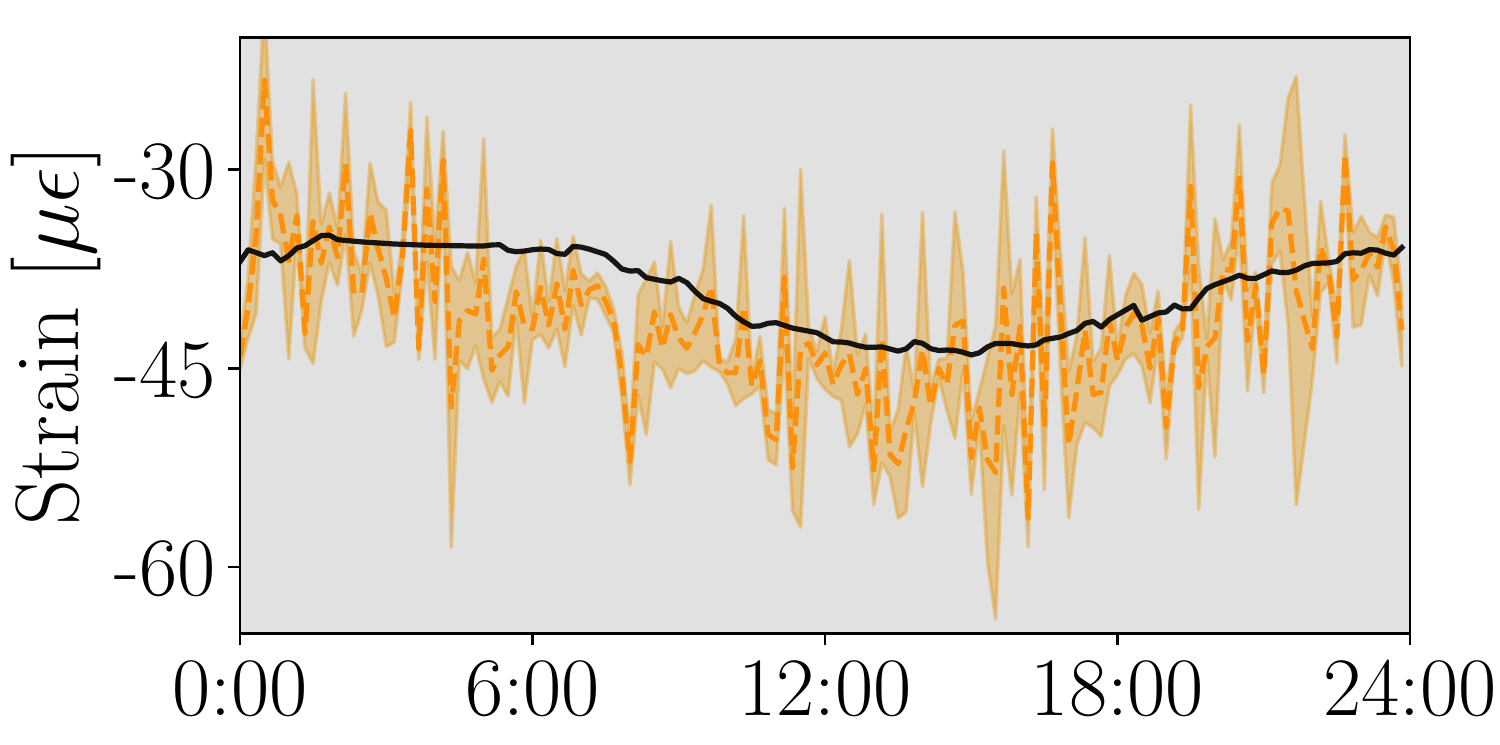}
	    \label{fig:mix2_pred_uq_2}} 
	    \subfigure[MM for Case 2 (Aug. 22, 2018)]{\includegraphics[width=0.3\linewidth]{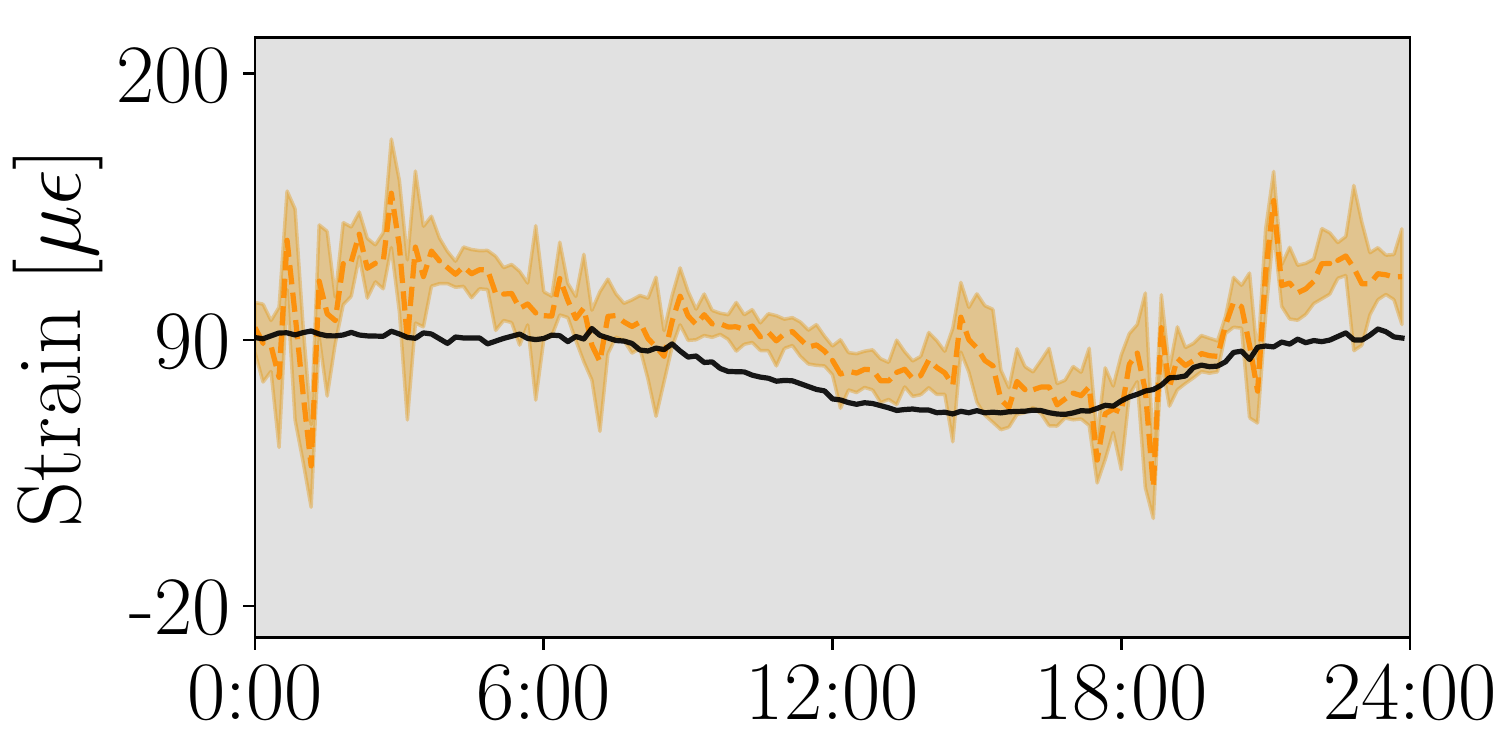}
	    \label{fig:mix2_pred_uq_3}} 
	\vspace{0pt}
	\caption{Uncertainty quantification of forecasting for four data missing cases. Note that the shading areas represent the time periods where data missing occurs, while the white box areas denote that the strain time series are successfully recorded. The black lines and the orange dashed lines depict the one-day field measurement and the predicted means respectively, and the yellow band is the area between plus/minus three standard deviations.}
	\label{fig:pred_uq_eval} 
\end{figure}

To further investigate the performance of the proposed approach, we conduct uncertainty quantification of the prediction. Thanks to the incremental learning architecture, we can achieve the convergence with fewer Monte Carlo samples (e.g., $10\sim50$) for small missing rates (e.g., 10\%). The probabilistic imputation and forecasting results are summarized in Figure \ref{fig:impu_uq_eval} and Figure \ref{fig:pred_uq_eval} respectively, where the predicted mean and three standard deviations of the strain response are shown for Sensor $\text{S}_{2}\text{-}4$ in comparison with the ground truth records on March 10, 2016, December 15, 2017 and August 22, 2018. 

For all these four missing scenarios, the imputation uncertainty is prominently smaller than the forecasting uncertainty. In the forecasting cases, it is observed that the missing data cause prediction fluctuations, which leads to deviation from the ground truth. Moreover, the forecasting uncertainty tends to be more unstable and larger when the missing rate becomes larger as shown in Figure \ref{fig:pred_uq_eval}\subref{fig:mix1_pred_uq_1}-\subref{fig:mix1_impu_uq_3} and \subref{fig:mix2_pred_uq_1}-\subref{fig:mix2_pred_uq_3}.

\subsection{Missing Rate Effect}\label{S:3.4}
\begin{figure}[t!]
	\centering
	    \subfigure[Random Missing]{\includegraphics[width=0.45\linewidth]{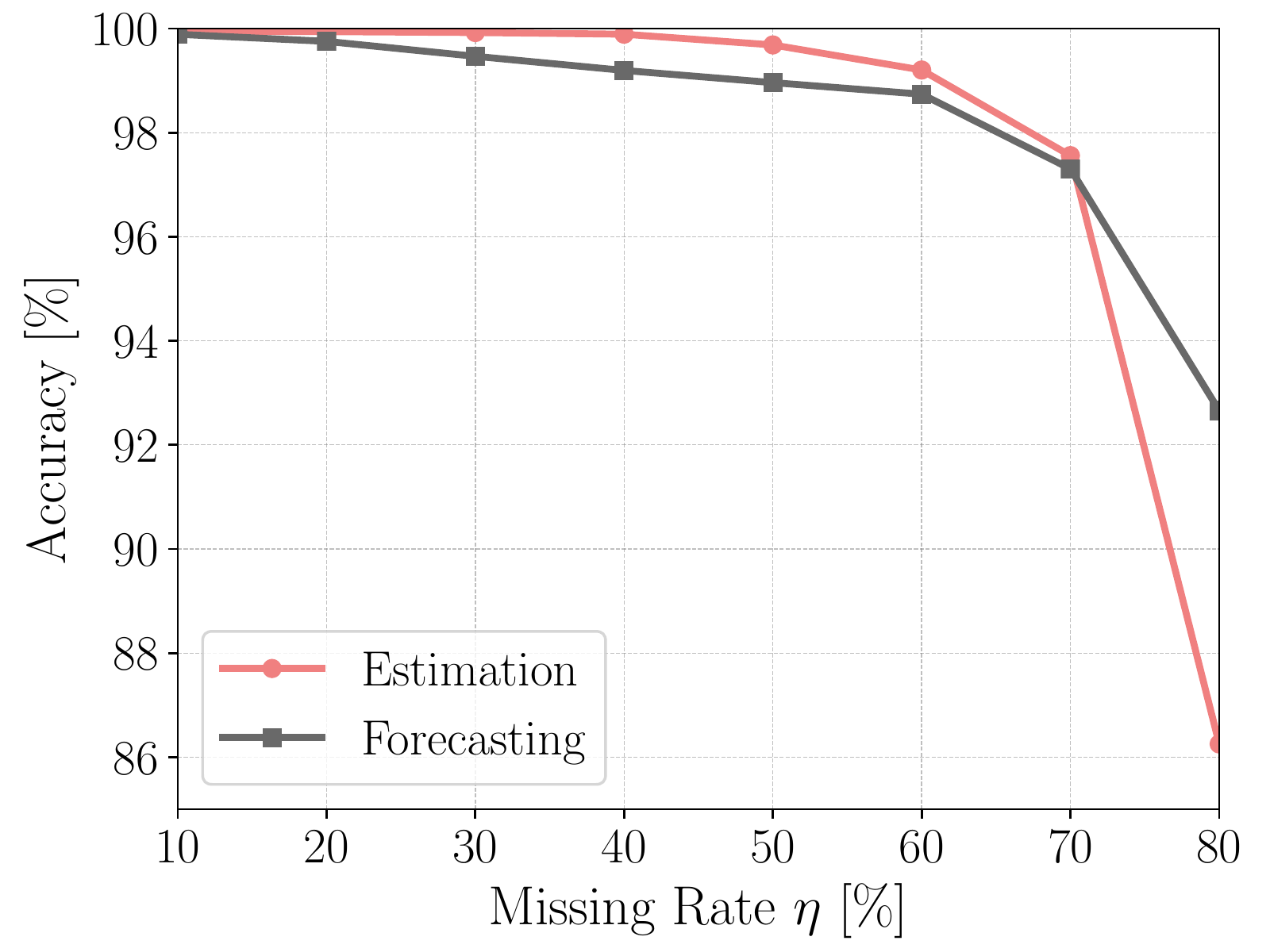}
	    \label{fig:rand_acc}} 
	    \hspace{1em}
	    \subfigure[Structured Missing]{\includegraphics[width=0.45\linewidth]{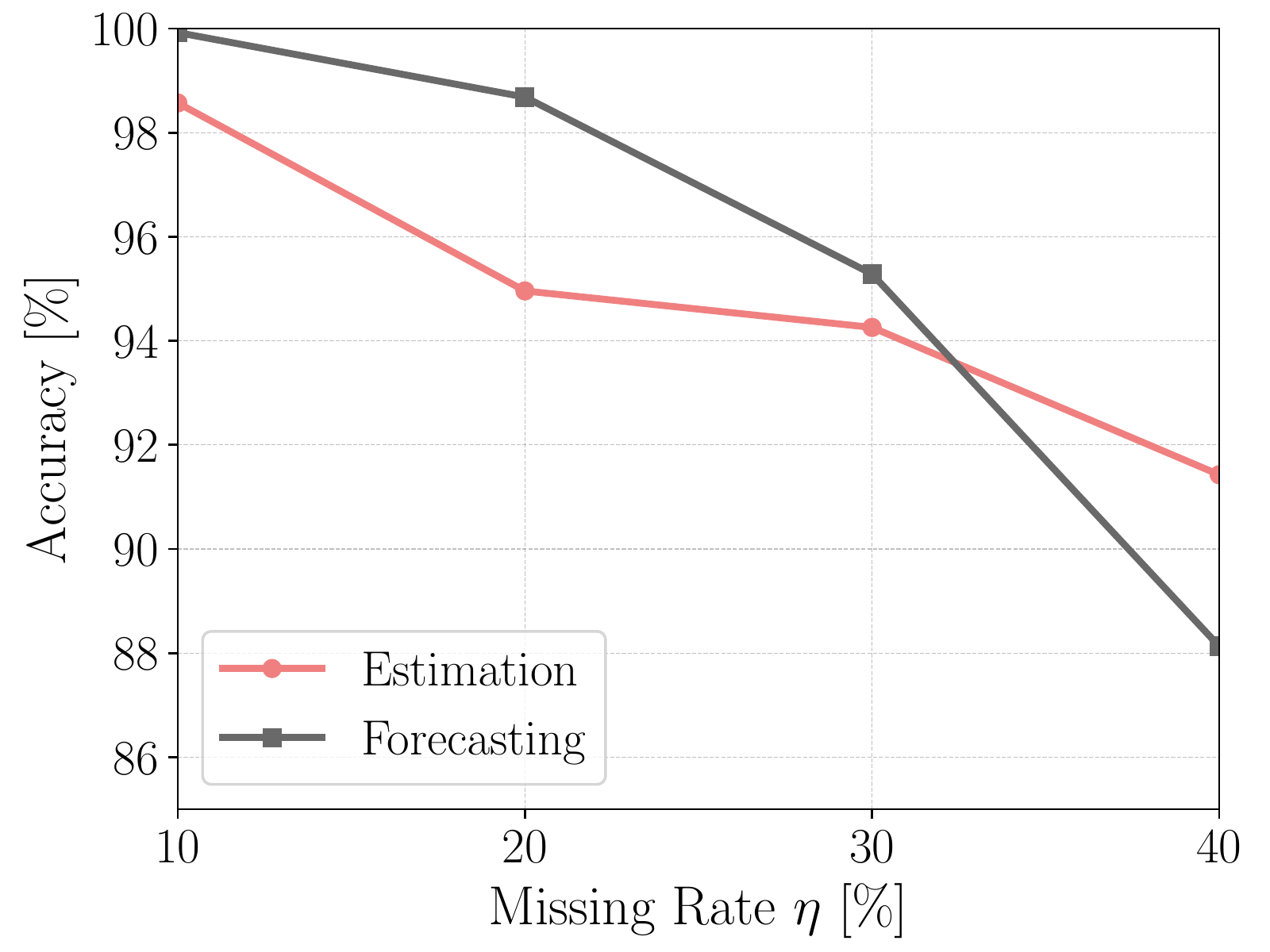}
	    \label{fig:struc_acc}}
	\vspace{0pt}
	\caption{The accuracy of imputation and forecasting with respect to different data missing rates.}
	\label{fig:accuracy} 
\end{figure}

In addition, we perform parametric studies on the influence of data missing rate on the accuracy of imputation and forecasting. The test experiments arrange the first 80\% portion of the recorded data for missing data recovery and the rest 20\% for response forecasting. Figure \ref{fig:accuracy} summarizes the parametric study result. For the random missing scenario shown in Figure \ref{fig:accuracy}\subref{fig:rand_acc}, the proposed method presents outstanding accuracy (over $95\%$ for both imputation and forecasting) given the missing rate $\eta$ up to $70\%$. The extreme case we consider here is the missing condition with $\eta = 80\%$. Nevertheless, the proposed approach still achieves over $86\%$ missing data recovery accuracy and more than $92\%$ forecasting accuracy. For the structured missing scenario (more practical and commonly seen in real-world applications), it is naturally more challenging to recover the missing data and forecast the response compared with the ideal random missing. As is seen in Figure \ref{fig:accuracy}\subref{fig:struc_acc}, the capacity limit of the proposed Bayesian tensor learning method shows to be $\eta = 40\%$ where the imputation accuracy surpasses $91\%$ while the forecasting has over $88\%$ accuracy. Interestingly, the mixed missing scenarios demonstrate quite perfect imputation and forecasting accuracy, namely, 99.78\% for imputation and 98.43\% for forecasting in Case 1, and 99.59\% for imputation and 98.00\% for forecasting in Case 2. This result is closely related to optimal tensor rank selection which is discussed in Section \ref{S:3.5}.

\subsection{Rank Analysis}\label{S:3.5}
\begin{figure}[t!]
	\centering
	    \subfigure[Random Missing]{\includegraphics[width=0.45\linewidth]{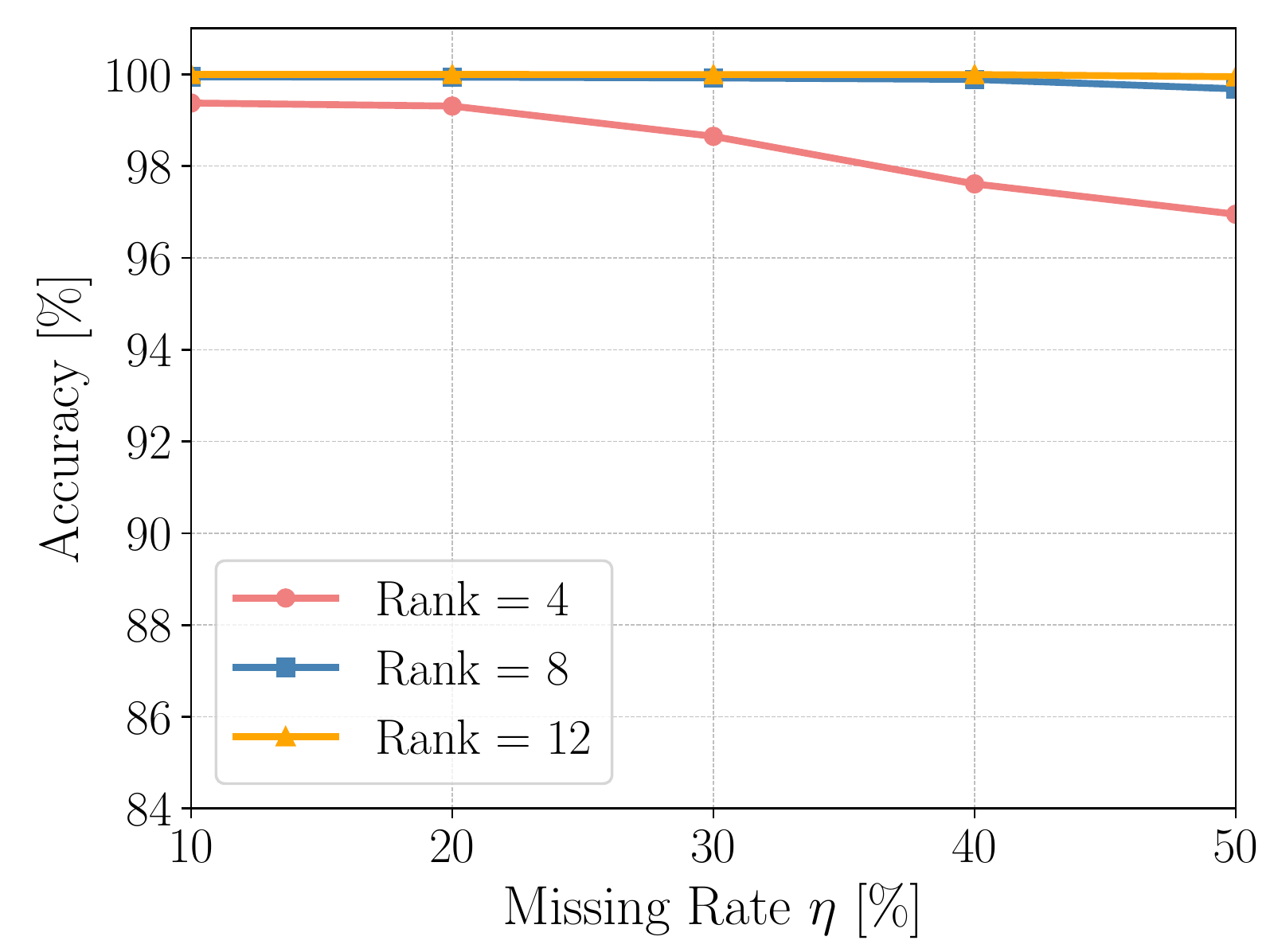}
	    \label{fig:rank_compare_rand}} 
	    \hspace{1em}
	    \subfigure[Structured Missing]{\includegraphics[width=0.45\linewidth]{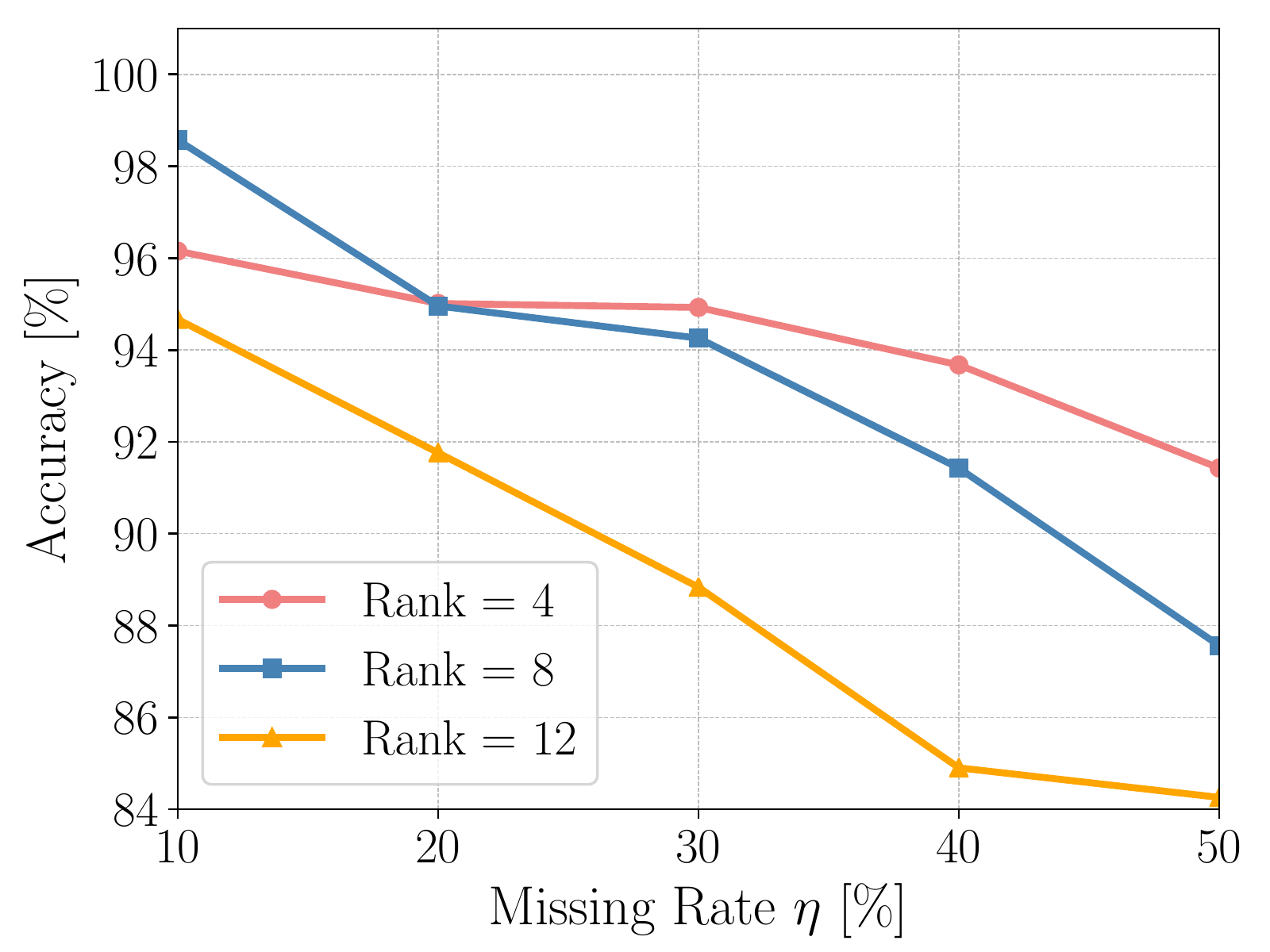}
	    \label{fig:rank_compare_struc}}
	\vspace{0pt}
	\caption{The performance imputation with respect to different tensor ranks.}
	\label{fig:rank_compare} 
\end{figure}

There exist many recent researches attempting to reveal the effect of imperfect data on tensor representation \cite{fan2017hyperspectral,chen2017general,liang2019learning,chang2020weighted}. According to \cite{liang2019learning}, it is believed that clean datasets exhibit correlations across time and modalities while the imperfect data with incomplete values break these natural correlations and lead to the requirement of a higher rank. Inspired by this study, we also quantitatively investigate the prediction performance of the proposed Bayesian tensor learning method with different ranks (e.g., 4, 8 and 12) under different data missing scenarios (e.g., random and structured), and summarize the result in Figure \ref{fig:rank_compare}. In particular, we test the imputation capability. As shown in Figure \ref{fig:rank_compare}\subref{fig:rank_compare_rand}, with the increasing rank, the estimation achieves a better accuracy in the random missing scenario, which agree with the observation in \cite{liang2019learning}. In other words, random missing destroys the spatiotemporal correlations so that we should increase the tensor rank for a more accurate result when dealing with this type of imperfect data condition.

For the structured missing scenario (see Figure \ref{fig:rank_compare}\subref{fig:rank_compare_struc}), it is surprising to see that we get a higher accuracy of missing data recovery with a lower rank under different missing rates. We empirically extrapolate that continuous element missing helps to build a more correlated tensor structure.

\section{Conclusions and Discussions}\label{S:4}
This paper presents an incremental Bayesian tensor learning method for spatiotemporal data imputation and response forecasting for SHM applications, with the incorporation of the AR process which contributes to the temporal feature modeling in an incremental learning scheme. With the existence of temperature data, the tensor model can easily gain a low-rank structure and utilize the correlation between strain and temperature for robust prediction of the strain response. In our validation experiments, we both consider the ideal random missing scenario and a more realistic missing condition--structured missing. Based on the learned latent features, the accurate estimation and forecasting results show the satisfactory performance of the proposed approach for processing incomplete SHM recordings, with uncertainty quantification capability. In addition, the extreme cases illustrate that acceptable imputation and forecasting accuracy can retain for the missing rate up to $\eta=80\%$ in random missing and up to $\eta=40\%$ in structured missing. Furthermore, the investigation into rank selection has revealed that a lower rank helps achieve better prediction performance for structured missing, while a higher rank is preferred for random missing.

There are three highlights of the proposed method. The first and the most notable significance is that we model the temporal dependency via the latent features instead of using incomplete data directly, which offers a robust and flexible modeling scheme for multivariate time series data. Secondly, it is unnecessary to know which of the entries in the tensor data are incomplete beforehand. Thirdly, the fully Bayesian method can avoid overfitting and relax parameter tuning. In the meanwhile, it also draws unfavorable deficiency of computational complexity due to the use of approximated Bayesian inference. Notwithstanding the most time consuming process remains in the imputation process, our proposed incremental Bayesian tensor learning algorithm can drastically reduce the computational time and make it efficient for data imputation and response forecasting for continuous SHM with streaming yet missing data.

The present study demonstrates that tensor learning has potential to become a promising area in SHM applications. Some future research directions and outlook are proposed herein. Firstly, as long as we have enough sensor locations and monitoring zones (e.g., distributed sensing), the higher order tensor decomposition for imputation and forecasting should be explored thanks to its possibility of outperforming the second-order tensor factorization \cite{ran2016tensor}. Secondly, the proposed approach can be extended to tackle issues of SHM data anomaly detection and de-noising on account of the power of tensor representation. Last but not least, the spatial feature can be described in a more realistic way by considering graph kernels \cite{yu2016temporal,bahadori2014fast}, which will be worthy to investigate.

\section*{Acknowledgement}
The authors would like to thank the Department of Bridge and Structural Engineering, China Merchants Chongqing Communications Technology Research and Design Institute Co. Ltd., for sharing the datasets which were used to validate the proposed methodology. In addition, the authors greatly acknowledge the open source codes \cite{code2020} of Bayesian temporal matrix factorization (BTMF), which were leveraged for numerical analyses in this study.


\bibliographystyle{elsarticle-num}
\bibliography{references.bib}

\end{document}